\documentclass[10pt,twocolumn,letterpaper]{article}

\usepackage[pagenumbers]{cvpr}

\makeatletter
\@namedef{ver@everyshi.sty}{}
\makeatother

\usepackage{times}
\usepackage{epsfig}
\usepackage{graphicx}
\usepackage{amsmath}
\usepackage{amssymb}

\usepackage[utf8]{inputenc}
\usepackage{multirow}
\usepackage{rotating}
\usepackage{verbatim}
\usepackage{booktabs}
\usepackage{array}
\usepackage{textcomp, gensymb}
\usepackage{diagbox}
\usepackage[T1]{fontenc}    
\usepackage{url}            
\usepackage{booktabs}       
\usepackage{amsfonts}       
\usepackage{nicefrac}       
\usepackage{microtype}      
\usepackage{setspace}
\usepackage{multirow}
\usepackage{amssymb}
\usepackage{diagbox}
\usepackage{mathtools}
\usepackage{wrapfig}
\usepackage{diagbox} 
\usepackage{wrapfig,lipsum,booktabs}
\usepackage{caption}
\usepackage{appendix}
\usepackage{textcomp}
\usepackage{pgfkeys}
	\def\addlegendimage{\csname pgfplots@addlegendimage\endcsname}
\usepackage{tkz-kiviat}
\usepackage{xfp}
\newcommand\blfootnote[1]{%
  \begingroup
  \renewcommand\thefootnote{}\footnote{#1}%
  \addtocounter{footnote}{-1}%
  \endgroup
}

\usepackage{pifont}

\newcommand{\xmark}{\text{\ding{55}}}
\usepackage{tikz}
\usepackage{tkz-kiviat}

\usepackage{color}
\usepackage{xcolor}
\definecolor{citecolor}{rgb}{0.21,0.49,0.74}
\usepackage{colortbl}
\usepackage{array}

\usepackage[pagebackref=true,breaklinks=true,letterpaper=true,citecolor=citecolor,colorlinks,bookmarks=false]{hyperref}

\definecolor{tblue}{RGB}{80,80,245}
\definecolor{tred}{RGB}{250,100,100}
\definecolor{tgreen}{RGB}{32,178,170}
\usepackage{soul}

\newcommand{\tablefirst}{\cellcolor{gray!20}}

\newcommand{\modelname}[0]{\mbox{OmniGlue}\xspace}

\newcommand{\green}[1]{{\color{tgreen}#1}}

\newcolumntype{x}[1]{>{\centering\arraybackslash}p{#1pt}}

\newcommand{\app}{\raise.17ex\hbox{$\scriptstyle\sim$}}

\def\x{$\times$}

\newlength\savewidth\newcommand\shline{\noalign{\global\savewidth\arrayrulewidth
  \global\arrayrulewidth 1pt}\hline\noalign{\global\arrayrulewidth\savewidth}}
\newcommand{\tablestyle}[2]{\setlength{\tabcolsep}{#1}\renewcommand{\arraystretch}{#2}\centering\footnotesize}

\usepackage{amssymb}
\usepackage{pifont}

\def\paperID{1523}
\def\confName{CVPR}
\def\confYear{2024}

\begin{document}

\title{\modelname{}: Generalizable Feature Matching with Foundation Model Guidance}

\author{
Hanwen Jiang$^\ast$ \and
Arjun Karpur$^\dagger$ \and
Bingyi Cao$^\dagger$ \and
Qixing Huang$^\ast$ \and
Andr\'{e} Araujo$^\dagger$ \\
\and
$^\ast$University of Texas at Austin\\
{
\tt\small (hwjiang,huangqx)@cs.utexas.edu}
\and
$^\dagger$Google Research \\
{
\tt\small (arjunkarpur,bingyi,andrearaujo)@google.com
}
}

\maketitle

\begin{abstract}

The image matching field has been witnessing a continuous emergence of novel learnable feature matching techniques, with ever-improving performance on conventional benchmarks.
However, our investigation shows that despite these gains, their potential for real-world applications is restricted by their limited generalization capabilities to novel image domains.
In this paper, we introduce \modelname{}, the first learnable image matcher that is designed with generalization as a core principle.
\modelname{} leverages broad knowledge from a vision foundation model to guide the feature matching process, boosting generalization to domains not seen at training time.
Additionally, we propose a novel keypoint position-guided attention mechanism which disentangles spatial and appearance information, leading to enhanced matching descriptors.
We perform comprehensive experiments on a suite of $7$ datasets with varied image domains, including scene-level, object-centric and aerial images.
\modelname{}'s novel components lead to relative gains on unseen domains of $20.9\%$ with respect to a directly comparable reference model, while also outperforming the recent LightGlue method by $9.5\%$ relatively.
Code and model can be found at \url{https://hwjiang1510.github.io/OmniGlue}.



\end{abstract}

\vspace{-0.15in}
\section{Introduction}

Local image feature matching techniques provide fine-grained visual correspondences between two images~\cite{Ma2020ImageMF}, which are critical for achieving accurate camera pose estimation~\cite{sarlin2020superglue, roessle2023end2end} and 3D reconstruction~\cite{Cadena2016PastPA, schonberger2016colmap, goesele2006multi, jiang2022few}.
The past decade has witnessed the evolution from hand-crafted~\cite{Lowe2004sift,Bay2006SURF} to learning-based image features~\cite{verdie2015tilde,yi2016lift,detone2018superpoint,Revaud2019R2D2,potje2023enhancing}.
More recently, 
novel learnable image matchers have been proposed~\cite{sarlin2020superglue, Sun2021LoFTR, pdcnet+, Lindenberger2023LightGlue, edstedt2023dkm}, demonstrating ever-improving performance on conventional benchmarks~\cite{Li2018MegaDepth, dai2017scannet, balntas2017hpatches}.

\begin{figure}[t]
\begin{center}
  	\resizebox{\linewidth}{!}{
        \usetikzlibrary{arrows}

\newcommand{\lattice}{4}
\newcommand{\naxis}{6}
\newcommand{\amax}{8.6}
\newcommand{\amin}{6.8}

\newcommand{\bmax}{22.4}
\newcommand{\bmin}{16.4}

\newcommand{\cmax}{13.2}
\newcommand{\cmin}{6.2}

\newcommand{\dmax}{12.4}
\newcommand{\dmin}{4.2}

\newcommand{\emax}{47.4}
\newcommand{\emin}{25.8}

\newcommand{\fmax}{31.3}
\newcommand{\fmin}{15.0}

\newcommand{\baseline}{0.5}
\newcommand{\distance}[3]{
    {((#3) - (#2))/ ((#1) - (#2)) * (1-\baseline)+\baseline}
}

\newcommand\ColorBox[1]{\textcolor{#1}{\rule{3ex}{3ex}}}

\newcommand{\annotMark}[5]{
	\pgfmathsetmacro{\xcor}{#3*cos{(#1*#2)}/(1/#4)};
	\pgfmathsetmacro{\ycor}{#3*sin{(#1*#2)}/(1/#4)};
	\draw (\xcor,\ycor)node[anchor=south]{#5};
}

\newcommand{\vu}{\mathbf{u}}
\newcommand{\vv}{\mathbf{v}}
\newcommand{\calT}{\mathcal{T}}
\newcommand{\calX}{\mathcal{X}}
\newcommand{\calY}{\mathcal{Y}}
\DeclareRobustCommand{\hlGray}[1]{{\sethlcolor{Gray}\hl{#1}}}
\newcommand{\thickhline}{\Xhline{3\arrayrulewidth}}

\definecolor{citecolor}{HTML}{0071bc}
\definecolor{color_ao}{gray}{0.5}
\definecolor{color_our}{rgb}{0.66,0.82,0.56}
\definecolor{color_pre}{rgb}{0.52,0.59,0.69}
\definecolor{Gray}{gray}{0.9}
\definecolor{LighterGray}{gray}{0.93}
\definecolor{LightGrayForTableRule}{gray}{0.92}
\definecolor{DarkGray}{gray}{0.5}
\definecolor{Black}{rgb}{0.0, 0.0, 0.0}
\definecolor{NiceBlue}{rgb}{0.11764705882352941, 0.5647058823529412, 1.0}
\definecolor{NiceGreen}{rgb}{0.0, 0.5, 0.0}

\begin{tikzpicture}[rotate=30, scale=0.95,every node/.style={inner sep=-15,outer sep=-15}]
	\tkzKiviatDiagram[lattice=\lattice, gap=1.0, step=1, label space=1.6]
	{\Large GSO-\\Hard,
		\Large DeepAerial,
		\Large NAVI-MultiView,
	    \Large NAVI-\\Wild,
		\Large MegaDepth \\(in-domain),
		\Large ScanNet }
	
		\tkzKiviatLine[thick, fill=color_our!50, color=color_our, opacity=0.5](
		\fpeval{(\distance{\amax}{\amin}{8.6})*\lattice},
		\fpeval{(\distance{\bmax}{\bmin}{22.4})*\lattice},
		\fpeval{(\distance{\cmax}{\cmin}{13.2})*\lattice},
		\fpeval{(\distance{\dmax}{\dmin}{12.4})*\lattice},
		\fpeval{(\distance{\emax}{\emin}{47.4})*\lattice},
		\fpeval{(\distance{\fmax}{\fmin}{31.3})*\lattice})
		\tkzKiviatLine[thick, fill=NiceBlue, color=NiceBlue, opacity=0.5](
        \fpeval{(\distance{\amax}{\amin}{7.2})*\lattice},
        \fpeval{(\distance{\bmax}{\bmin}{16.4})*\lattice},
        \fpeval{(\distance{\cmax}{\cmin}{11.8})*\lattice},
        \fpeval{(\distance{\dmax}{\dmin}{10.6})*\lattice},
        \fpeval{(\distance{\emax}{\emin}{42.2})*\lattice},
        \fpeval{(\distance{\fmax}{\fmin}{25.5})*\lattice})
		\tkzKiviatLine[thick, fill=gray!50, color=gray, opacity=0.5](
        \fpeval{(\distance{\amax}{\amin}{6.8})*\lattice},
        \fpeval{(\distance{\bmax}{\bmin}{17.5})*\lattice},
        \fpeval{(\distance{\cmax}{\cmin}{6.2})*\lattice},
        \fpeval{(\distance{\dmax}{\dmin}{4.2})*\lattice},
        \fpeval{(\distance{\emax}{\emin}{25.8})*\lattice},
        \fpeval{(\distance{\fmax}{\fmin}{4.6})*\lattice})

	\annotMark{0}{360/\naxis}{2.0}{1}{\Large \amin};
	\annotMark{0}{360/\naxis}{4.4}{1}{\Large \amax};
	\annotMark{1}{360/\naxis}{2.0}{1}{\Large \bmin};
	\annotMark{1}{360/\naxis-10}{4.7}{1}{\Large \bmax};
	\annotMark{2}{360/\naxis}{2.0}{1}{\Large \cmin};
	\annotMark{2}{360/\naxis}{4.6}{1}{\Large \cmax};
	\annotMark{3}{360/\naxis}{1.7}{1}{\Large \dmin};
	\annotMark{3}{360/\naxis}{3.8}{1}{\Large \dmax};
	\annotMark{4}{360/\naxis}{1.7}{1}{\Large \emin};
	\annotMark{4}{360/\naxis-10}{3.8}{1}{\Large \emax};
	\annotMark{5}{360/\naxis}{2.0}{1}{\Large \fmin};
	\annotMark{5}{360/\naxis}{3.7}{1}{\Large \fmax};

	\node[anchor=south west,xshift=-0pt,yshift=20pt] at (current bounding box.south east)
{
	\begin{tabular}{@{}lp{4cm}@{}}
		\ColorBox{gray!50} & \Large SIFT+MNN \\
		\ColorBox{NiceBlue} & \Large SuperGlue \\
		\ColorBox{color_our} & \Large \textbf{OmniGlue (ours)}
	\end{tabular}
};
\end{tikzpicture}
    }
\end{center}
\vspace{-0.1in}
  \caption{
  \textbf{\modelname{} is a generalizable learnable matcher.}
  Introducing foundation model guidance and an enhanced attention mechanism, \modelname{} learns effective image matching that transfers well to image domains not seen during training.
  We compare it against reference methods SIFT \cite{Lowe2004sift} and SuperGlue \cite{sarlin2020superglue}, with substantial improvements on a suite of diverse datasets: outdoor scenes (MegaDepth-1500 \cite{Li2018MegaDepth} pose AUC@$5\degree$), indoor scenes (ScanNet \cite{dai2017scannet} pose accuracy @$5\degree$), aerial scenes (DeepAerial \cite{park2020two} PCK@$1\%$) and object-centric images (GSO-Hard \cite{Downs2022GoogleSO} and NAVI-MultiView / NAVI-Wild \cite{jampani2023navi}, pose accuracy @$5\degree$).
  }
  \vspace{-0.15in}
\label{fig: teaser}
\end{figure}
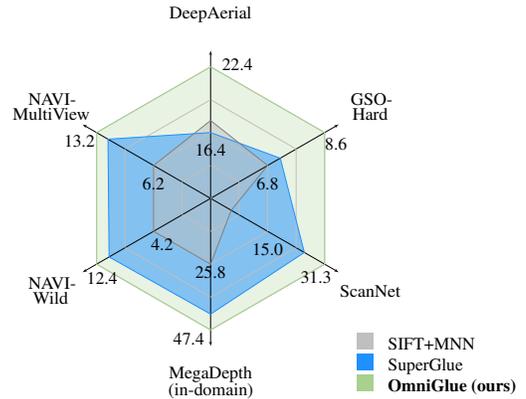

\blfootnote{$^*$This work was completed while Hanwen was an intern at Google.}

Despite substantial progress, these advancements overlook an essential aspect: the \textbf{generalization capability} of image matching models.
Today, most local feature matching research \cite{Sun2021LoFTR,edstedt2023dkm,Lindenberger2023LightGlue} focuses on specific visual domains with abundant training data (\eg, outdoor and indoor scenes), leading to models that are highly specialized for the training domain.
Unfortunately, we observe that the performance of these methods usually drops dramatically on out-of-domain data (\eg, object-centric or aerial captures), which may not even be significantly better than traditional approaches in some cases. 
For this reason, traditional domain-agnostic techniques, such as SIFT~\cite{Lowe2004sift}, are still widely used to obtain poses for downstream applications~\cite{li2020frodo,tyszkiewicz2022raytran,barron2023zipnerf,mildenhall2020nerf}.
Due to the cost of collecting high-quality correspondence annotations,
we believe it is unrealistic to assume that abundant training data would be available for each image domain, like in some other vision tasks~\cite{deng2009imagenet, lin2014coco}. Thus, the community should focus on developing architectural improvements to make learnable matching methods generalize.

Motivated by the above observations, we propose \textbf{\modelname{}}, the first learnable image matcher that is designed with generalization as a core principle.
Building on top of domain-agnostic local features~\cite{detone2018superpoint}, we introduce novel techniques for improving the generalizability of matching layers: foundation model guidance and keypoint-position attention guidance. As shown in Fig.~\ref{fig: teaser}, with the introduced techniques, we enable \modelname{} to generalize better on out-of-distribution domains while maintaining quality performance on the source domain.

Firstly, we incorporate broad visual knowledge of a foundation model.
By training on large-scale data, the foundation model, DINOv2~\cite{Oquab2023DINOv2}, performs well in diverse image domains on a variety of tasks, including robust region-level matching~\cite{Oquab2023DINOv2, jiang2023vqloc, Zhang2023ATO}.
Even though the granularity of matching results yielded from foundational models is limited, these models provide generalizable guidance on potential matching regions when a specialized matcher cannot handle the domain shift. Thus, we use DINO to guide the inter-image feature propagation process, downgrading irrelevant keypoints and encouraging the model to fuse information from potentially matchable regions.

Secondly, we also guide the information propagation process with keypoint position information.
We discover that previous positional encoding strategies \cite{sarlin2020superglue} hurt performance when the model is applied to different domains -- which motivates us to disentangle it from the matching descriptors used to estimate correspondence. We propose a novel keypoint-position guided attention mechanism designed to avoid specializing too strongly in the training distribution of keypoints and relative pose transformations.

Experimentally, we assess \modelname{}'s generalization across diverse visual domains, spanning synthetic and real images, from scene-level to object-centric and aerial datasets, with small-baseline and wide-baseline cameras.
We demonstrate significant improvements compared to previous work.
In more detail, our contributions are as follows.

\noindent\textbf{Contributions.}
\textbf{(1)} We introduce foundation model guidance to the learnable feature matching process, which leverages broad visual knowledge to enhance correspondences in domains that are not observed at training time, boosting pose estimation accuracy by up to $\textbf{5.8\%}$ ($14.4\%$ relatively).
\textbf{(2)} A new strategy for leveraging positional encoding of keypoints, which avoids an overly reliant dependence on geometric priors from the training domain, boosting cross-domain transfer by up to $\textbf{6.1\%}$ ($14.9\%$ relatively).
\textbf{(3)} We perform comprehensive experiments on $7$ datasets from varied domains, demonstrating the limited generalizability of existing matching methods and \modelname{}'s strong improvements, with relative gains of $\textbf{20.9\%}$ on average in all novel domains.
\textbf{(4)} By fine-tuning \modelname{} using limited amount of data from the target domain, we show that \modelname{} can be easily adapted with an improvement up to $\textbf{8.1\%}$ ($94.2\%$ relatively). 

\begin{figure*}[t]
\begin{center}
   \includegraphics[width=\linewidth]{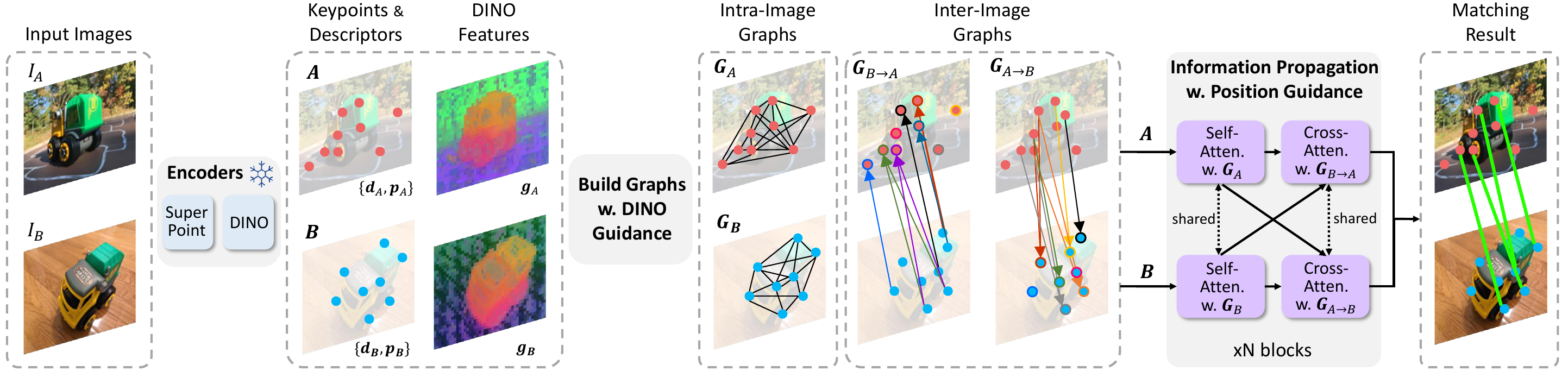}
   \vspace{-0.2in}
   \caption{
   \textbf{\modelname{} overview.} We use frozen DINO and SuperPoint to detect keypoints and extract features. Then, we build densely connected intra-image keypoint graphs and leverage DINO features to build inter-image graphs. We refine the keypoint features based on the constructed graphs, performing information propagation. In this process, we use keypoint positions solely for guidance, disentangling them from the keypoint local descriptors. Finally, the matching results are produced based on the updated keypoint local descriptors. 
   }
   \vspace{-0.25in}
   \label{fig: main}
\end{center}
\end{figure*}

\section{Related Work}

\noindent\textbf{Generalizable Local Feature Matching.} Prior to the deep learning era, researchers focused on developing generalizable local feature models.
For example, SIFT \cite{Lowe2004sift}, SURF \cite{Bay2006SURF} and ORB \cite{Rublee2011ORBAE} have been widely used for image matching tasks across diverse image domains.
Still today, many computer vision systems ignore recent advances in learnable local features and rely on hand-crafted methods, for example, to obtain poses for downstream applications~\cite{li2020frodo,tyszkiewicz2022raytran,barron2023zipnerf,mildenhall2020nerf}.
One of the main reasons for such old hand-crafted methods to continue being adopted is that most of the recent learning-based methods \cite{Tyszkiewicz2020DISKLL, Edstedt2023DeDoDeDD, Ono2018LFNet, Revaud2019R2D2, noh2017large} are specialized to domains with abundant training data, such as outdoor building scenes, and do not generalize well to other domains.
Recently, the community shifted the main focus to develop learnable image matchers, which associate local features produced by off-the-shelf methods~\cite{detone2018superpoint} or jointly learn feature description and association~\cite{Sun2021LoFTR}. 
While they demonstrate better performance compared with hand-crafted matching systems, they make the entire image matching pipeline even more domain-specific.
Our experiments show that learnable matchers specialize strongly in the training domain, with limited generalization. 
Our proposed \modelname{} improves the generalization capability of existing learnable matchers by introducing guidance from foundation models and improved positional encoding.


\noindent\textbf{Sparse Learnable Matching.} Sparse learnable image matching methods~\cite{sarlin2020superglue, Chen2021LearningTM, Lindenberger2023LightGlue} associate sparse keypoints, produced by keypoint detectors.
For example, SuperGlue~\cite{sarlin2020superglue} uses SuperPoint~\cite{detone2018superpoint} for keypoint detection and leverages the attention mechanism~\cite{Vaswani2017Attention} to perform intra- and inter-image keypoint feature propagation.
However, SuperGlues shows limited generalization capability. 
One reason is that it entangles the local descriptors and positional information of the keypoints, making the matching process overly dependent on learned positional patterns. 
It hinders the generalizability to data with different position-related matching patterns.
To solve this problem, \modelname{} proposes to disentangle them during the feature propagation, releasing the reliance on positional patterns and improving the generalization capability to images from diverse domains. 


\noindent\textbf{(Semi-)Dense Learnable Matching.}
Dense image matching methods jointly learn the image descriptors and the matching module, performing pixel-wise matching on the entire input images~\cite{Sun2021LoFTR, Chen2022ASpanFormerDI, Wang2022MatchFormerIA, edstedt2023dkm, Truong2021pdcnet}. They benefit from the end-to-end learning pipeline and demonstrate better performance in the training domain. For example, the semi-dense method LoFTR introduces a coarse-to-fine correspondence prediction paradigm~\cite{Sun2021LoFTR}. 
Another line of work directly predicts the matching results as a 4D correlation volume~\cite{edstedt2023dkm, Truong2021pdcnet}. However, we notice that some of them generalize worse on new domains compared with sparse methods. Thus, \modelname{} chooses to focus on sparse methods, which can have better potential to be generalizable due to the use of domain-agnostic local descriptors.

\noindent\textbf{Matching with Additional Image Representations.}
Leveraging robust image representations is a promising avenue toward generalizable image matching. One line of work uses geometric image representations, e.g., depth map~\cite{wang2023curvature} and NOCS map~\cite{karpur2023lfm3d}, to augment the image matching process. 
However, they are dependent on a highly accurate monocular estimation of these geometric representations. Differently, SFD2~\cite{xue2023sfd2semantic} uses semantic segmentation results to reject indistinguishable keypoints in background regions. Nevertheless, the semantic segmentation model has to be trained on each specific target domain. 
Recently, large vision models, e.g., self-supervised vision backbones~\cite{He2019moco, Caron2021dino, Oquab2023DINOv2} and Diffusion models~\cite{tang2023emergent, Zhang2023ATO, hedlin2023unsupervised} demonstrate robust semantic understanding properties. By training on large data, these models showcase strong generalization capability across diverse domains~\cite{jiang2023vqloc, liu2023zero123, jiang2023leap}, which enables them to obtain coarse patch-level matching results. However, performing matching using image features extracted by these models demonstrates limited performance on regions/keypoints without strong semantic information and the accuracy is limited~\cite{Zhang2023ATO, jiang2023doduo}.
Instead of directly incorporating these coarse signals into the keypoint features and using them to perform matching, \modelname{} uses DINOv2 features to identify potentially related regions and guide the attention-based feature refinement process. 
Thanks to the wide domain knowledge encoded in this model, \modelname{} can boost the generalization ability of our method to diverse domains.
\vspace{-2mm}
\section{OmniGlue}
\vspace{-2mm}
We first introduce the overview and technical details of our method \modelname{}. Then we compare \modelname{} with SuperGlue and LightGlue for clarifying their differences.

\subsection{Model Overview}

Fig.~\ref{fig: main} presents a high-level overview of our \modelname{} method, with four main stages.
First, image features are extracted using two complementary types of encoders: SuperPoint \cite{detone2018superpoint}, focusing on generic fine-grained matching; and DINOv2 \cite{Oquab2023DINOv2}, an image foundation model which encodes coarse but broad visual knowledge.
Second, we build keypoint association graphs using these features, both intra and inter-image.
In contrast to previous work, our inter-image graph leverages DINOv2 guidance, which provides a coarse signal capturing general similarity between SuperPoint keypoints.
Third, we propagate information among the keypoints in both images based on the built graphs, using self and cross-attention layers for intra and inter-image communication, respectively.
Crucially, we disentangle positional and appearance signals at this stage, different from other models that overlook this aspect. 
This design enables feature refinement to be guided by both keypoint spatial arrangement and their feature similarities, but without contaminating the final descriptors with positional information, which hinders generalizability.
Finally, once the refined descriptors are obtained, optimal matching layers are applied to produce a mapping between the keypoints in the two images.
These stages are described in more detail in the following section.

\subsection{OmniGlue Details}

\noindent\textbf{Feature Extraction.} The inputs are two images with shared content, denoted as $I_A$ and $I_B$. 
We denote the SuperPoint keypoint sets of the two images as $\mathbf{A} := \{A_1, ..., A_N \}$ and $\mathbf{B} := \{ B_1, ..., B_M\}$. Note that $N$ and $M$ are the number of identified keypoints of $I_A$ and $I_B$, respectively. Each keypoint is associated with its SuperPoint local descriptor $\mathbf{d} \in \mathbb{R}^{C}$. 
Additionally, normalized keypoint locations are encoded with positional embeddings, and we further refine them using MLP layers. We denote the resulting positional features of a keypoint as $\mathbf{p} \in \mathbb{R}^{C}$.
Furthermore, we extract dense DINOv2 feature maps of the two images. We interpolate the feature maps using the location of SuperPoint keypoints to obtain DINOv2 descriptors for each keypoint, denoted as $\mathbf{g} \in \mathbb{R}^{C'}$.
For clarity, we denote the three features of the $i^{th}$ keypoint in set $\mathbf{A}$ as $\mathbf{d}^A_i$, $\mathbf{p}^A_i$ and $\mathbf{g}^A_i$. The features of the keypoints in set $\mathbf{B}$ are denoted accordingly. The goal of our \modelname{} model is to estimate correspondences between the two keypoint sets.

\noindent\textbf{Graph Building Leveraging DINOv2.} 
We build four keypoint association graphs: two inter-image graphs and two intra-image graphs. 
The two inter-image graphs represent the connectivity between the keypoints of the two images, from $I_A$ to $I_B$ and vice versa. 
We denote them as $\mathbf{G}_{A\xrightarrow{}B}$ and $\mathbf{G}_{B\xrightarrow{}A}$, respectively. The two inter-image graphs are directed, where information is propagated from the source node to the target node. 

\begin{figure}[t]
\begin{center}
   \includegraphics[width=\linewidth]{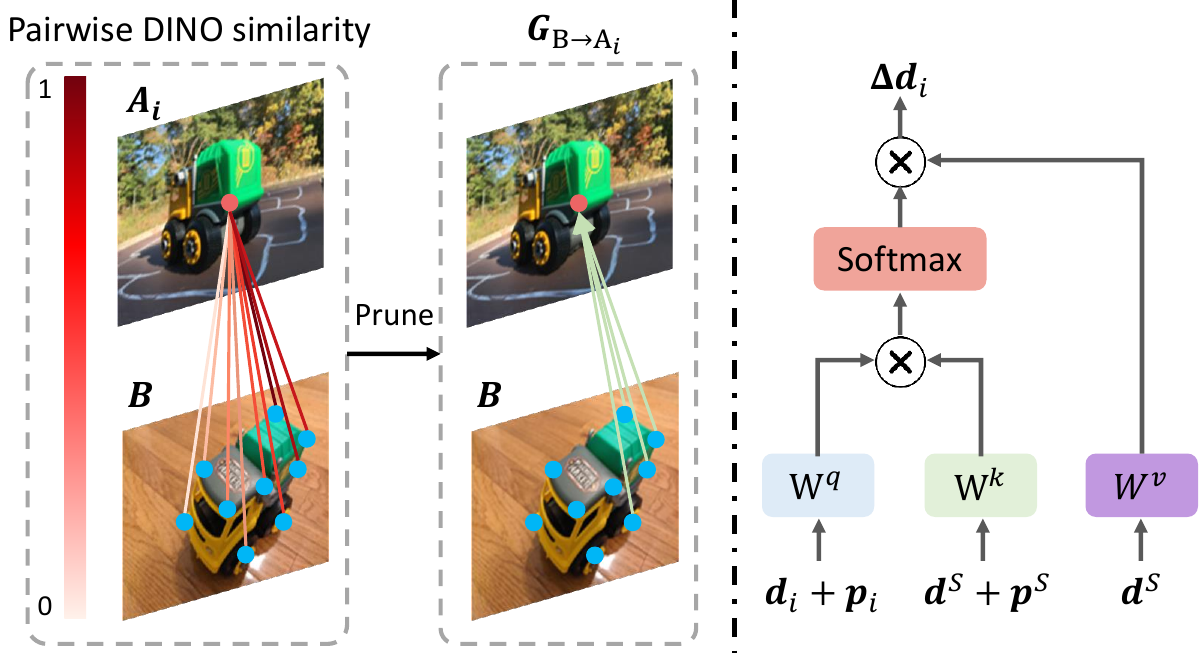}
   \vspace{-0.2in}
   \caption{
   \small{\textbf{(Left) Building inter-image graph.} We prune the dense pairwise graph based on the DINO feature similarity. \textbf{(Right) Position-guided attention.}  The keypoint position is involved in computing attention weights, while the output attention update is only composed of local descriptor components. 
   }
   }
  \vspace{-0.3in}
   \label{fig: graph}
\end{center}
\end{figure}

We leverage DINOv2 features to guide the building of the inter-image graphs. As depicted in Fig.~\ref{fig: graph} (left), we take $\mathbf{G}_{B\xrightarrow{}A_i}$ as an example. For each keypoint $A_i$ in keypoint set $\mathbf{A}$, we compute its DINOv2 feature similarities with all keypoints in set $\mathbf{B}$. Note that we perform channel-wise normalization on the DINOv2 features $\mathbf{g}^A_i$ and $\mathbf{g}^B$ before computing the similarities. 
We select the top half of keypoints in set $\mathbf{B}$ with the largest DINOv2 similarities to connect with $A_i$, which prunes the densely-connected pairwise graph between the keypoints of the two images.
We perform the same operation on all keypoints in $A$ to obtain $\mathbf{G}_{B\xrightarrow{}A}$, and the graph $\mathbf{G}_{A\xrightarrow{}B}$ is built in a similar manner. 

Similarly, the intra-image graphs represent the connectivity between keypoints belonging to the same image. We denote them as $\mathbf{G}_{A}$ and $\mathbf{G}_{B}$, which are undirected -- information is propagated bi-directionally between connected keypoints. Each keypoint is densely connected with all other keypoints within the same image.

\noindent\textbf{Information Propagation with Novel Guidance.} We perform information propagation based on the keypoint graphs. 
This module contains multiple blocks, where each block has two attention layers. The first one updates keypoints based on the intra-image graphs, performing self-attention; The second updates keypoints based on the inter-image graphs, performing cross-attention.
In particular, this stage introduces two novel elements compared to previous work, which we show are critical towards generalizable matching: suitable guidance from DINOv2 and from keypoint positions.

First, DINOv2 guidance:
during cross-attention, for keypoint $A_i$, it only aggregates information from the DINOv2-pruned potential matching set selected from $\mathbf{B}$, instead of all its keypoints.
This is particularly helpful for generalized image matching, where DINO's broad knowledge may guide the feature matching process in a domain that the model has not seen at training time.
In this manner, information from irrelevant keypoints will not be fused into the query keypoint features.
This process also encourages the cross-attention module to focus on distinguishing the matching point in the smaller potential matching set. 
Note, however, that we do not forcibly limit the matching space to the potential matching sets, as DINO may also be incorrect in some cases.

Second, we introduce refined keypoint guidance.
We observe that prior methods entangle keypoint positional features and local descriptors during feature propagation~\cite{sarlin2020superglue}, which makes the model overly dependent on learned position-related priors -- our ablation experiments in Section~\ref{sec:experiments} highlight this issue.
The learned priors are vulnerable under image pairs with matching patterns that were not seen at training time, limiting the generalization capability.
To deal with this issue, we propose a novel position-guided attention, which disentangles the keypoint positional features $\mathbf{p}$ and the local descriptors $\mathbf{d}$. 
The positional information is used as spatial context in this module and is not incorporated in the final local descriptor representation used for matching.

With these novel elements, our attention layer, illustrated in Fig.~\ref{fig: graph} (right), is defined as follows, where we take the example of keypoint $A_i$:
\begin{align}
    \label{eq: eq_1}
    \mathbf{d}^A_i \leftarrow \mathbf{d}^A_i + \text{MLP}([\mathbf{d}^A_i \vert \Delta \mathbf{d}^A_i]), \text{where} \\
    \label{eq: eq_2}
    \Delta \mathbf{d}^A_i = \text{Softmax}(\frac{\mathbf{q}^{A}_i (\mathbf{k}^S)^T}{\sqrt{C}}) \cdot \mathbf{v}^S \in \mathbb{R}^{C}, \text{and} \\
    \label{eq: eq_3}
    \mathbf{q}^{A}_i = \text{W}^q(\mathbf{d}^A_i + \mathbf{p}^A_i) + \mathbf{b}^q \in \mathbb{R}^{C},\\
    \label{eq: eq_4}
    \mathbf{k}^{S} = \text{W}^k(\mathbf{d}^S + \mathbf{p}^S)  + \mathbf{b}^k \in \mathbb{R}^{K\times C},\\
    \label{eq: eq_5}
    \mathbf{v}^{S} = \text{W}^v(\mathbf{d}^S) + \mathbf{b}^v\in \mathbb{R}^{K\times C}.
\end{align}
As described in Eq.~\ref{eq: eq_1}, the attention has a residual connection, which integrates the attention update value $\Delta \mathbf{d}^A_i$ . The notation $\leftarrow$ is the updating operation and $[\cdot|\cdot]$ is the channel-wise concatenation.
To compute the attention update value, as described in Eq.~\ref{eq: eq_2}, we compute the feature similarity between the keypoint $A_i$ and its source connected keypoints in a graph, which is denoted as $S$ containing $K$ keypoints. The query, key and value of the attention are $\mathbf{q}^{A}_i$, $\mathbf{k}^{S}$, and $\mathbf{v}^{S}$, respectively. Specifically, as shown in Eq.~\ref{eq: eq_3}-\ref{eq: eq_5}, the query and key are computed by fusing both local descriptors and positional features. The value, however, is transformed from only the local descriptors. We note that the weights ($\textbf{W}$) and bias ($\textbf{b}$), which map features into query, key and value tokens in attention, are not shared across different attention layers.
In self-attention ($\mathbf{G}_A$ and $\mathbf{G}_B$), $S$ is composed by all keypoints; in cross-attention ($\mathbf{G}_{A\xrightarrow{}B}$ and $\mathbf{G}_{B\xrightarrow{}A}$), $S$ contains the keypoints identified by DINO.

Intuitively, the query and key compute the attention weights, where both feature affinity and spatial correlations are considered. 
However, the attention update value, $\Delta \mathbf{d}^A_i$, is composed of local descriptor components only.
This design allows the model to reason about spatial correlation between keypoints using their positional features while avoiding an over-reliance on it.

\noindent\textbf{Matching Layer and Loss Function.}
We use the refined keypoint representations to produce a pairwise similarity matrix $\mathbf{S} \in \mathbb{R}^{N\times M}$, where $\mathbf{S}_{i,j} = \mathbf{d}_i^A \cdot (\mathbf{d}_j^B)^T$.
Then we use the Sinkhorn algorithm~\cite{sinkhorn1967concerning} to refine the similarities, which produces the matching matrix $\mathbf{M} \in [0,1]^{N\times M}$, where $\mathbf{M}_{i,j}$ represents the matching probability between keypoint $A_i$ and $B_j$. To train \modelname{}, we minimize
the negative log-likelihood of the matching matrix with ground truth~\cite{sarlin2020superglue, Sun2021LoFTR}.

\vspace{-1mm}
\subsection{Comparison Against SuperGlue and LightGlue}

It is important to highlight differences between our model and reference sparse learnable feature matching methods, SuperGlue \cite{sarlin2020superglue} and LightGlue \cite{Lindenberger2023LightGlue}.
While neither of these is designed to target generalizability to multiple domains, there are common elements in the model structure, so we would like to emphasize our novelty.

Both works use attention layers for information propagation.
Differently, \modelname{} leverages a foundation model to guide this process, which significantly helps with transferring to image domains that are not observed during training.

In terms of local descriptor refinement, \modelname{} departs from SuperGlue to disentangle positional and appearance features. For reference, SuperGlue represents keypoint with entangling the two features as $\mathbf{d}+\mathbf{p}$, where positional features are also used to produce matching results.
Similar to our design, LightGlue removes the dependency of the updated descriptors on the positional features. However, it proposes a very specific positional encoding formulation, based on rotary encodings, only in self-attention layers.

Overall, SuperGlue is the closest model to \modelname{}, serving as a directly comparable reference where our contributions can be clearly ablated.
For this reason, in the following section, we use SuperGlue as the main reference comparison for experimental validation.

\begin{table}[t]
\centering
\renewcommand{\arraystretch}{1.25}
\resizebox{\linewidth}{!}{
\begin{tabular}{ccccccccc}\hline
    & (1) & (2) & (3) & (4) & (5) & (6) & (7) & (8)\\
    & \multirow{2}{*}{\textbf{Type}} & \multirow{2}{*}{\textbf{Scene}} & \textbf{Real} & \textbf{Syn.} & \multirow{2}{*}{\textbf{Mask}} & \textbf{Cam.} & \textbf{Diff.} & \multirow{2}{*}{\textbf{Task}} \\
    &  &  & \textbf{Img.} & \textbf{Trans.} & & \textbf{Bl.} & \textbf{Bg.} & \\\shline
\textit{MegaDepth} & Scene & Outdoor & \checkmark & \xmark & \xmark & Large & \xmark & Corr. \& Pose Est.\\
\textit{GSO-Hard} & Object & None & \xmark & \xmark & \xmark & Large & \xmark & Pose Est.\\
\textit{GSO-Easy} & Object & None & \xmark & \xmark & \xmark & Small & \xmark & Pose Est.\\
\textit{NAVI-MV} & Object & In \& Outdoor & \checkmark & \xmark & \checkmark & Large & \xmark & Pose Est.\\
\textit{NAVI-Wild} & Object & In \& Outdoor & \checkmark & \xmark & \checkmark & Large & \checkmark & Pose Est.\\
\textit{ScanNet} & Scene & Indoor & \checkmark & \xmark & \xmark & Large & \xmark & Pose Est.\\
\textit{SH} & Scene & Outdoor & \checkmark & \checkmark & \xmark & Small & \xmark & Corr. Est.\\
\textit{DeepAerial} & Scene & Aerial & \checkmark & \checkmark & \xmark & N/A & \checkmark & Image Reg. \\
\hline
\end{tabular}
}
\vspace{-0.1in}
\caption{\small{Dataset and task comparisons on: (1) The general type; (2) The background scene type; (3) Use of real (\checkmark) or rendered (\xmark) images; (4) Whether the pose transformation is synthetic; (5) Whether foreground masks are used to filter correspondence predictions; (6) The camera baseline type; (7) Whether two input images have different backgrounds; (8) Evaluated tasks: Correspondence Estimation, Pose Estimation or Image Registration. 
}}
\vspace{-0.1in}
\label{tabel: data}
\end{table}

\begin{figure*}[t]
\begin{center}
   \includegraphics[width=\linewidth]{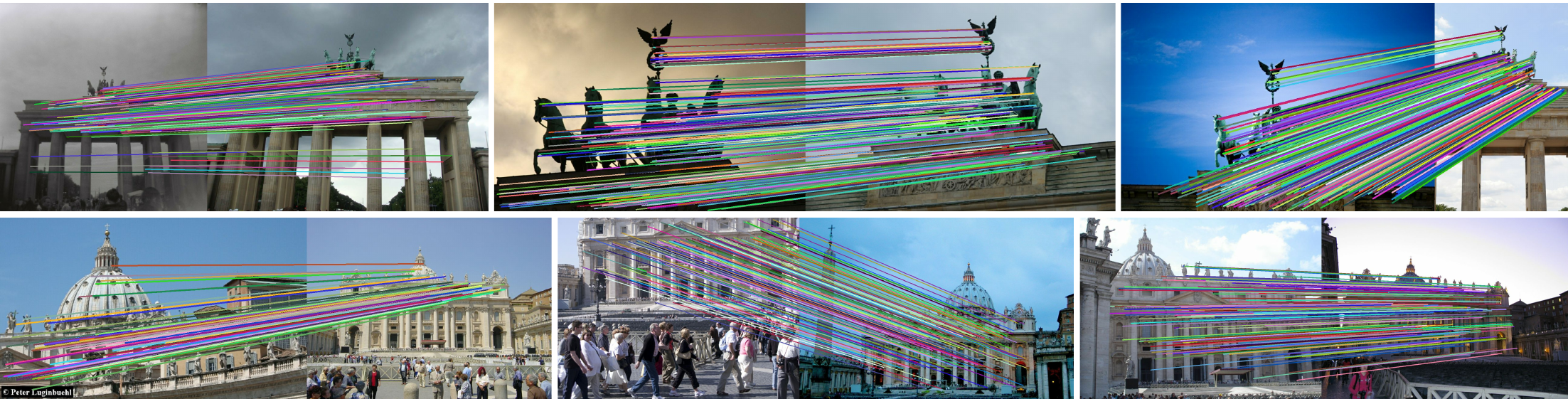}
   \vspace{-0.25in}
   \caption{
   Visualization of correspondences predicted by \modelname{} on the MegaDepth-1500 benchmark. We distinguish the matches by different colors. We show results for scene "0022" and "0015" on the top and bottom rows, respectively.
   }
   \vspace{-0.25in}
   \label{fig: md1500}
\end{center}
\end{figure*}

\vspace{-2mm}
\section{Experiments}
\vspace{-2mm}
\label{sec:experiments}

We first introduce the experiment setup and then present our results as well as ablation studies.

\subsection{Experimental Setup}
\vspace{-1mm}
We list the datasets and tasks used for evaluating \modelname{} in Table~\ref{tabel: data}. We include details of \textbf{datasets} as follows:

\begin{itemize}

    \item \textbf{Synthetic Homography (SH)} contains images from the Oxford and Paris dataset~\cite{radenovic2018revisiting}. We generate random crops and homography transformations to sample image patch pairs, similar to~\cite{sarlin2020superglue}. Two subsets are generated, SH100 and SH200, wherein the perturbations of the image corners for homography generation are within 100 and 200 pixels, respectively. For each subset, we generate roughly 9 million training pairs and 10K test pairs.
    
    \item \textbf{MegaDepth (MD)}~\cite{Li2018MegaDepth} is a large-scale outdoor image dataset. The ground-truth matches are computed using SfM~\cite{schonberger2016colmap}. We follow the train/test split of prior works~\cite{Sun2021LoFTR}, with roughly 625K training pairs and 1500 test pairs.

    \item \textbf{Google Scanned Objects (GSO)}~\cite{Downs2022GoogleSO} comprises $1400$ daily object model scans of 17 categories. We render synthetic images with large (60\degree - 90\degree) rotation (Hard subset) and small (15\degree - 45\degree) rotation (Easy subset) camera baselines, intentionally distinct from the training distribution. We produce $50$ image pairs for each object model, resulting in around $140$K test cases.

    \item \textbf{NAVI}~\cite{jampani2023navi} focuses on objects and encompasses a variety of both indoor and outdoor images. It is divided into two subsets: the multiview subset ($25$K image pairs), featuring input images captured in the same environment; and the wild subset ($36$K image pairs), where the two input images are taken in different environments with distinct backgrounds, lighting conditions and camera models.
    
    \item \textbf{ScanNet}~\cite{dai2017scannet} collects indoor images. We follow the split of prior works~\cite{Sun2021LoFTR} with 1500 evaluation pairs.

    \item \textbf{DeepAerialMatching}~\cite{park2020two} provides aligned pairs of satellite images under varying conditions (i.e. different seasons, weather, time-of-day). We introduce random 2D rotations and crop $520\times520$ image patches to produce image pairs with known affine transformations (500 in total).
\end{itemize}

\noindent\textbf{Tasks and metrics.} We assess the models across three tasks: (1) \textit{Correspondence estimation}, evaluated with correspondence-level precision and recall (for sparse methods only). Following SuperGlue~\cite{sarlin2020superglue}, we employ thresholds of $<3px$ and $>5px$ to label a correspondence as correct and incorrect, respectively. 
(2) \textit{Camera pose estimation}, evaluated with pose accuracy (\% of correct poses within $\{5^{\circ}, 10^{\circ}, 20^{\circ}\}$ of error) and AUC, with accuracy being used by default unless otherwise specified. The poses are derived from the estimated correspondences using RANSAC~\cite{Fischler1981RandomSC}, and we use Rodrigues' formula to calculate relative rotation error between the predicted/ground truth rotation matrices; 
(3) \textit{Aerial image registration}, evaluated with percentage of correct keypoints (PCK). We use RANSAC-based affine estimation from the estimated correspondences, and apply the predicted/ground truth affine transformations to 20 test keypoints with fixed positions to calculate the PCK within $\tau \cdot max(h, w)$ pixels of error, for $\tau \in \{0.01, 0.03, 0.05\}$.

\noindent\textbf{Baselines.}
We compare OmniGlue against:
\begin{itemize}
    \item \textbf{SIFT}~\cite{Lowe2004sift} and \textbf{SuperPoint}~\cite{detone2018superpoint} provide domain-agnostic local visual descriptors for keypoints. We generate matching results using both nearest neighbor + ratio test (NN/ratio) and mutual nearest neighbor (MNN), with the best outcomes being reported. 

    \item \textbf{Sparse matchers:} \textbf{SuperGlue}~\cite{sarlin2020superglue} employs attention layers for intra- and inter-image keypoint information aggregation, using descriptors derived from SuperPoint~\cite{detone2018superpoint}. It is the closest reference of \modelname{}. \textbf{LightGlue}~\cite{Lindenberger2023LightGlue} improves SuperGlue~\cite{sarlin2020superglue} with better performance and speed. Besides, we also test with \textbf{DINOv2}~\cite{Oquab2023DINOv2}+SuperGlue, by substituting SuperPoint descriptors with DINO features.

    \item \textbf{(Semi-)Dense matchers:} \textbf{LoFTR}~\cite{Sun2021LoFTR} and \textbf{PDCNet}~\cite{Truong2021pdcnet} are used as reference dense matching techniques, to contextualize our sparse matching performance with respect to other types of approaches.

\end{itemize}

\noindent\textbf{Implementation details.}
In line with SuperGlue~\cite{sarlin2020superglue}, we implement $9$ contextual reasoning blocks, each comprising an intra-image aggregation layer (self-attention) and an inter-image aggregation layer (cross-attention). This configuration results in a total of $18$ attentional layers. Across all sparse methods, we use $1024$ keypoints and $256$-dimensional descriptors. See more training details in supplementary.

\begin{table}[t]
\centering
\renewcommand{\arraystretch}{1.25}
\resizebox{0.85\linewidth}{!}{
\begin{tabular}{c|c|c}\shline
    Setting $\rightarrow$ & \multicolumn{2}{c}{\textbf{Test Performance (in-domain)}} \\ \hline
    & \multicolumn{1}{c|}{\textit{SH100}} & \multicolumn{1}{c}{\textit{SH200}} \\ \hline
    DINOv2~\cite{Oquab2023DINOv2}+SG~\cite{sarlin2020superglue} & 87.6 / 88.4 & 79.8 / 80.2 \\
    SP\cite{detone2018superpoint}+SG~\cite{sarlin2020superglue} & \textbf{99.2} / 99.4 & 95.4 / 96.0 \\ \hline
    \textbf{OmniGlue (ours)} & \tablefirst \textbf{99.2} / \textbf{99.5} & \tablefirst \textbf{96.4} / \textbf{98.0} \\ \hline \hline
    Setting $\rightarrow$ & \multicolumn{2}{c}{\textbf{Test Generalization} (\textit{src $\rightarrow$ trg})} \\ \hline
    & \multicolumn{1}{c|}{\textit{SH100 $\rightarrow$ SH200}} & \multicolumn{1}{c}{\textit{SH200 $\rightarrow$ MD}} \\ \hline
    DINOv2~\cite{Oquab2023DINOv2}+SG~\cite{sarlin2020superglue} & 72.6 / 77.3 & 19.2 / 18.8 \\
    SP\cite{detone2018superpoint}+SG~\cite{sarlin2020superglue} & 78.3 / 75.6 & 34.9 / 39.0 \\ \hline
    \textbf{OmniGlue (ours)} & \tablefirst \textbf{90.0} / \textbf{89.6} & \tablefirst \textbf{36.0} / \textbf{54.7} \\ 
    \small \textbf{\green{relative gain (\%)}} & \small \textbf{\green{+14.9}} / \textbf{\green{18.5}} & \small \textbf{\green{+4.3}} / \textbf{\green{+40.3}}\\ \shline

\end{tabular}
}
\vspace{-0.1in}
\caption{\small{Results for in-domain (top) and zero-shot generalization to out-of-domain datasets (bottom), for models trained on Synthetic Homography (SH) datasets. We measure precision / recall at the correspondence level.}}
\vspace{-0.2in}
\label{tabel: SH_exp}
\end{table}

\begin{figure*}[t]
\begin{center}
   \includegraphics[width=\linewidth]{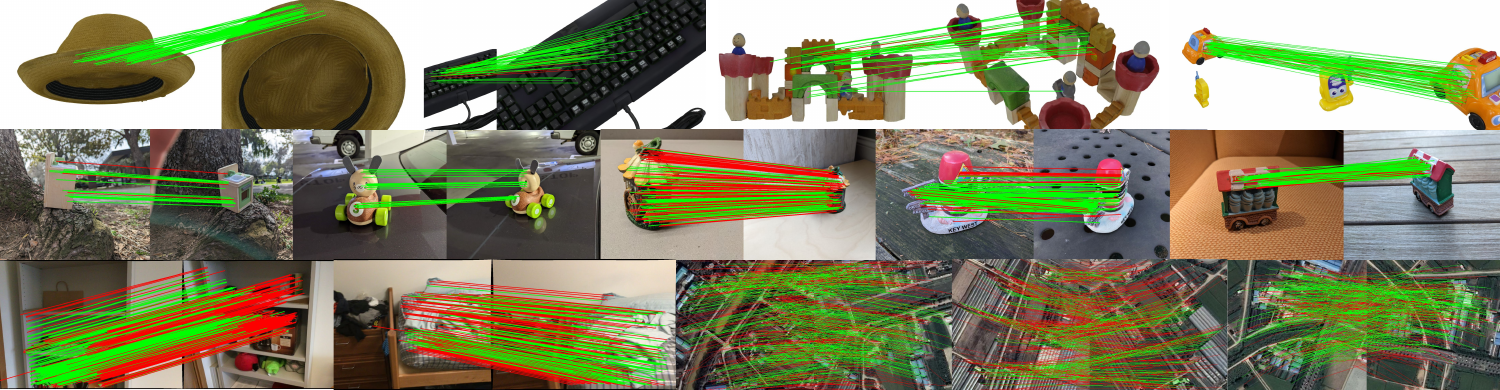}
   \vspace{-0.25in}
   \caption{
   Zero-shot generalization to novel domains. The top and middle row show results on GSO and NAVI, the last row shows results on ScanNet and DeepAerial. We draw the correct and incorrect estimated correspondences as green and red, respectively.
   }
   \vspace{-0.2in}
   \label{fig: ood-viz}
\end{center}
\end{figure*}

\begin{table*}[t]
\centering
\renewcommand{\arraystretch}{1.25}
\resizebox{\linewidth}{!}{
\begin{tabular}{c|c|c|c|c|c|c|c}\shline
    & \multicolumn{1}{c|}{\textbf{In-domain}} & \multicolumn{6}{c}{\textbf{Out-of-domain} (Zero-shot Generalization)} \\
     
    & \multirow{2}{*}{\textit{MegaDepth-1500}} & \multicolumn{2}{c|}{\textit{Google Scanned Object}} & \multicolumn{2}{c|}{\textit{NAVI}} & \multirow{2}{*}{\textit{ScanNet}} &
    \multirow{2}{*}{\textit{DeepAerial}} \\
    &  & \multicolumn{1}{c|}{Hard (60-90 deg.)} & \multicolumn{1}{c|}{Easy (15-45 deg.)} & \multicolumn{1}{c|}{Multiview} & \multicolumn{1}{c|}{Wild} &  \\ 
    Method & \multicolumn{1}{c|}{AUC@5\degree / 10\degree / 20\degree } & \multicolumn{1}{c|}{Acc@5\degree / 10\degree / 20\degree} & \multicolumn{1}{c|}{Acc@5\degree / 10\degree / 20\degree} & \multicolumn{1}{c|}{Acc@5\degree / 10\degree / 20\degree} & \multicolumn{1}{c|}{Acc@5\degree / 10\degree / 20\degree} & \multicolumn{1}{c|}{Acc@5\degree / 10\degree / 20\degree} & 
    \multicolumn{1}{c}{PCK@$1\%/3\%/5\%$} \\ \shline
    \multicolumn{7}{l}{\textsc{Dense and Semi-Dense methods}} \\ \hline
    PDCNet~\cite{Truong2021pdcnet} & 51.5 / 67.5 / 78.2 & 5.1 / 8.9 / 14.9 & 24.8 / 36.7 / 49.3 & 3.9 / 7.1 / 11.6 & 6.6 / 11.6 / 17.0 & \tablefirst \textbf{38.6} / \textbf{60.0} / 71.3 & 14.0 / 20.9 / 22.6 \\
    LoFTR~\cite{Sun2021LoFTR} & \textbf{52.8} / \textbf{69.2} / \textbf{81.2} & 7.6 / 14.0 / 22.9 & 38.2 / 54.1 / 67.5 & 12.5 / 22.7 / 34.2 & 9.8 / 18.4 / 29.8 & 36.2 / 56.1 / 68.6 & 17.8 / 23.7 / 25.0 \\ \hline \hline
    \multicolumn{7}{l}{\textsc{Descriptor+Hand-crafted rules}} \\ \hline
    SIFT~\cite{Lowe2004sift}+MNN & 25.8 / 41.5 / 54.2 & 6.8 / 12.1 / 20.3 & 32.5 / 46.2 / 60.3 & 6.2 / 11.9 / 22.7 & 4.2 / 8.1 / 23.1 &  4.6 / 10.6 / 20.2 & 17.5 / 25.9 / 32.2\\
    SuperPoint~\cite{detone2018superpoint}+MNN & \textbf{31.7} / \textbf{46.8} / \textbf{60.1} & 5.4 / 10.5 / 18.8 & 28.9 / 43.4 / 58.0 & 10.0 / 19.2 / 31.6 & 8.2 / 16.0 / 28.0 & 18.8 / 35.2 / 49.6 & 16.0 / 24.3 / 31.9\\ \hline \hline
    \multicolumn{7}{l}{\textsc{Sparse methods}} \\ \hline
    DINOv2~\cite{Oquab2023DINOv2}+SG~\cite{sarlin2020superglue} & 31.5 / 40.8 / 45.3 & 3.6 / 7.3 / 15.1 & 12.0 / 22.7 / 38.7 & 7.3 / 15.6 / 28.3 & 8.4 / 17.2 / 30.6 & 9.7 / 26.7 / 41.5 & 11.4 / 18.2 / 23.1\\
    SuperGlue~\cite{sarlin2020superglue} & 42.2 / 61.2 / 76.0 & 7.2 / 13.2 / 21.6 & 32.3 / 48.4 / 62.9 & 11.8 / 21.9 / 34.4 & 10.6 / 19.8 / 31.8 & 25.5 / 43.4 / 57.3 & 16.4 / 26.2 / 28.8\\
    LightGlue~\cite{Lindenberger2023LightGlue} & \textbf{47.6} / 64.8 / \textbf{77.9} & 7.5 / 13.8 / 21.7 & 36.4 / 53.2 / 66.9 & \textbf{13.2} / 24.0 / 34.8 & 9.7 / 17.6 / 25.9 & 36.7 / 59.4 / \textbf{71.6} & 18.1 / 25.8 / 27.3 \\ 
    \multirow{1}{*}{\textbf{OmniGlue (ours)}} & 47.4 / \textbf{65.0} / 77.8
    & \tablefirst \textbf{8.6} / \textbf{15.3} / \textbf{25.0} & \tablefirst \textbf{38.4} / \textbf{54.8} / \textbf{68.8} & \tablefirst \textbf{13.2} / \textbf{24.8} / \textbf{37.7} & \tablefirst \textbf{12.4} / \textbf{22.8} / \textbf{35.0} & 31.3 / 50.2 / 65.0 & \tablefirst \textbf{22.4} / \textbf{33.5} / \textbf{36.6}\\  
    \small \textbf{\green{rel. gain (\%) over~}}\cite{sarlin2020superglue} & \small \textbf{\green{+12.3}} / \textbf{\green{+6.2}} / \textbf{\green{+2.4}} & \small \textbf{\green{+19.4}} / \textbf{\green{+15.9}} / \textbf{\green{+15.7}} & \small \textbf{\green{+18.9}} / \textbf{\green{+13.2}} / \textbf{\green{+9.4}} & \small \textbf{\green{+11.9}} / \textbf{\green{+13.4}} / \textbf{\green{+9.6}} & \small \textbf{\green{+16.7}} / \textbf{\green{+15.2}} / \textbf{\green{+10.1}} & \small \textbf{\green{+22.0}} / \textbf{\green{+15.7}} / \textbf{\green{+13.4}} & \small \textbf{\green{+36.6}} / \textbf{\green{+27.9}} / \textbf{\green{+27.0}}\\\shline
    
\end{tabular}
}
\vspace{-0.15in}
\caption{Results for in-domain (left, measured with AUC) and zero-shot generalization to out-of-domain datasets (right, measured with pose accuracy / PCK), for models trained on the MegaDepth dataset. We highlight the best results on out-of-domain data and show our relative improvement against our base method SuperGlue. All sparse methods use 1024 keypoints.}
\label{tabel: main}
\vspace{-0.15in}
\end{table*}

\vspace{-0.5mm}
\subsection{Results}
\vspace{-0.5mm}
Following SuperGlue and LightGlue, we first initialize \modelname{} by training it on SH100. Then we further pre-train \modelname{} on SH200, and finally train \modelname{} on MegaDepth (MD). We evaluate \modelname{} and all baseline methods on the test splits of each training domain, and test their generalization to both subsequent training datasets or out-of-domain test datasets. Finally, we experiment with adapting \modelname{} to out-of-domain images with limited target domain training data.

\noindent\textbf{From Synthetic Homography to MegaDepth.} 
As depicted in Table~\ref{tabel: SH_exp}, in comparison to the base method SuperGlue, \modelname{} not only exhibits superior performance on the in-domain data but also demonstrates robust generalization. Even with a minimal data distribution shift from SH100 to SH200, SuperGlue experiences substantial drops in performance with a $20\%$ reduction in precision and recall. This result implies that SuperGlue is overly dependent on learned position-related patterns and is unable to handle further image warping distortion. In contrast, \modelname{} showcases strong generalization capability, surpassing SuperGlue with a $12\%$ improvement in precision and a  $14\%$ boost in recall. Similarly, during the transfer from SH200 to Megadepth, \modelname{} outperforms SuperGlue with a drastic $15\%$ improvement in recall. 

\noindent\textbf{From MegaDepth to other Domains.}
As shown in Table~\ref{tabel: main}, \modelname{} not only achieves comparable performance on MegaDepth-1500 with the state-of-the-art sparse matcher LightGlue, but also demonstrates better generalization capability on 5 out of 6 novel domains, when compared to all other methods. In detail, on MegaDepth-1500, \modelname{} showcases $12.3\%$ relative gain (pose AUC $@5\degree$) over the base method SuperGlue. On the 6 novel domains, \modelname{} shows $20.9\%$ and $9.5\%$ averaged relative gains (for pose and registration accuracy at the tightest thresholds) over SuperGlue and LightGlue, respectively. Moreover, \modelname{} demonstrates larger performance gains on harder novel domains against LightGlue, i.e., on GSO-Hard, NAVI-Wild, and DeepAerial. We show visualization in Fig.~\ref{fig: ood-viz} and Fig~\ref{fig: md1500} for zero-shot generalization on novel domains and its performance on the source domain.

Notably, the reference dense matchers, which achieve better performance on the in-domain MegaDepth dataset, generalize worse.
Their performances are close, or even worse, to SuperGlue, which has $10\%$ lower in-domain AUC@$5\degree$. 
We conjecture this may be due to the joint learning of visual descriptors and the matching module, making them easier to specialize strongly to the training domain.

\noindent\textbf{Low-Shot Fine-tuning on Target Domain.}
In certain real-world scenarios, a limited set of target domain data may be available for fine-tuning. To test this scenario, we fine-tune \modelname{} on the target domain (object-centric GSO dataset), comparing its performance with the base model, SuperGlue. We create small training subsets by utilizing only a few dozen object scans. 
Notably, these small training sets consist of instances from the sneaker object category only, covering a significantly minor subset of the testing object category distribution.

As depicted in Table~\ref{tabel: finetune}, \modelname{} is more readily adapted to the target domain. In detail, when scaling from $0$ to $30$ instances for training, \modelname{} consistently exhibits enhanced performance for both test subsets. With just $10$ instances for training, \modelname{} improves pose estimation accuracy by $5.3\%$ and $4.0\%$ on the two subsets. Expanding the training sets by incorporating $10$ more objects leads to a further performance improvement of $2\%$.
Furthermore, \modelname{} consistently surpasses SuperGlue, achieving a relative gain of approximately $10\%$ across all experiments. The results collectively demonstrate the applicability of \modelname{} in real-world scenarios as a versatile and generalizable method.

\begin{table}[t]
\centering
\renewcommand{\arraystretch}{1.25}
\resizebox{0.84\linewidth}{!}{
\begin{tabular}{cc|c|c}\shline
    \#Train & \multirow{2}{*}{Model} & \multicolumn{1}{c|}{Hard (60-90 deg.)} & \multicolumn{1}{c}{Easy (15-45 deg.)}\\
    Inst. &  & \multicolumn{1}{c|}{@5\degree / 10\degree / 20\degree} & \multicolumn{1}{c}{@5\degree / 10\degree / 20\degree}\\ \shline
    \multirow{2}{*}{0} & SG & 7.2 / 13.2 / 21.6 & 32.3 / 48.4 / 62.9\\
    & OG & \tablefirst \textbf{8.6} / \textbf{15.3} / \textbf{25.0}  & \tablefirst \textbf{38.4} / \textbf{54.8} / \textbf{68.8}\\ \hline
    \multirow{3}{*}{10} & SG & 11.6 / 20.8 / 31.7 & 38.9 / 55.7 / 68.6\\
    & OG & \tablefirst \textbf{13.9} / \textbf{24.6} / \textbf{36.8}  & \tablefirst \textbf{42.4} / \textbf{60.1} / \textbf{74.0}\\ 
    & \small \textbf{\green{rel. gain ($\%$)}} & \small \textbf{\green{+61.6}} / \textbf{\green{+60.8}} / \textbf{\green{+47.2}} & \small \textbf{\green{+10.4}} / \textbf{\green{+9.7}} / \textbf{\green{+7.6}} \\\hline
    \multirow{3}{*}{20} & SG & 13.0 / 22.9 / 35.2 & 40.3 / 57.0 / 70.5\\
    & OG & \tablefirst \textbf{15.3} / \textbf{27.0} / \textbf{39.7} & \tablefirst \textbf{44.1} / \textbf{61.5} / \textbf{75.0}\\ 
    & \small \textbf{\green{rel. gain ($\%$)}} & \small \textbf{\green{+77.9}} / \textbf{\green{+76.5}} / \textbf{\green{+58.8}} & \small \textbf{\green{+14.8}} / \textbf{\green{+12.2}} / \textbf{\green{+9.0}} \\ \hline
     \multirow{3}{*}{30} & SG & 14.6 / 25.2 / 37.9 & 42.0 / 59.2 / 71.2\\
    & OG & \tablefirst \textbf{16.7} / \textbf{29.1} / \textbf{42.3} & \tablefirst \textbf{45.8} / \textbf{62.5} / \textbf{76.0}\\
    & \small \textbf{\green{rel. gain ($\%$)}} & \small \textbf{\green{+94.2}} / \textbf{\green{+90.2}} / \textbf{\green{+69.2}} & \small \textbf{\green{+19.3}} / \textbf{\green{+14.1}} / \textbf{\green{+10.5}} \\ \shline
\end{tabular}
}
\vspace{-0.1in}
\caption{Fine-tuning results of SuperGlue~\cite{sarlin2020superglue} (SG) and our method OmniGlue (OG) on Google Scanned Object (GSO) dataset. We use dozens of sneaker object instances to generate training data and test on all 17 GSO categories. We also show a relative gain compared with the zero-shot performance.}
\label{tabel: finetune}
\vspace{-0.15in}
\end{table}

\vspace{-0.8mm}
\subsection{Ablation Study and Insights}
\vspace{-0.5mm}


We conduct a comprehensive ablation study on each proposed module, as detailed in Table~\ref{tabel: ablation}. Please note that the numbers reported on the GSO dataset are based on a subset, encompassing half of all test cases, for rapid evaluation. 

\noindent\textbf{The effectiveness of each proposed technique.}
The results in Table~\ref{tabel: ablation} (1) highlight the effectiveness of our foundation model guidance, which enhances the generalization capability on out-of-domain data.  Additionally, the third row of Table~\ref{tabel: ablation} (2) illustrates the impact of the position-guided attention, showcasing improvement in both in-domain and out-of-domain data. Furthermore, we conduct ablations with different approaches to disentangling keypoint positional features. The first two rows of Table~\ref{tabel: ablation} (2) demonstrate that performance degrades when either not using any positional features or applying the position-guidance only on self-attention (without positional guidance on cross-attention). This emphasizes the effectiveness of our position-guided attention in facilitating information propagation within both intra- and inter-image contexts. Besides, after removing the positional embeddings, the model shows better generalization even though the in-domain performance drops. This result implies that the inappropriate way that SuperGlue uses positional information limits its generalization.

\noindent\textbf{The ways of incorporating DINO features.}
As shown in Table~\ref{tabel: ablation} (3), we explore different methods of incorporating DINOv2. 
The first involves merging DINO features and SuperPoint local descriptors. This integration is performed before the information propagation module using an MLP. The experiment reveals a decline in performance, suggesting that the two features are not compatible, likely due to the coarse granularity of DINO. The manner in which these features can be effectively merged remains an open problem.


The second method entails applying DINOv2 guidance for constructing both intra and inter-image graphs, demonstrating diminished performance compared to (5). We hypothesize that the reason lies in the fact that intra-image information propagation (self-attention) requires a global context, particularly for distinguishing all keypoints in the feature space. Reducing connectivity on the intra-image graph adversely affects the global context, aligning with findings in the study of attention span in SuperGlue.

\noindent\textbf{Details of foundation model guidance.} We ablate the hyperparameter used to determine the number of source keypoint in a graph, as presented in Table~\ref{tabel: ablation} (4). The results indicate that selecting the top half of keypoints in the other image for building inter-image graphs is the optimal choice.

\begin{table}[t]
\centering
\renewcommand{\arraystretch}{1.3}
\resizebox{\linewidth}{!}{
\begin{tabular}{cc|c|c|c}\shline
    & & \multicolumn{1}{c|}{\textbf{In-domain}} & \multicolumn{2}{c}{\textbf{Out-of-domain}} \\
     
    & & \multirow{1}{*}{\textit{MegaDepth}} & \multicolumn{2}{c}{\textit{Google Scanned Object}}\\
    & &  & \multicolumn{1}{c|}{Hard} & \multicolumn{1}{c}{Easy} \\ 
    & & P / R & \multicolumn{1}{c|}{@5\degree / 10\degree / 20\degree} & \multicolumn{1}{c}{@5\degree / 10\degree / 20\degree} \\ \shline
    (0) & SuperGlue~\cite{sarlin2020superglue} & 67.2 / 68.3 & 9.0 / 16.9 / 27.3 & 40.4 / 60.5 / 76.6\\ \hline
    (1) & only DINO-guide & 66.6 / 68.0 & 10.0 / 18.7 / 29.6 & 46.2 / 65.4 / 79.5\\ \hline
    \multirow{3}{*}{(2)} & only no pos. emb. - all & 60.5 / 58.1 & 9.1 / 17.2 / 27.7 & 43.5 / 63.2 / 78.2\\ 
    & only no pos. emb. - cross & 63.3 / 62.1 & 9.3 / 17.0 / 28.0 & 44.8 / 64.1 / 79.4 \\
    & only pos. guidance & \textbf{69.2} / 73.9 & 9.8 / 18.0 / 28.6 & 46.4 / 66.6 / 80.2\\\hline
    \multirow{2}{*}{(3)}  & (2) + DINO-SP-merge & 62.6 / 65.6 & 7.8 / 14.9 / 24.9 & 42.5 / 61.3 / 75.4\\
    & (2) + DINO-guide-intra+inter & 66.4 / 72.2 & 10.5 / 19.4 / 30.5 & 47.1 / 66.8 / 80.8\\ \hline
    \multirow{3}{*}{(4)} & (2) + DINO-guide-0.3 & 66.8 / 73.3 & 10.3 / 19.3 / 30.8 & 47.3 / 67.1 / 81.0\\ 
    & (2) + DINO-guide-0.4 & 66.8 / 73.1 & 10.2 / 18.9 / 30.4 & 47.2 / 66.9 / 80.8 \\
    & (2) + DINO-guide-0.6 & 66.7 / 74.1 & 10.2 / 19.1 / 30.3 & 47.7 / 67.4 / 81.1\\ \hline
    (5) & (2) + DINO-guide-0.5 (full) & 66.2 / 74.1 & \textbf{11.0} / \textbf{20.4} / \textbf{32.0} & \textbf{48.7} / \textbf{68.4} / \textbf{82.3}\\ \shline
\end{tabular}
}
\vspace{-0.1in}
\caption{\small{Ablation study on (1) only with DINO guidance, (2) only with the disentangled keypoint representation variants, (3) DINO guidance variants analysis (based on (2) with position guidance), (4) DINO guidance threshold analysis, and (5) full model \modelname{}.}}
\label{tabel: ablation}
\vspace{-0.15in}
\end{table}

\section{Conclusions and Future Work}
We propose \modelname{}, the first learnable image matcher that is designed with generalization as a core principle. We introduce the broad visual knowledge of a foundation model, which guides the graph-building process. We identify the limitation of the previous descriptor-position entangled representation and present a novel attention module to deal with it. We demonstrate that \modelname{} outperforms prior work with better cross-domain generalization. Moreover, \modelname{} can also be easily adapted to a target domain with a limited amount of data collected for fine-tuning.
For future work, it is also worth exploring how to leverage unannotated data in target domains to improve generalization. Both of better architectural designs and better data strategies can pave the way for a foundational matching model. 

{\scriptsize{
\noindent\textbf{Acknowledgements.} We would like to acknowledge support from NSF IIS-2047677, HDR-1934932, CCF-2019844, and the IARPA WRIVA program.}
}



\appendix
\section*{Appendix}

\section{Additional Model Details}

\modelname{} undergoes training with $750,000$ iterations using a batch size of $48$ on $8$ NVIDIA Tesla V100 GPUs. 
The initial learning rate is set at $3e-5$, with a decay rate of $0.999991$ and a hinge step of $55000$. 
For DINOv2~\cite{Oquab2023DINOv2} feature extraction, we use the images with a maximum resolution (long side) of $630$, maintaining the aspect ratio during image resizing, for reduce the computation. 
The DINOv2 backbone employed ViT-14-base~\cite{dosovitskiy2020image}. We use the improved positional embedding scheme proposed in LFM-3D~\cite{karpur2023lfm3d}. 

\section{Target Domain Visualization}
\begin{figure*}[t]
    \centering

    \begin{subfigure}{0.155\textwidth}
        \includegraphics[width=\linewidth]{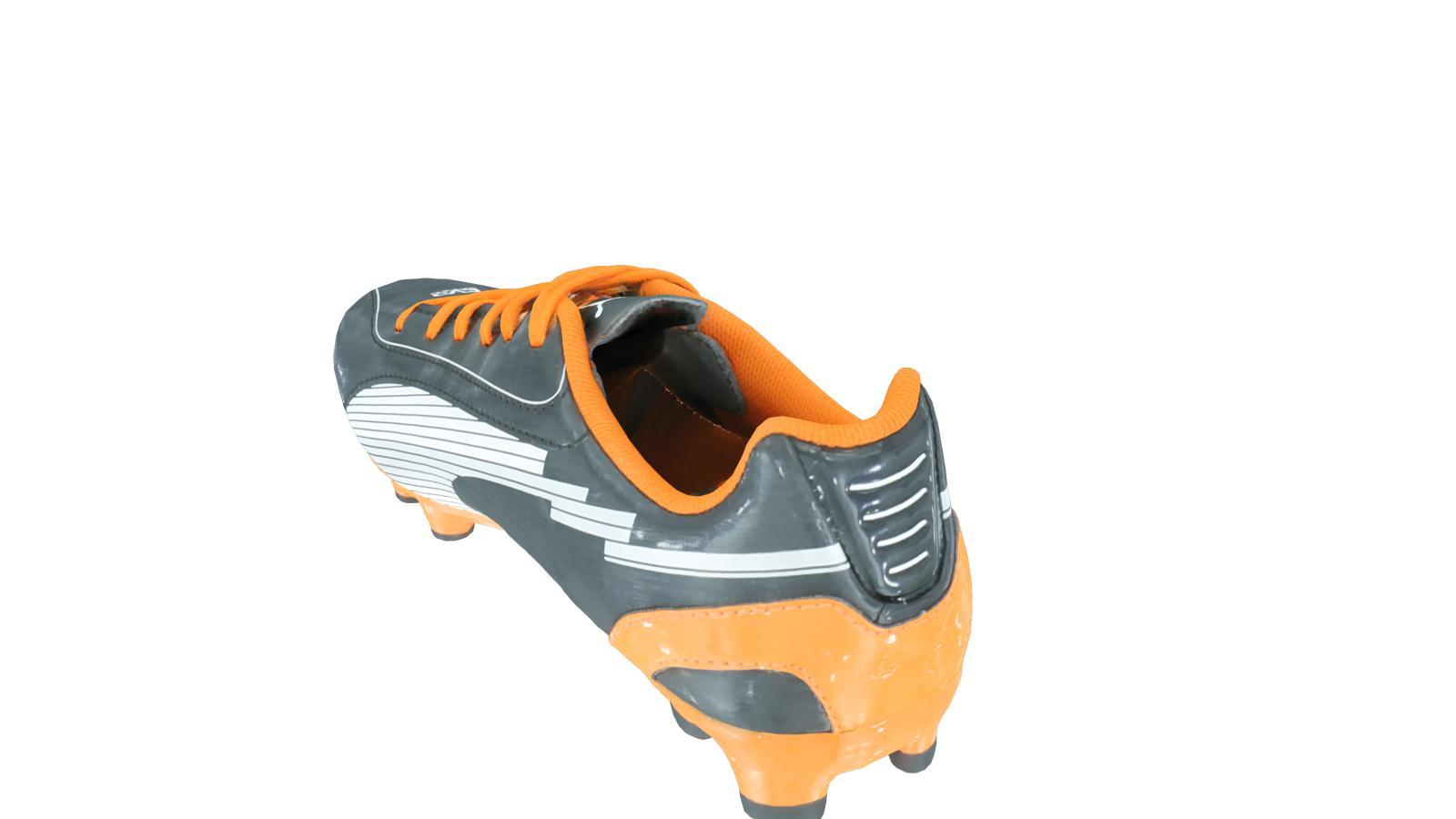}
    \end{subfigure}
    \begin{subfigure}{0.155\textwidth}
        \includegraphics[width=\linewidth]{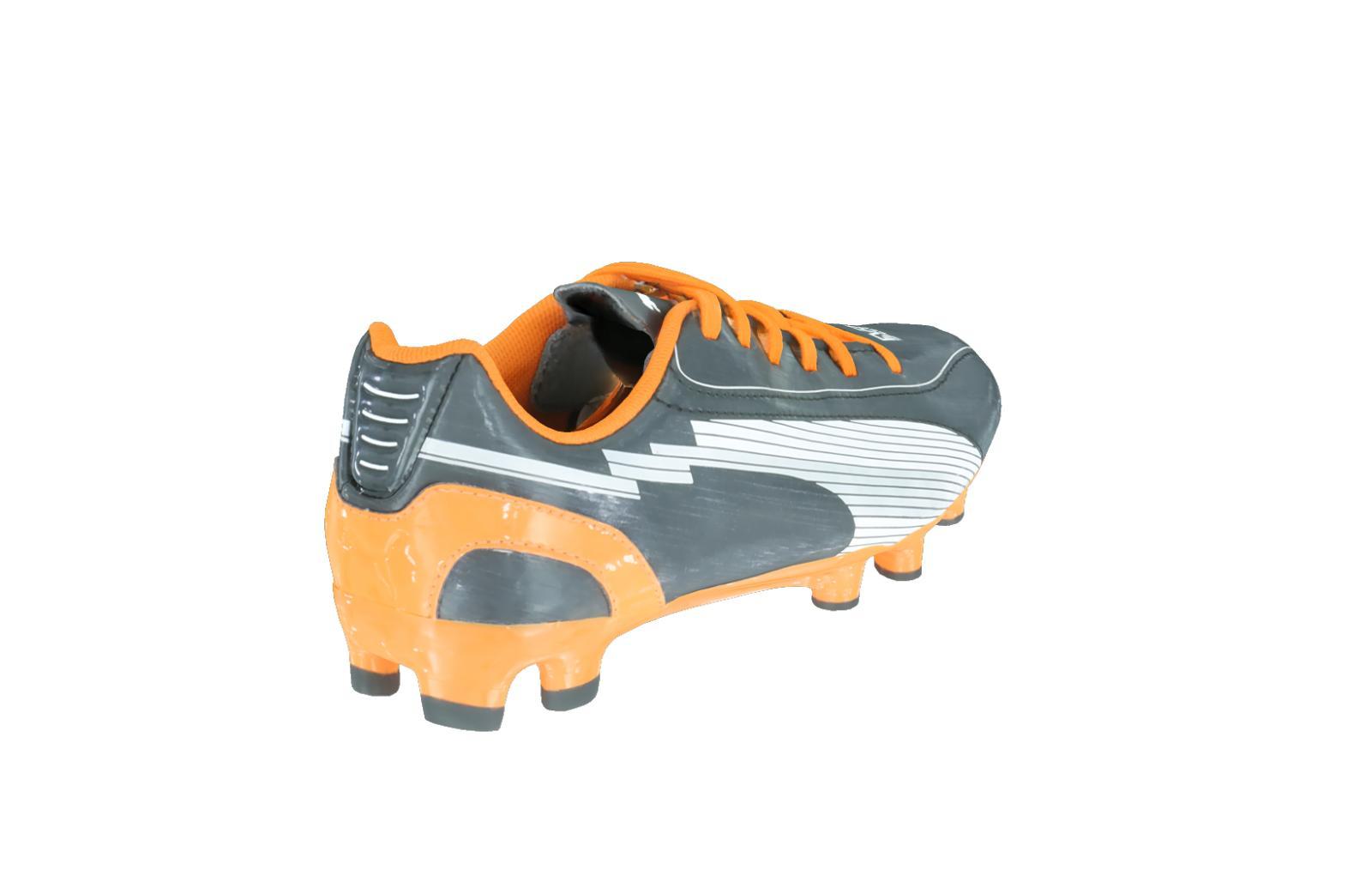}
    \end{subfigure}
    \hspace{3mm} 
    \begin{subfigure}{0.155\textwidth}
        \includegraphics[width=\linewidth]{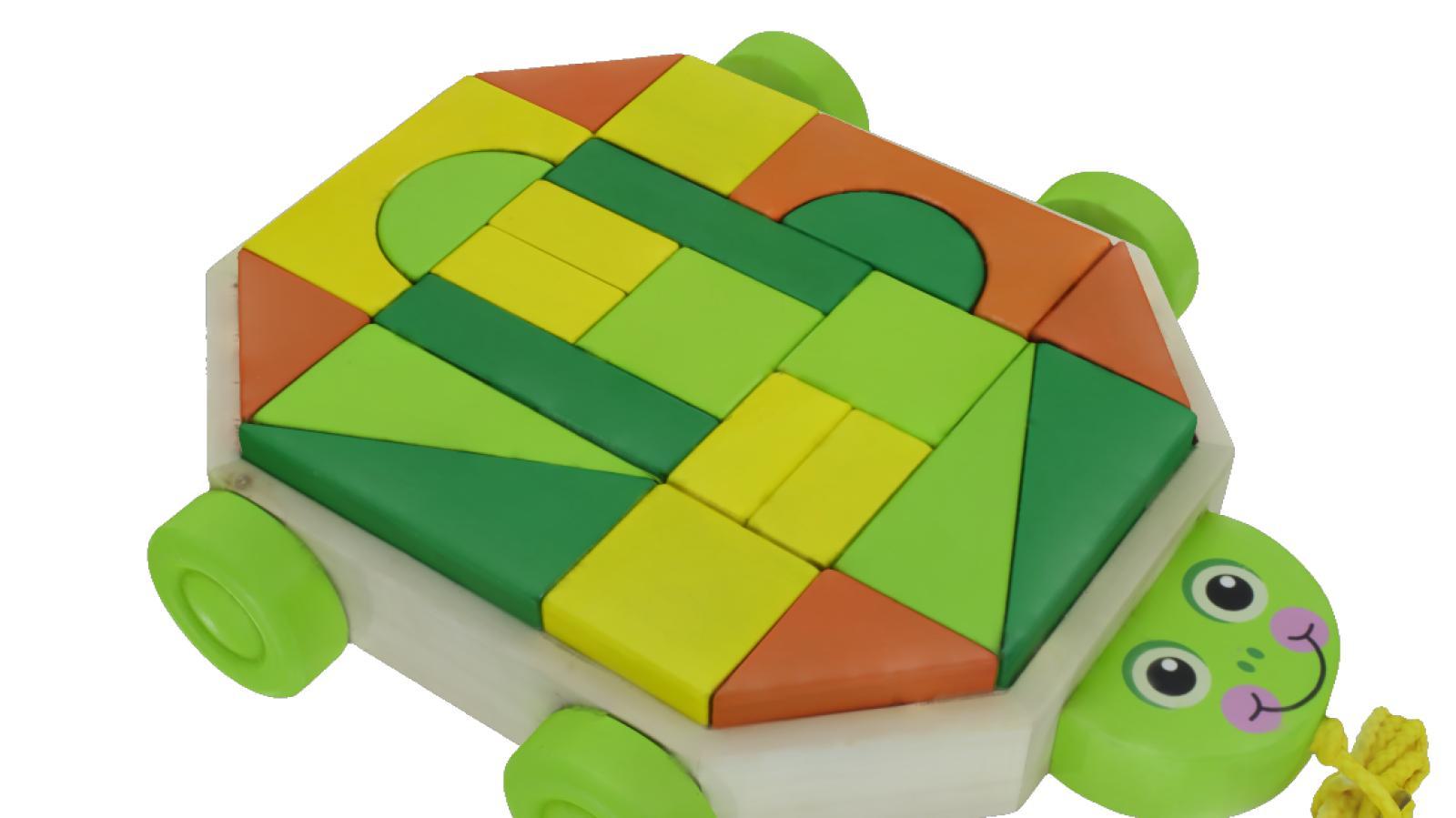}
    \end{subfigure}
    \begin{subfigure}{0.155\textwidth}
        \includegraphics[width=\linewidth]{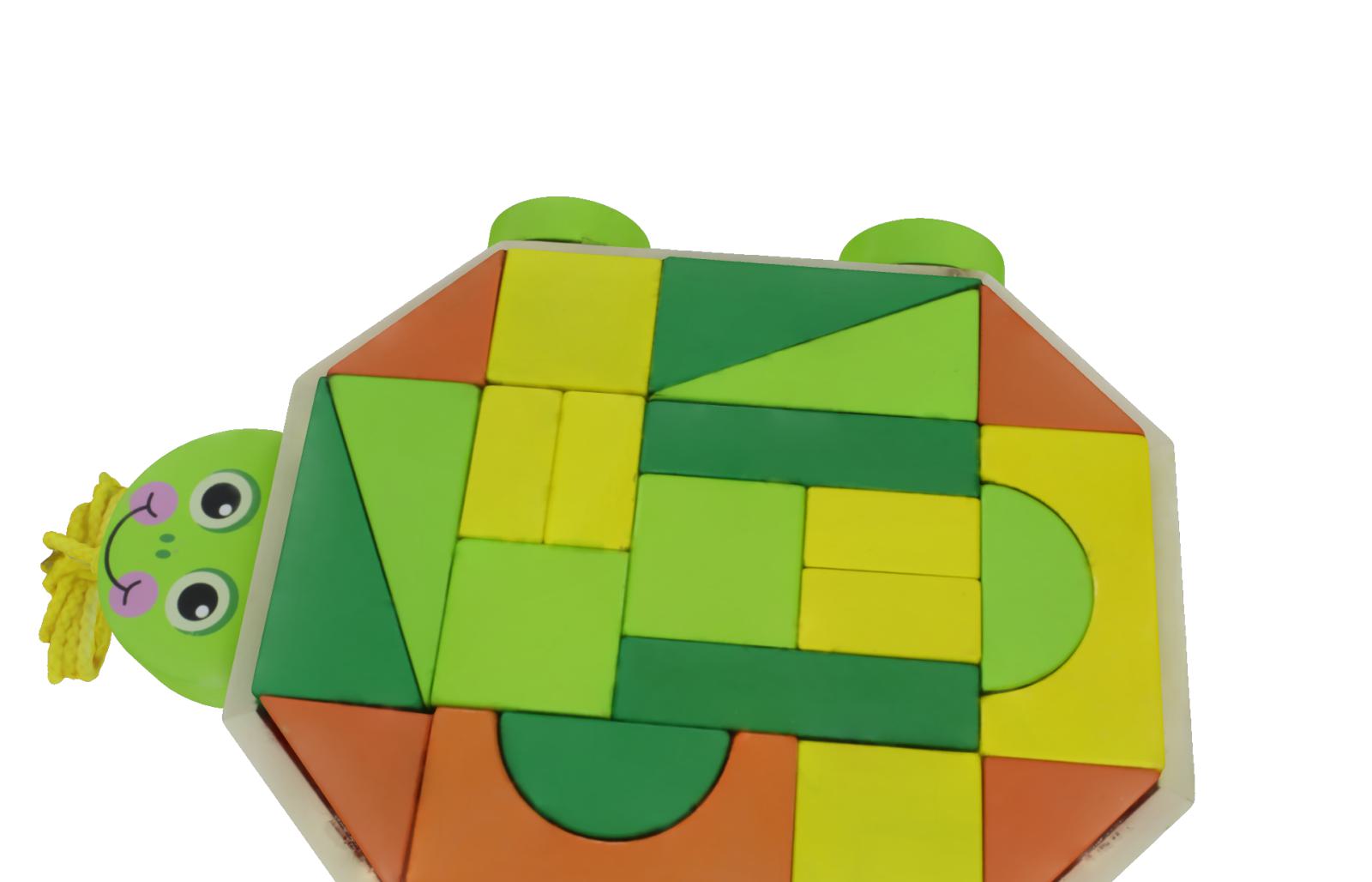}
    \end{subfigure}
    \hspace{3mm} 
    \begin{subfigure}{0.155\textwidth}
        \includegraphics[width=\linewidth]{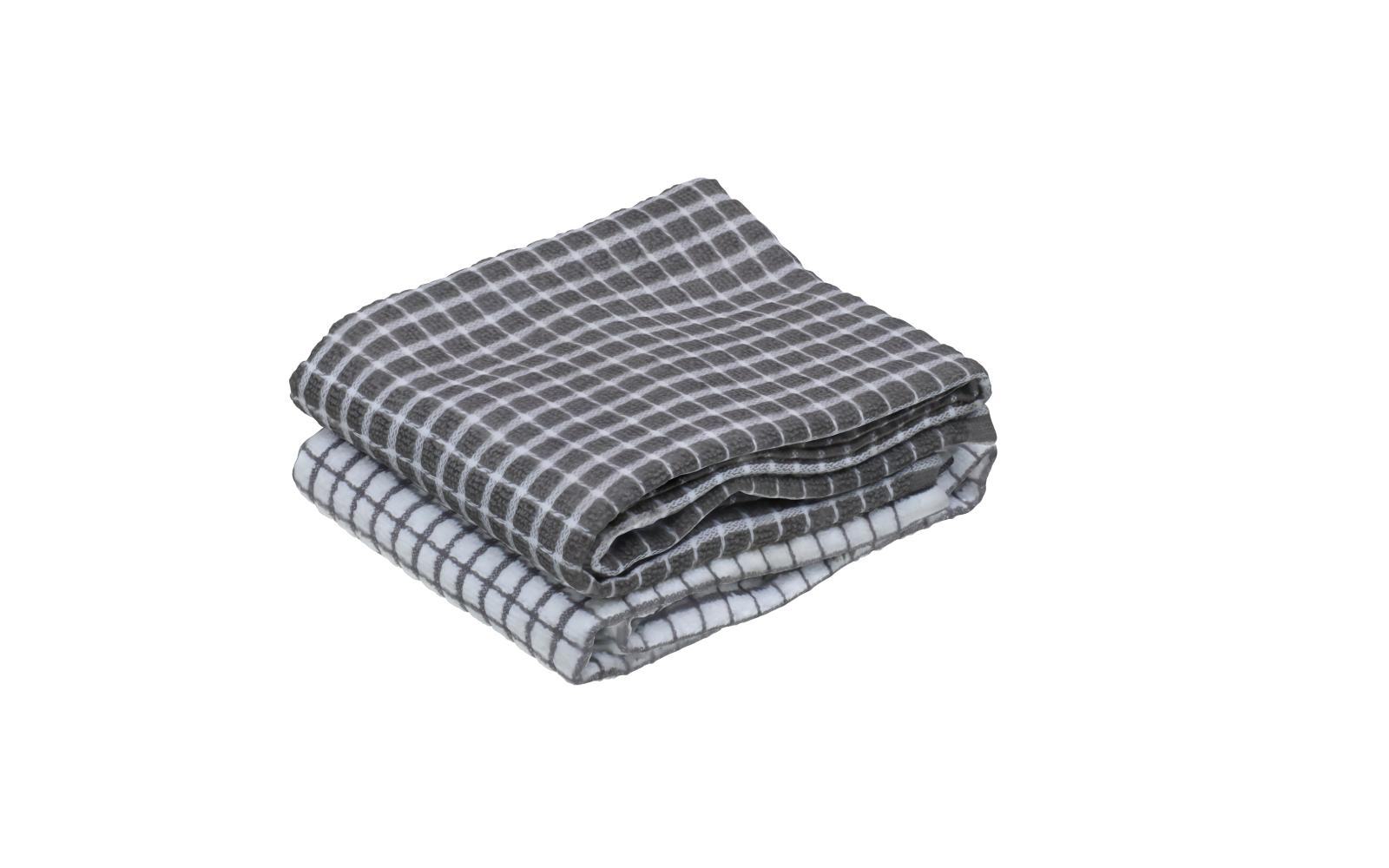}
    \end{subfigure}
    \begin{subfigure}{0.155\textwidth}
        \includegraphics[width=\linewidth]{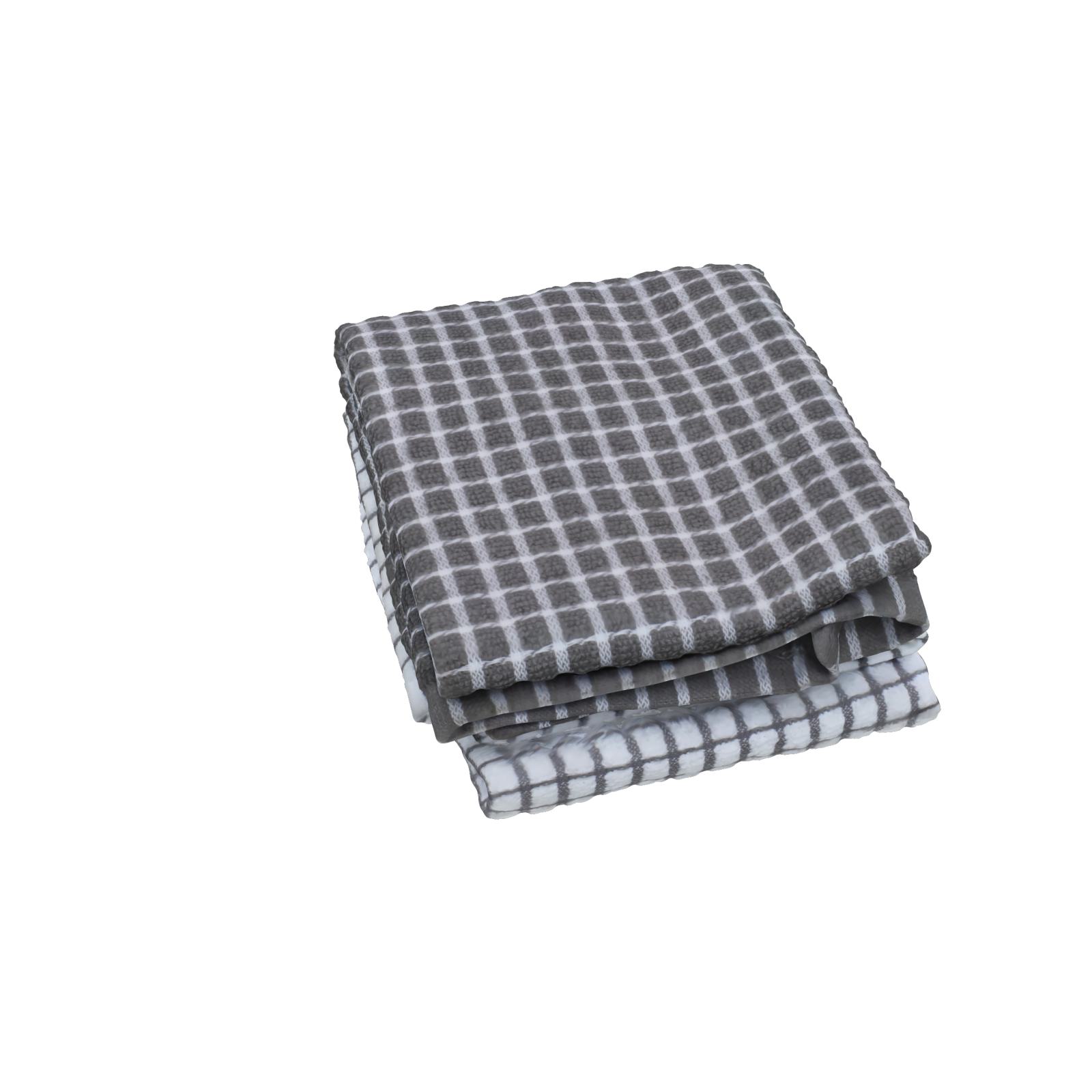}
    \end{subfigure}

    \bigskip
    
    \begin{subfigure}{0.155\textwidth}
        \includegraphics[width=\linewidth]{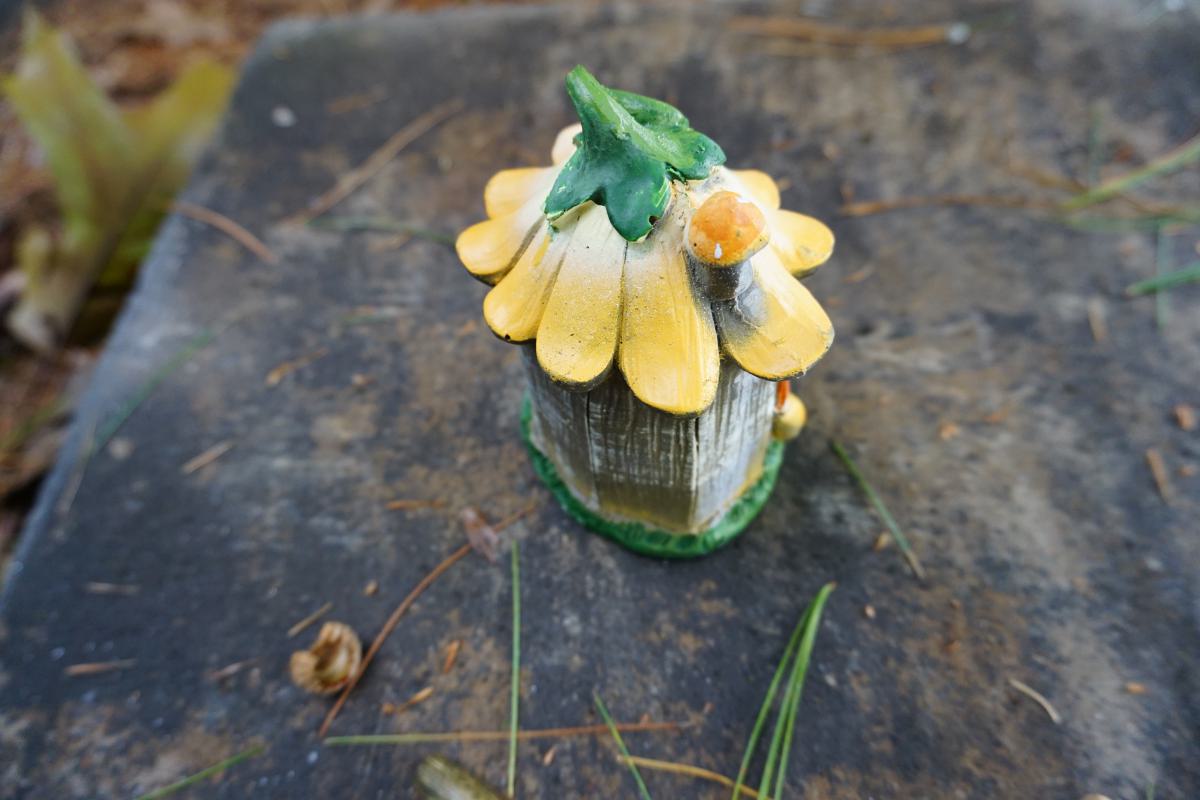}
    \end{subfigure}
    \begin{subfigure}{0.155\textwidth}
        \includegraphics[width=\linewidth]{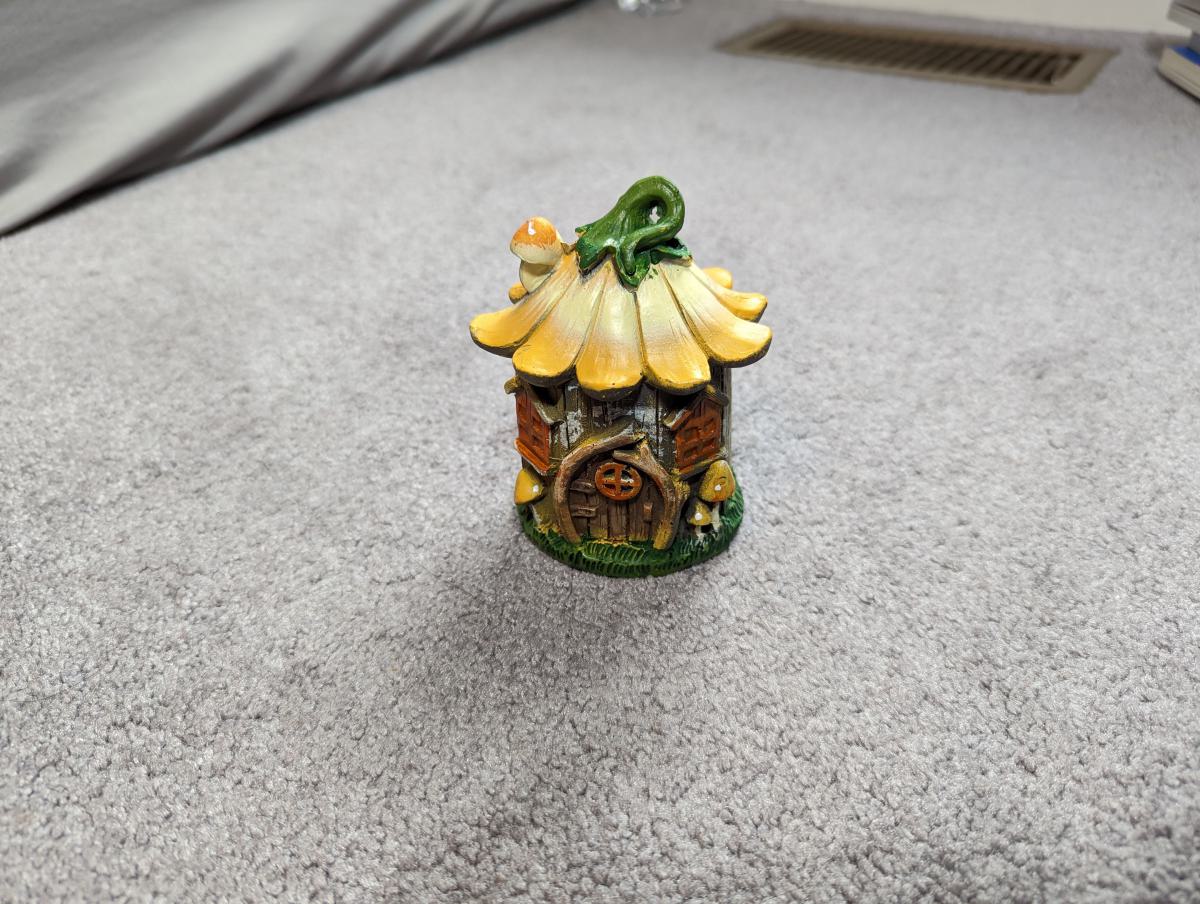}
    \end{subfigure}
    \hspace{3mm} 
    \begin{subfigure}{0.155\textwidth}
        \includegraphics[width=\linewidth]{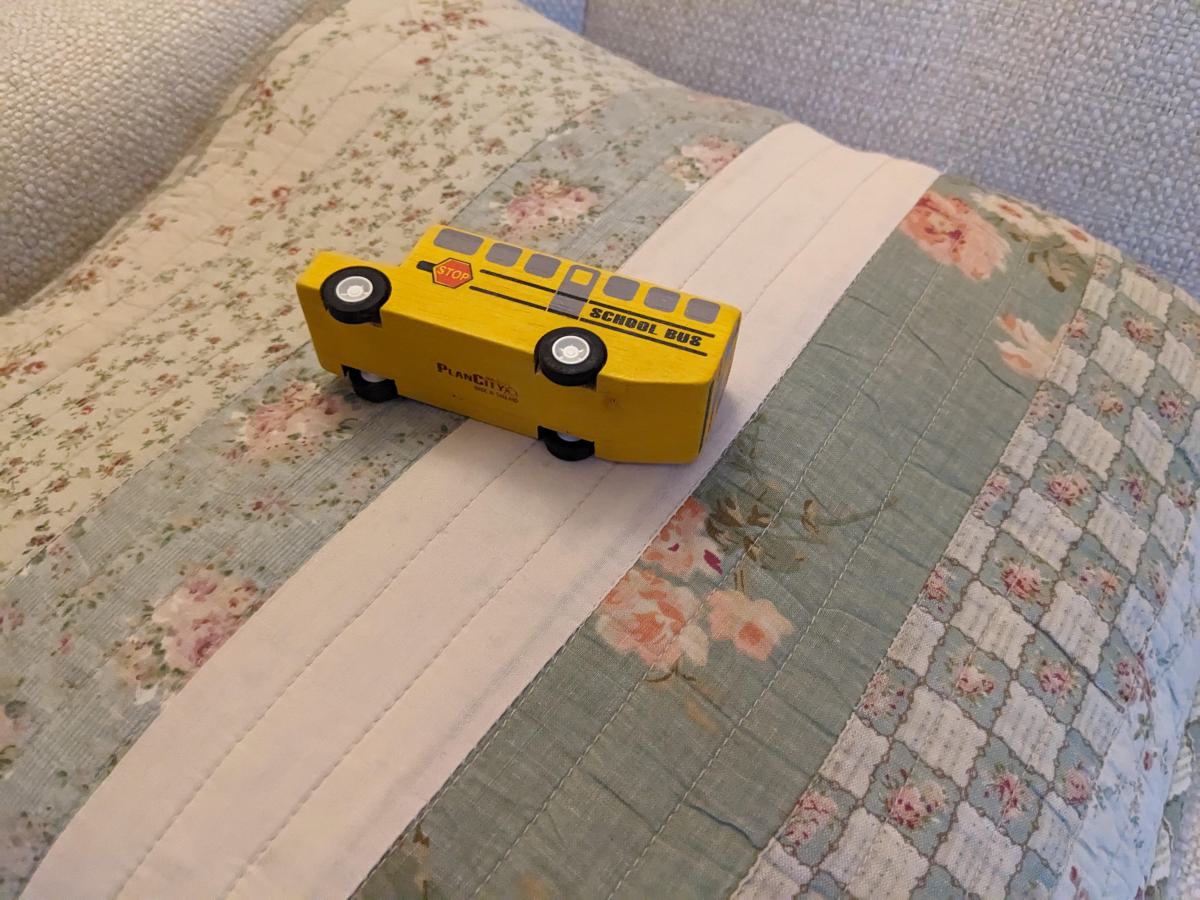}
    \end{subfigure}
    \begin{subfigure}{0.155\textwidth}
        \includegraphics[width=\linewidth]{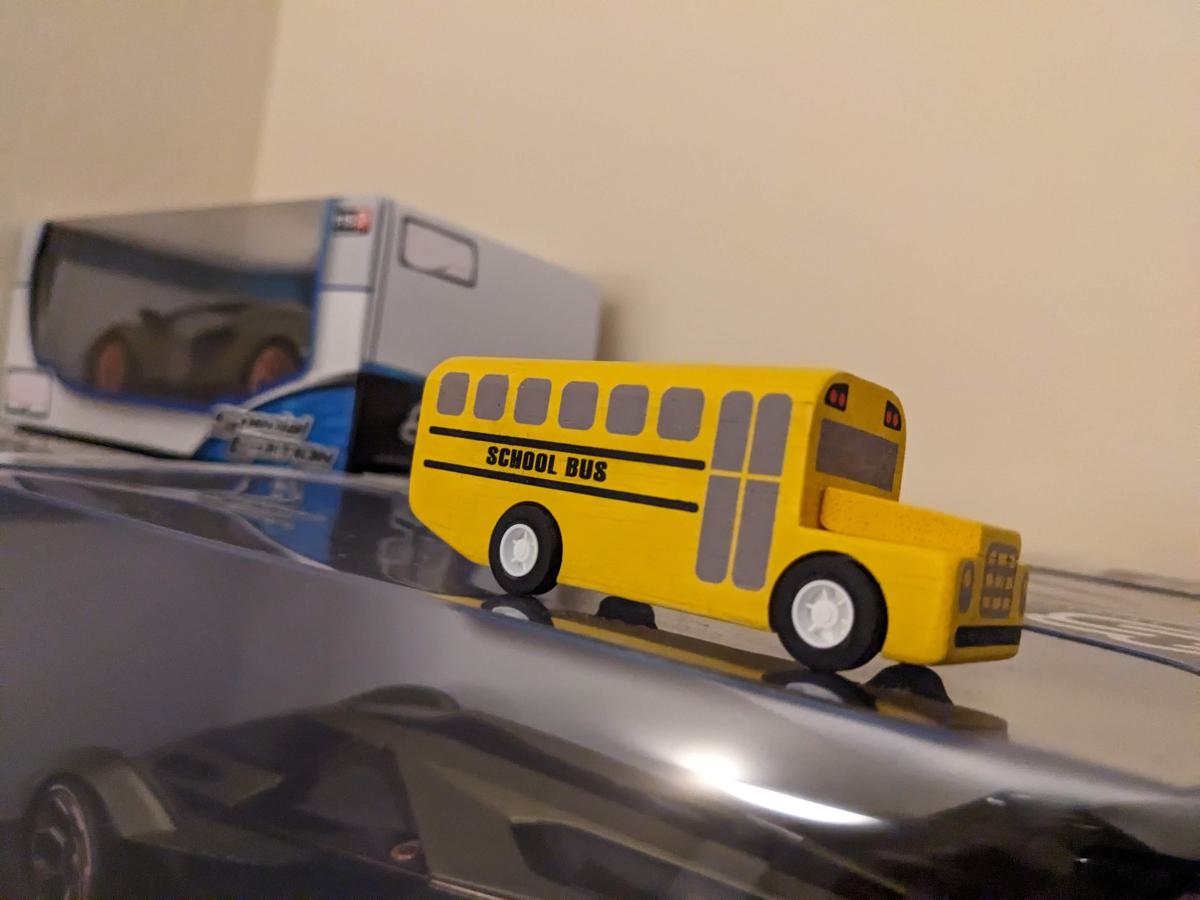}
    \end{subfigure}
    \hspace{3mm} 
    \begin{subfigure}{0.155\textwidth}
        \includegraphics[width=\linewidth]{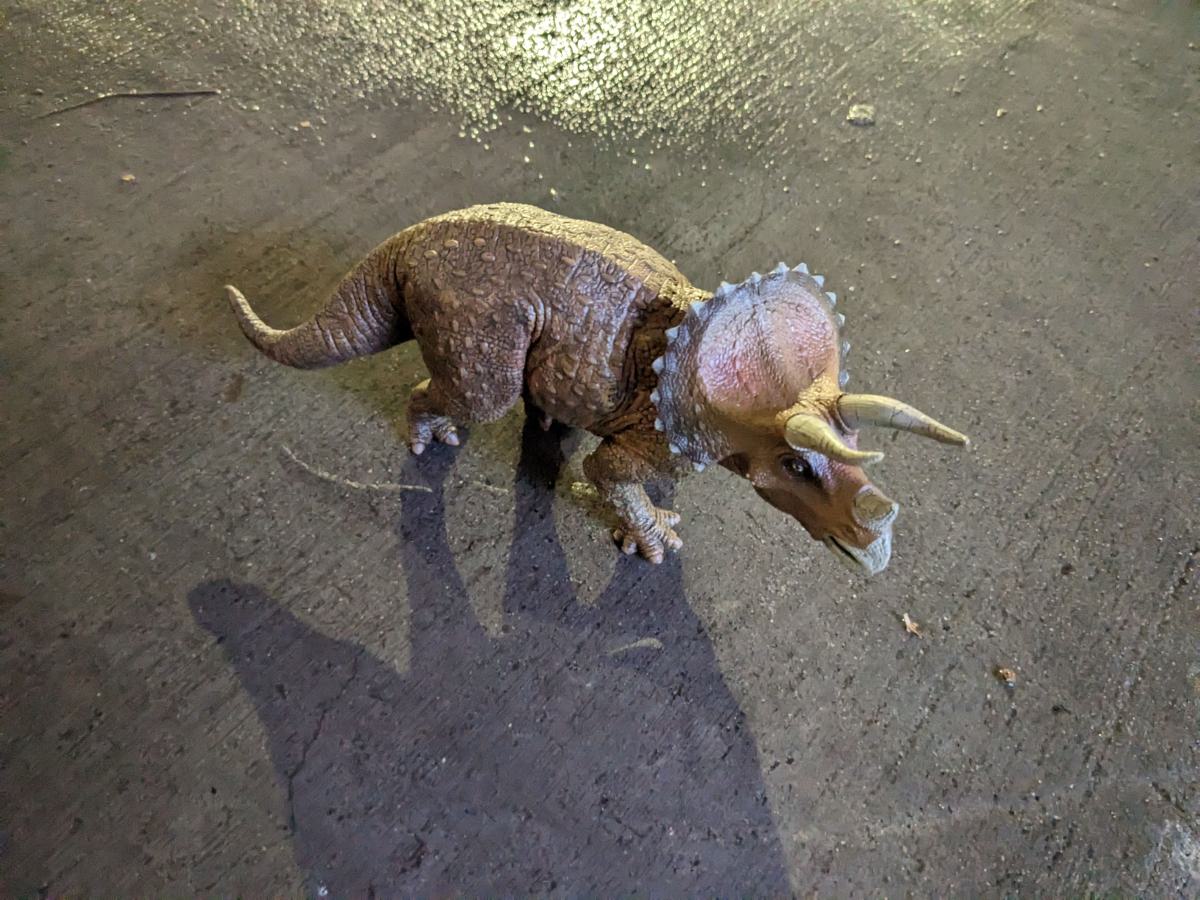}
    \end{subfigure}
    \begin{subfigure}{0.155\textwidth}
        \includegraphics[width=\linewidth]{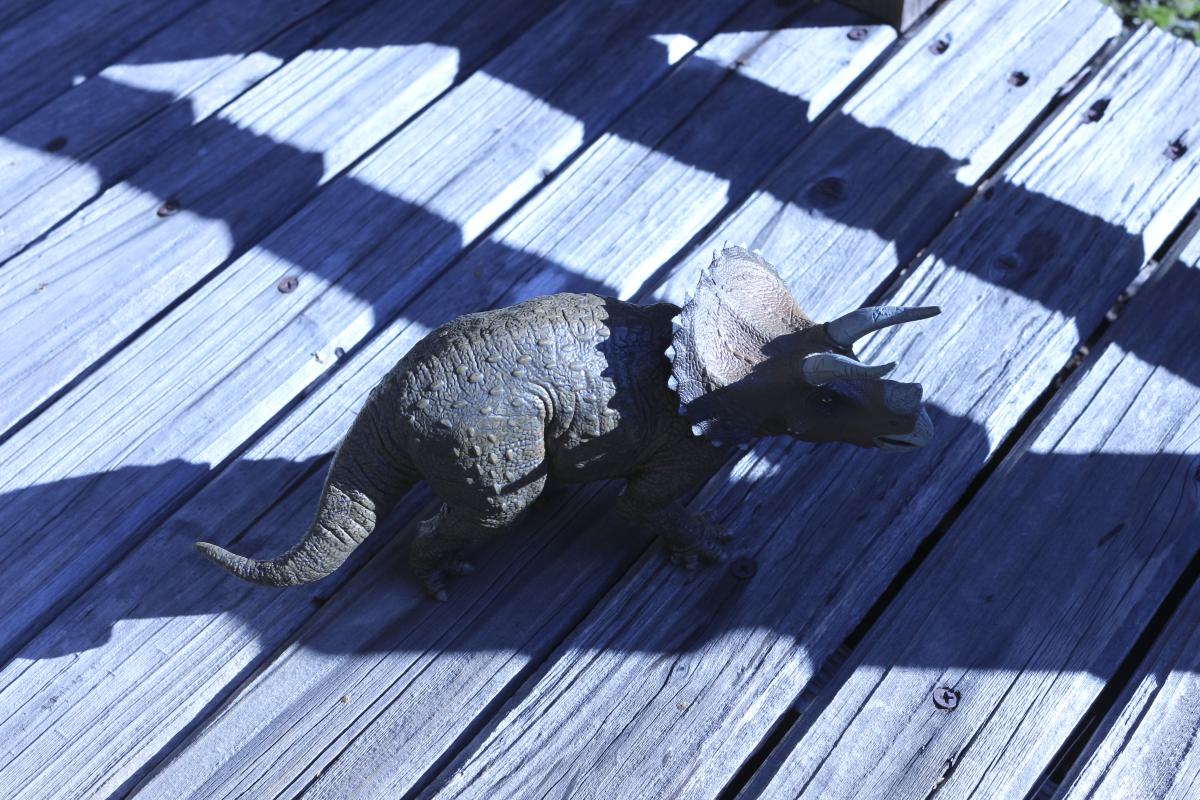}
    \end{subfigure}

    \bigskip
    
    \begin{subfigure}{0.155\textwidth}
        \includegraphics[width=\linewidth]{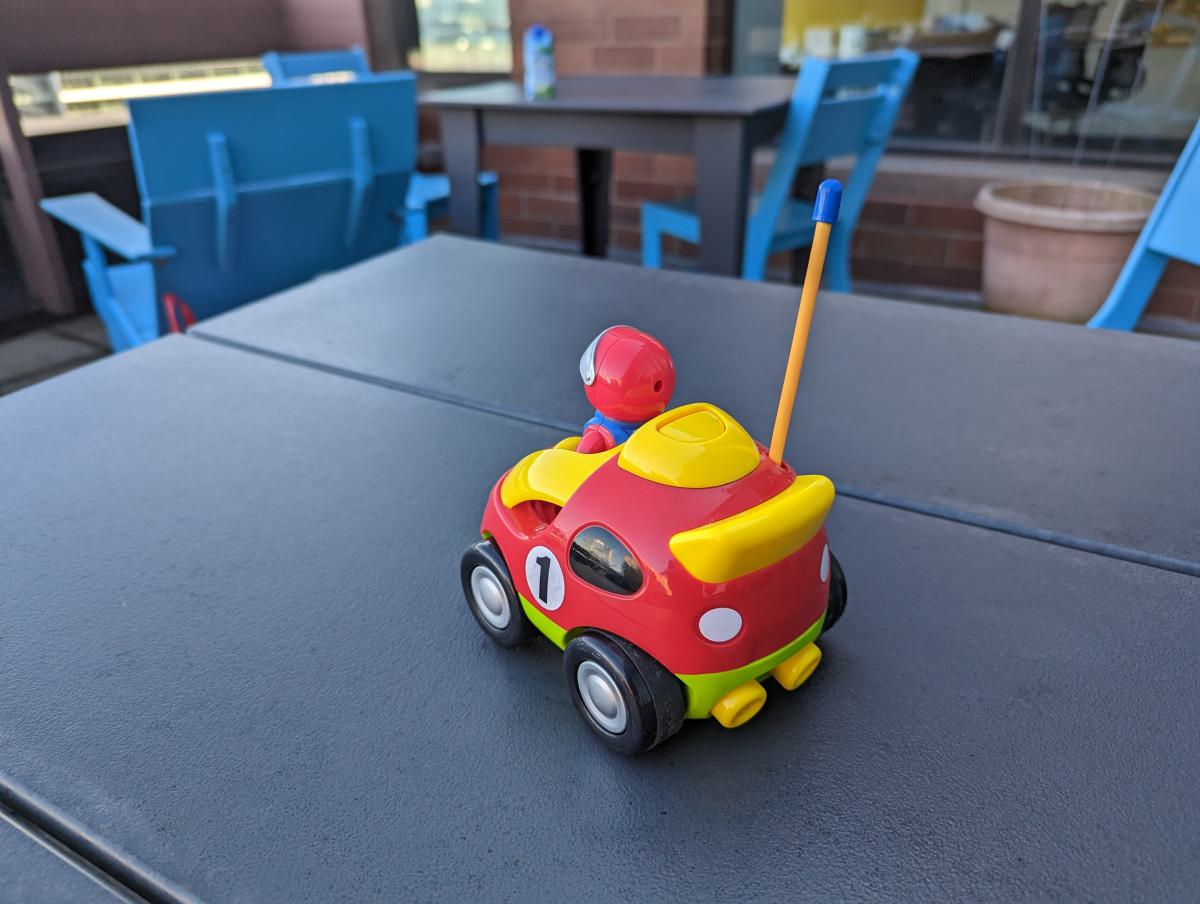}
    \end{subfigure}
    \begin{subfigure}{0.155\textwidth}
        \includegraphics[width=\linewidth]{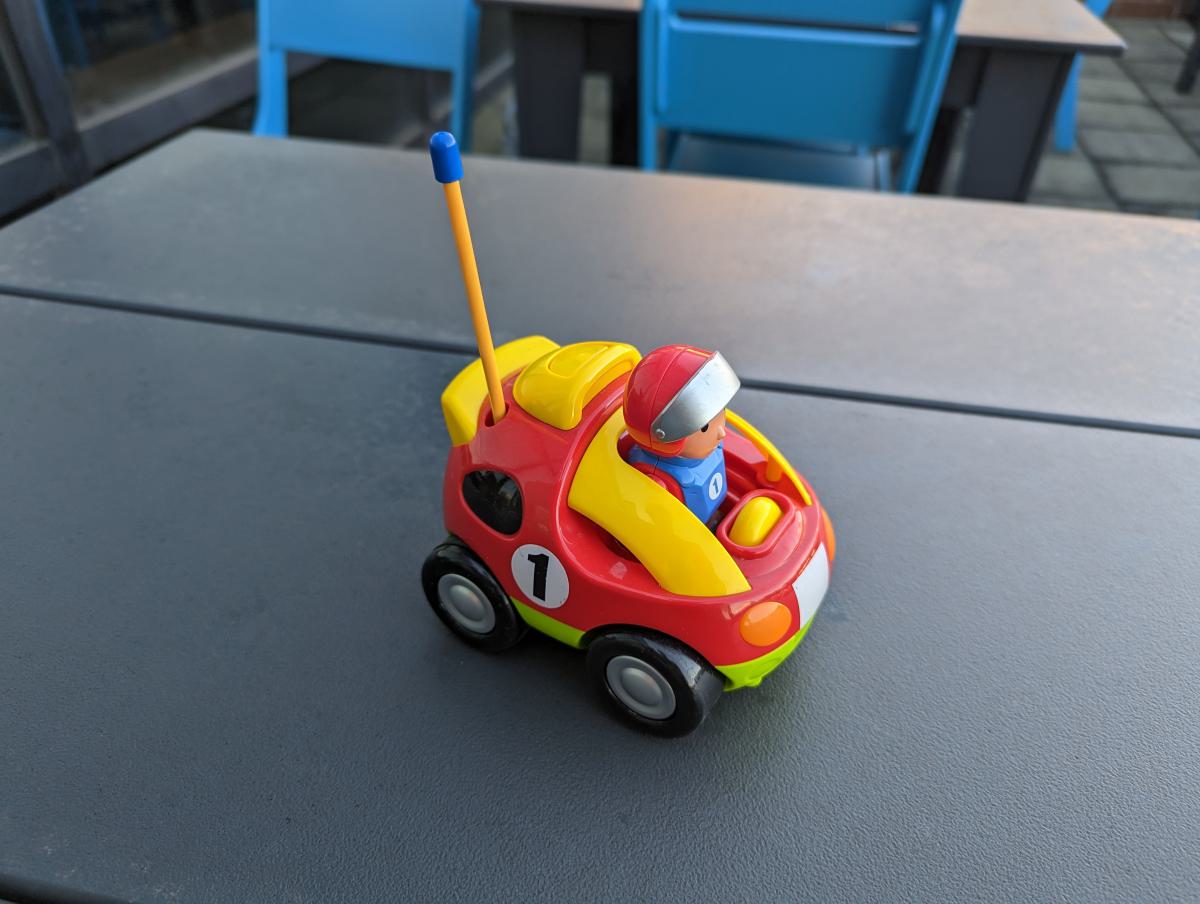}
    \end{subfigure}
    \hspace{3mm} 
    \begin{subfigure}{0.155\textwidth}
        \includegraphics[width=\linewidth]{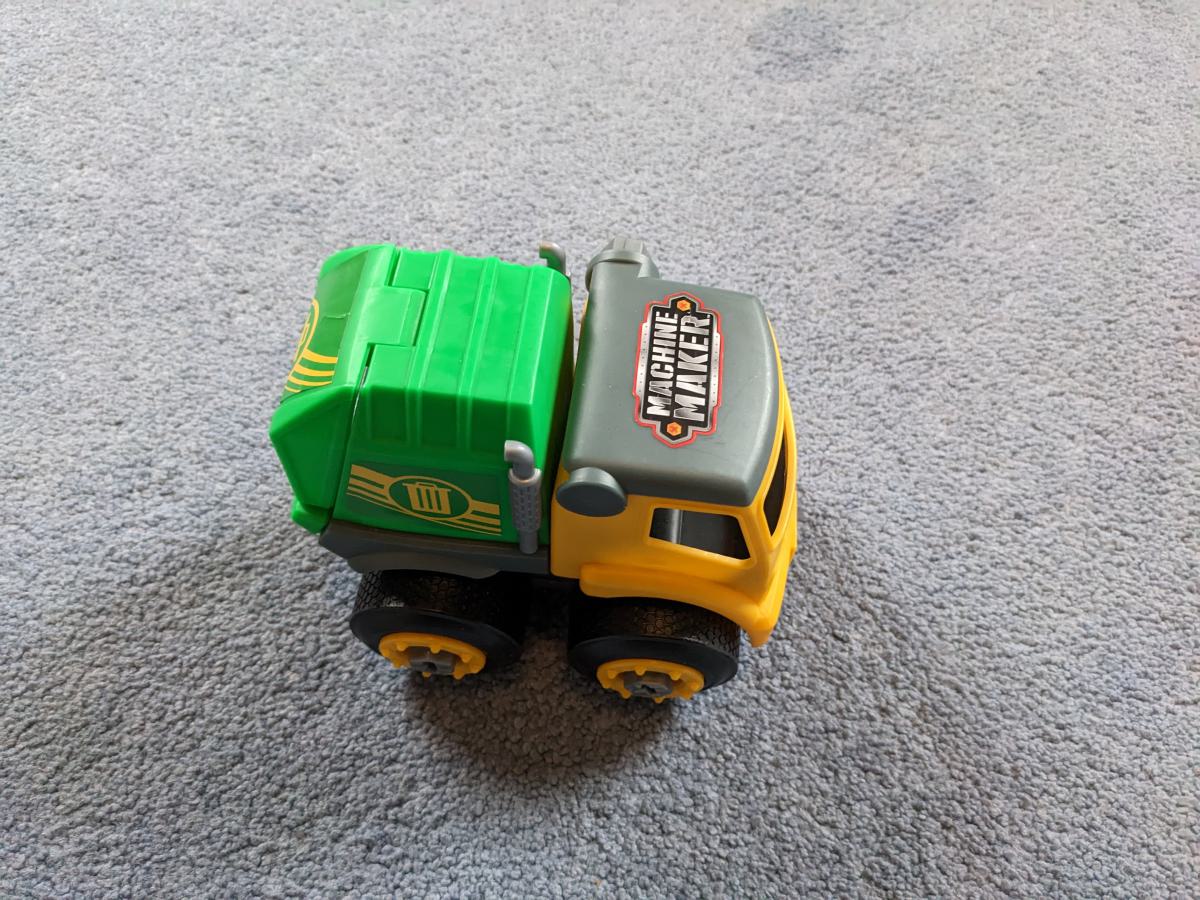}
    \end{subfigure}
    \begin{subfigure}{0.155\textwidth}
        \includegraphics[width=\linewidth]{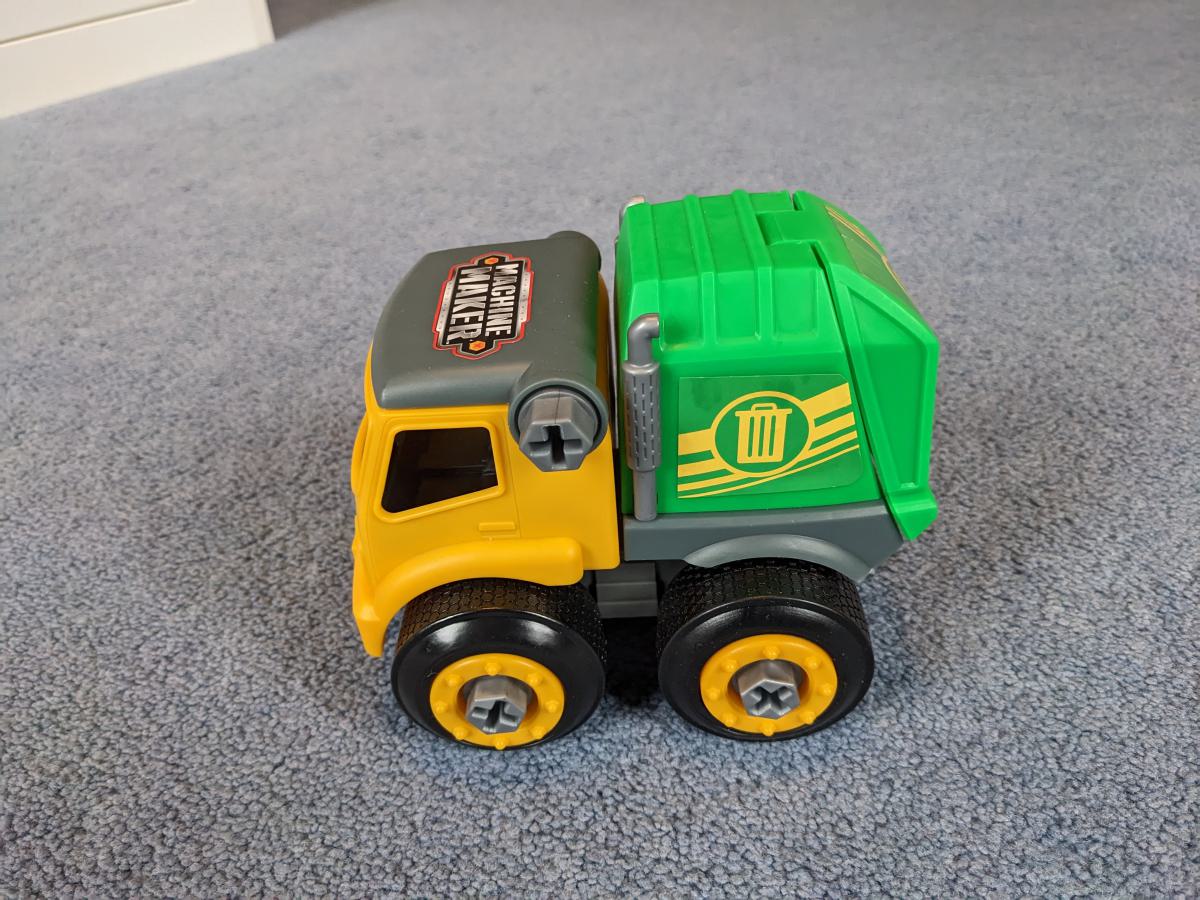}
    \end{subfigure}
    \hspace{3mm} 
    \begin{subfigure}{0.155\textwidth}
        \includegraphics[width=\linewidth]{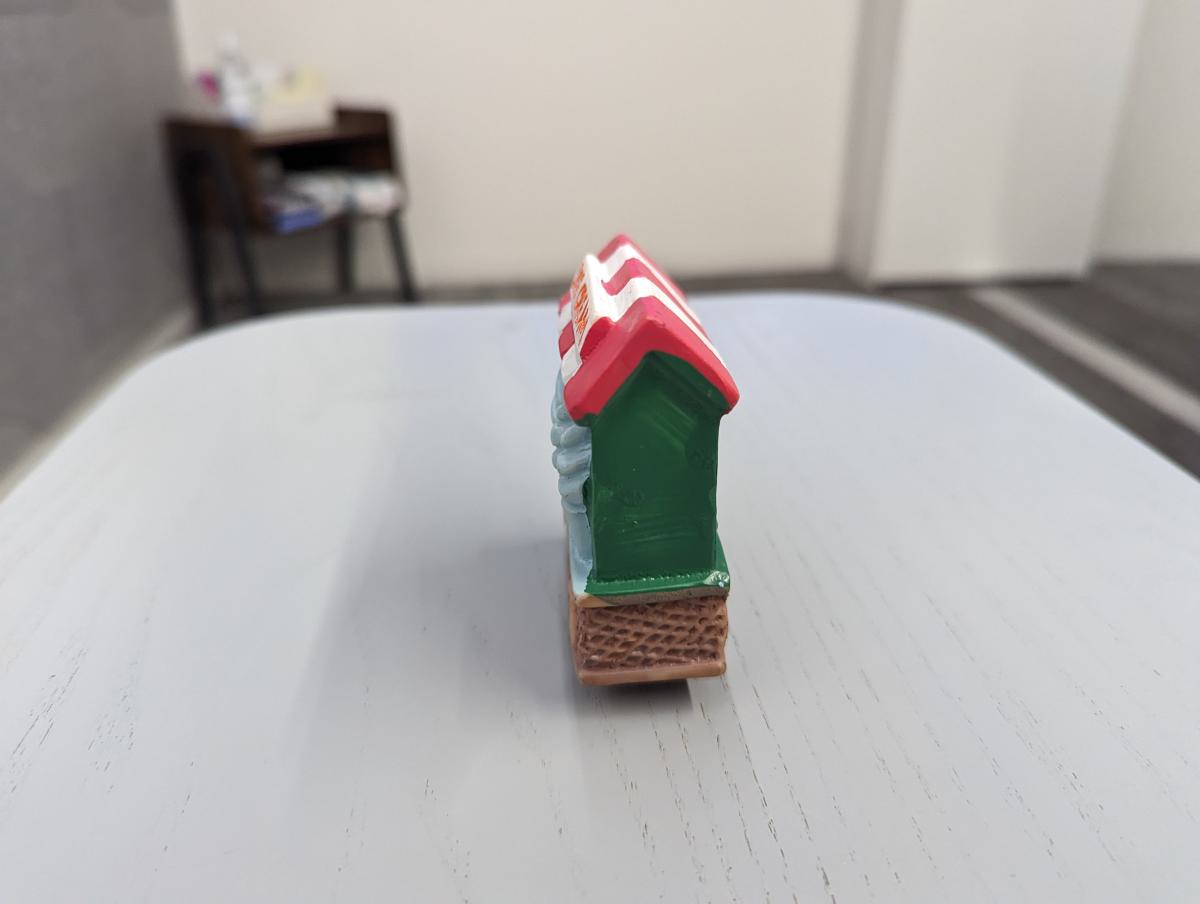}
    \end{subfigure}
    \begin{subfigure}{0.155\textwidth}
        \includegraphics[width=\linewidth]{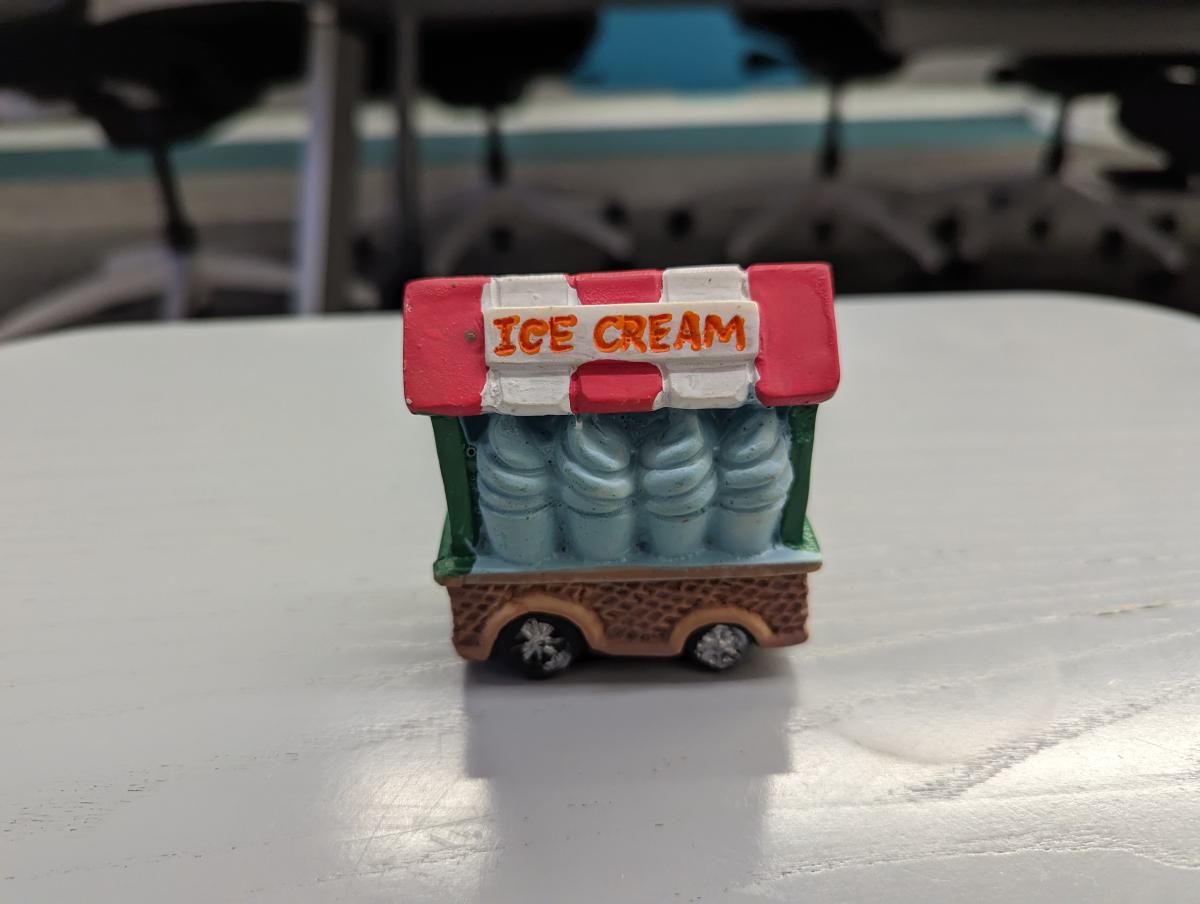}
    \end{subfigure}

    \bigskip
    
    \begin{subfigure}{0.155\textwidth}
        \includegraphics[width=\linewidth]{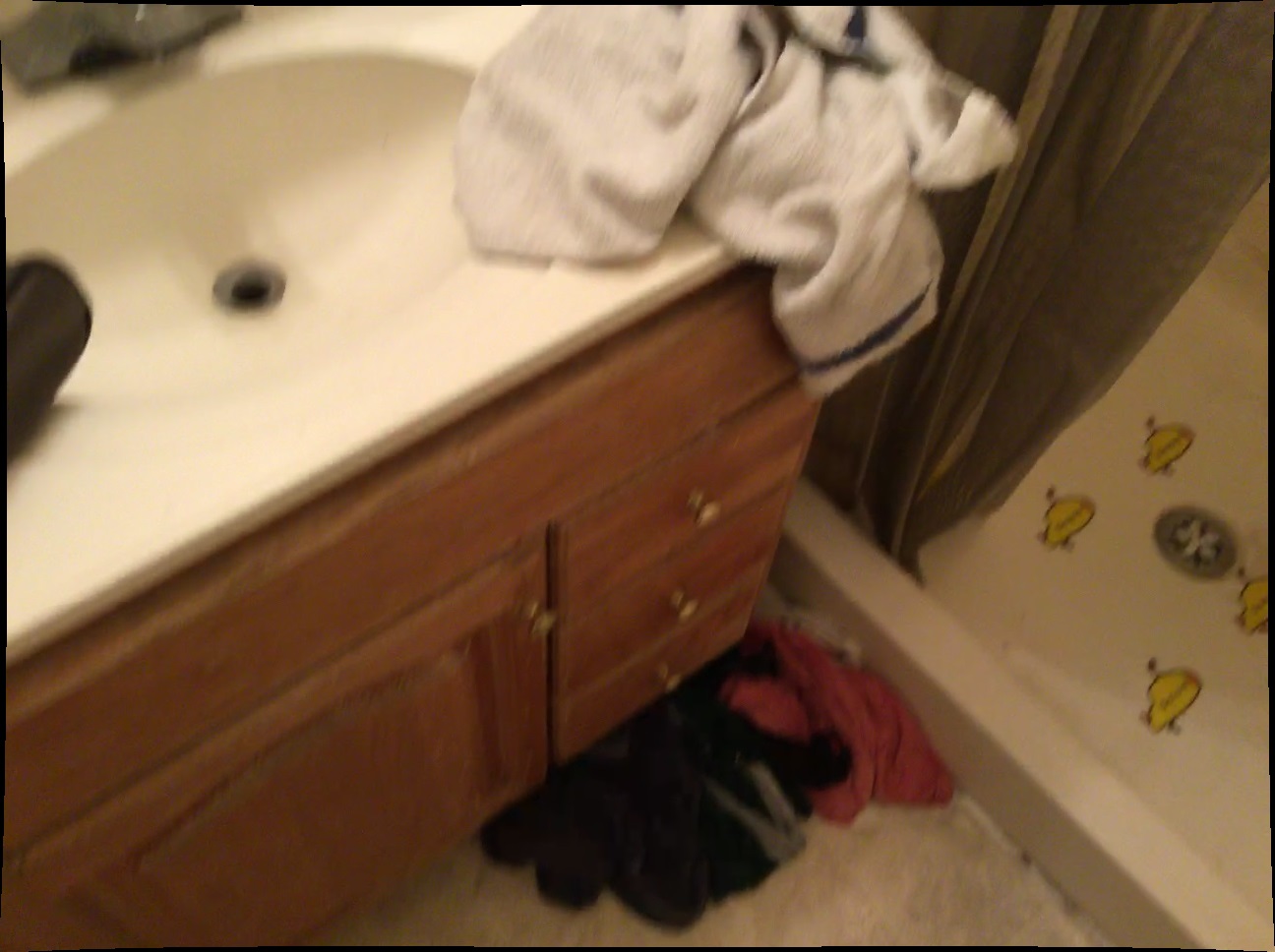}
    \end{subfigure}
    \begin{subfigure}{0.155\textwidth}
        \includegraphics[width=\linewidth]{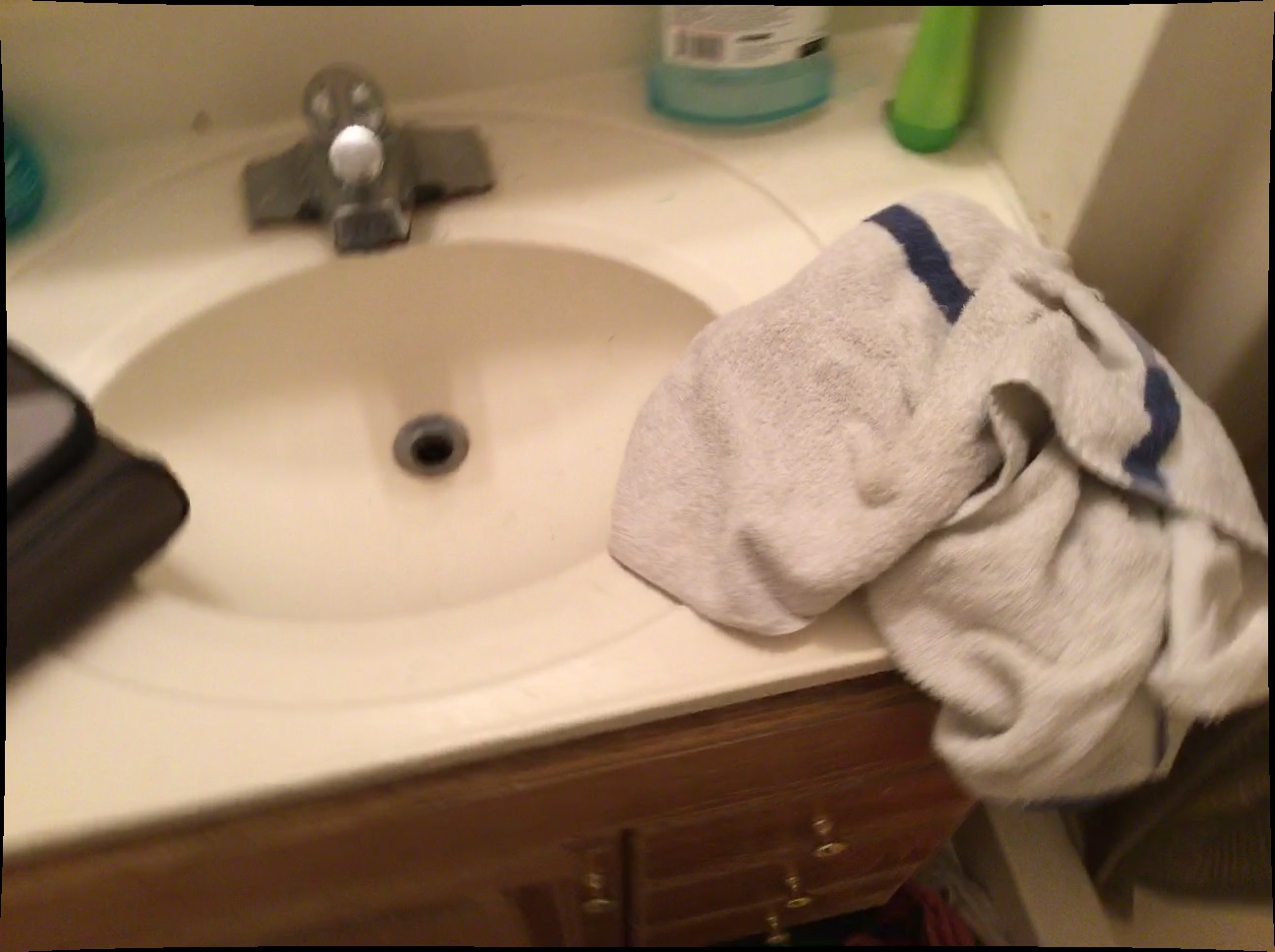}
    \end{subfigure}
    \hspace{3mm} 
    \begin{subfigure}{0.155\textwidth}
        \includegraphics[width=\linewidth]{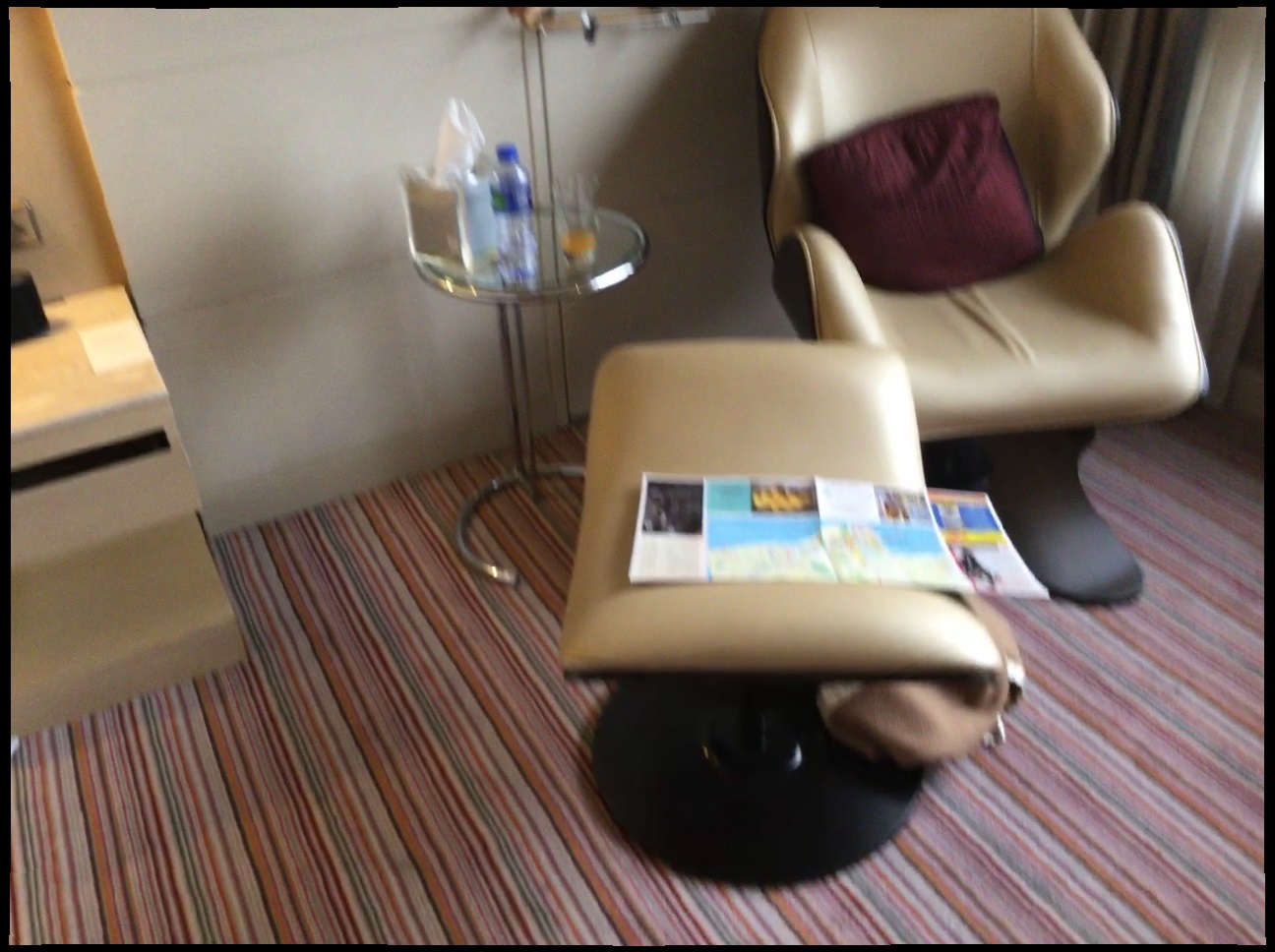}
    \end{subfigure}
    \begin{subfigure}{0.155\textwidth}
        \includegraphics[width=\linewidth]{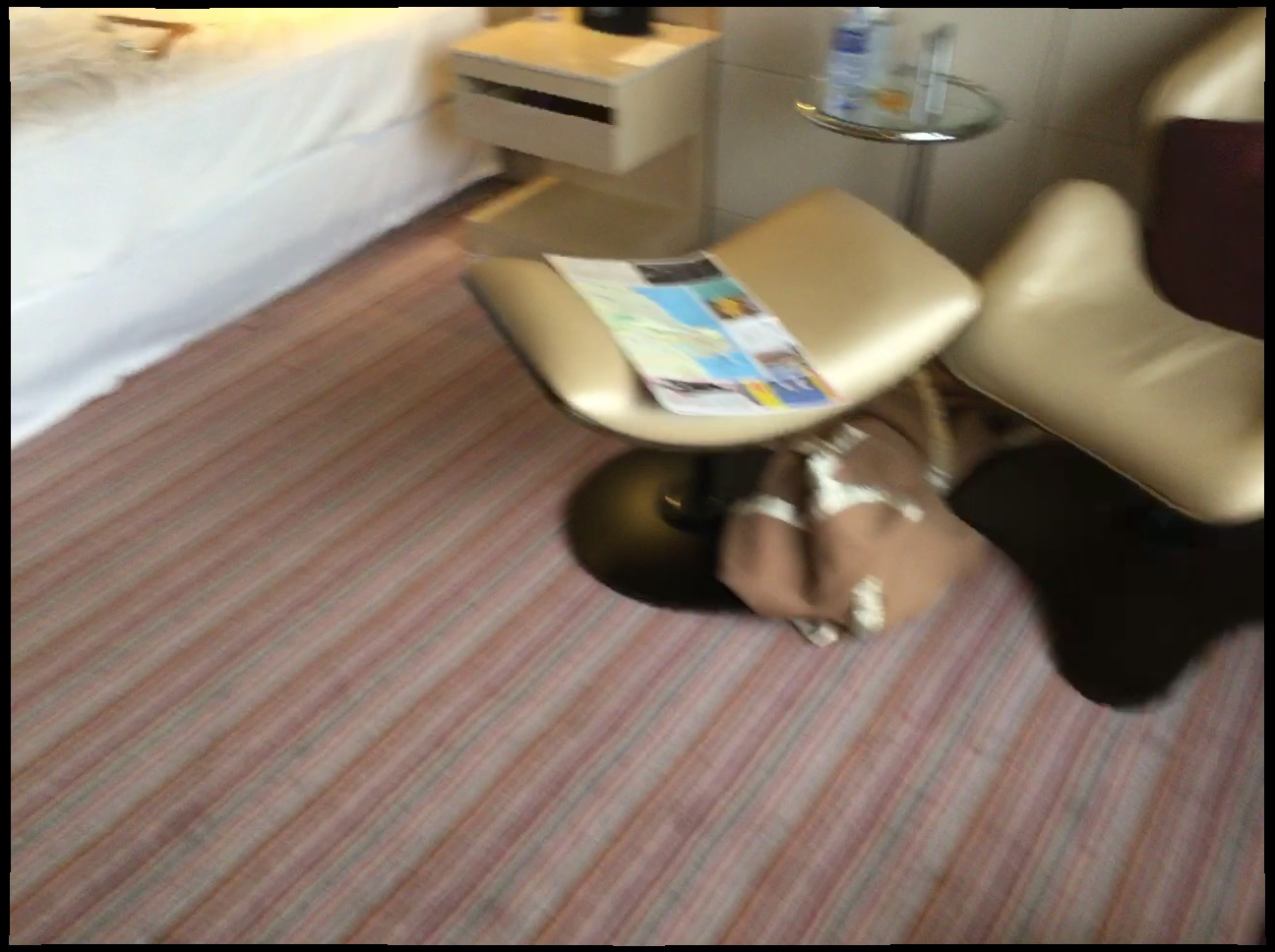}
    \end{subfigure}
    \hspace{3mm} 
    \begin{subfigure}{0.155\textwidth}
        \includegraphics[width=\linewidth]{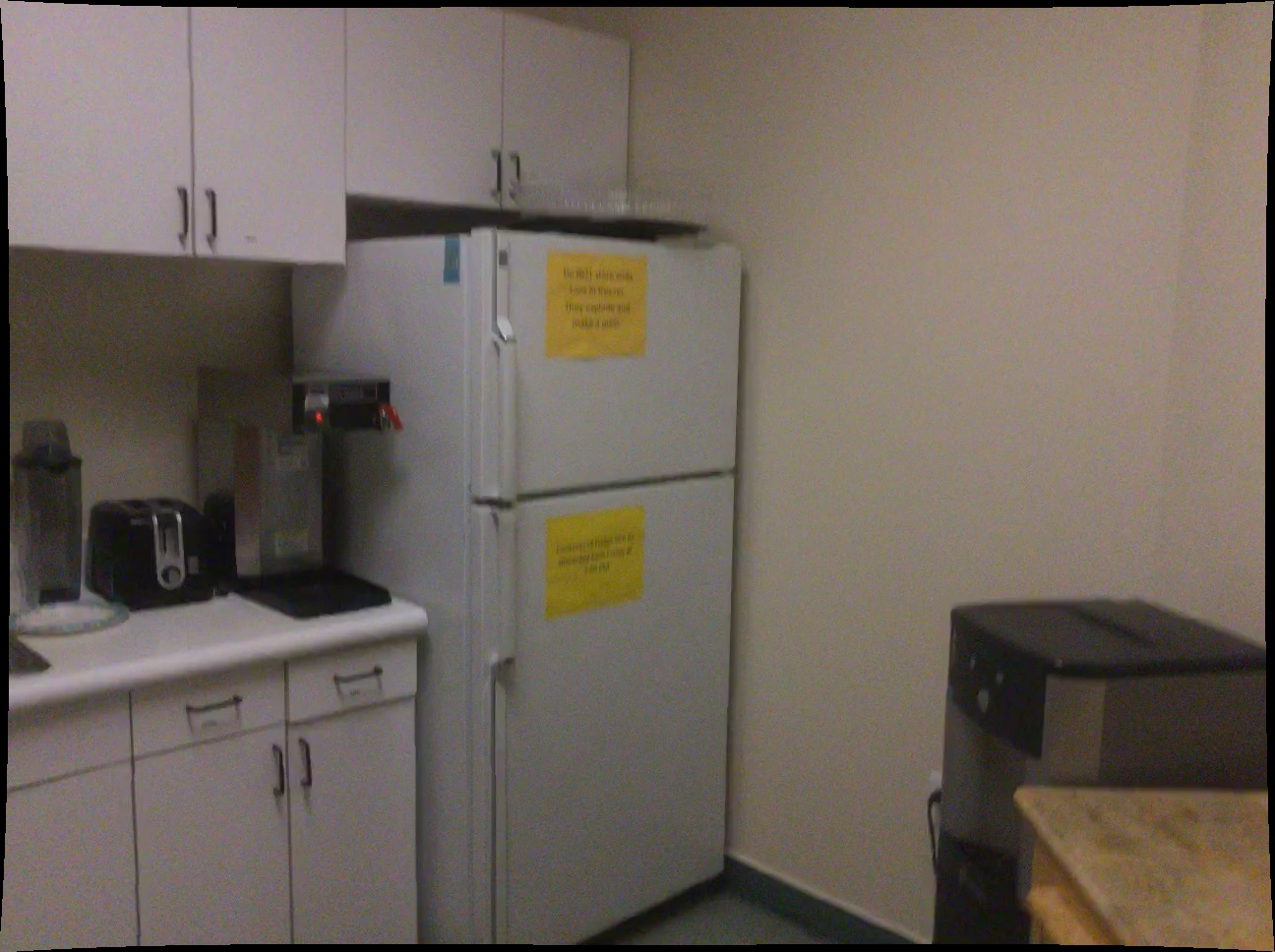}
    \end{subfigure}
    \begin{subfigure}{0.155\textwidth}
        \includegraphics[width=\linewidth]{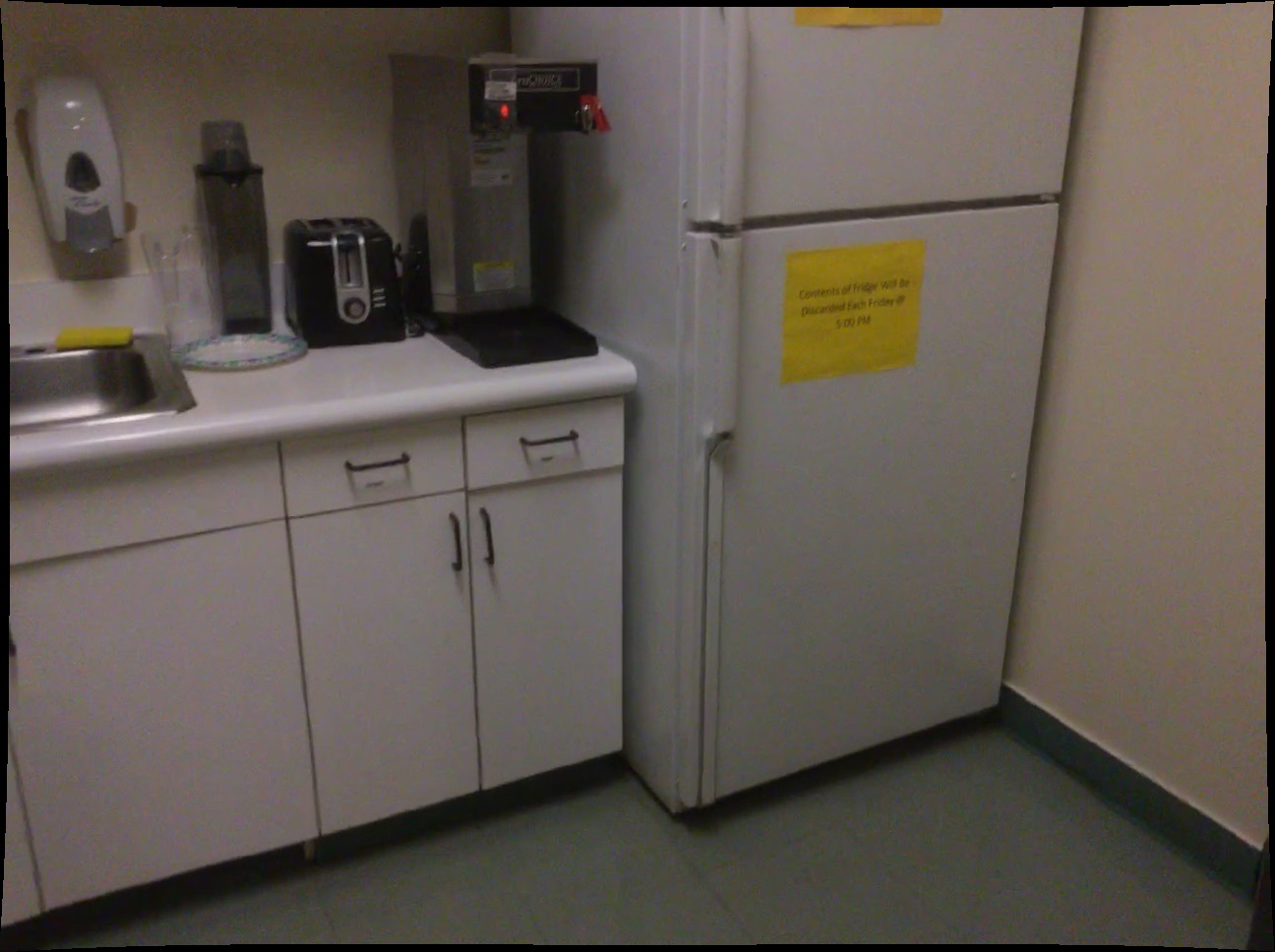}
    \end{subfigure}

    \bigskip
    
    \begin{subfigure}{0.155\textwidth}
        \includegraphics[width=\linewidth]{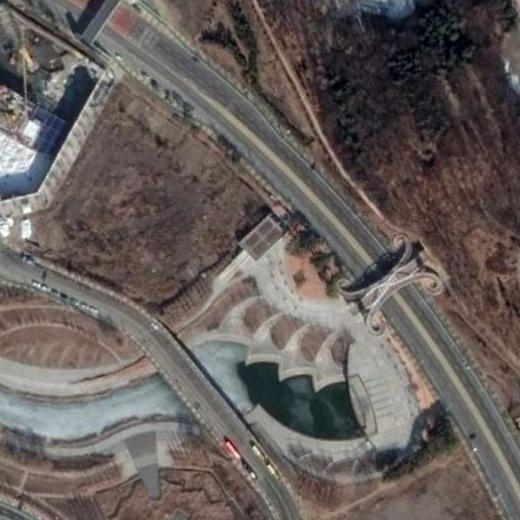}
    \end{subfigure}
    \begin{subfigure}{0.155\textwidth}
        \includegraphics[width=\linewidth]{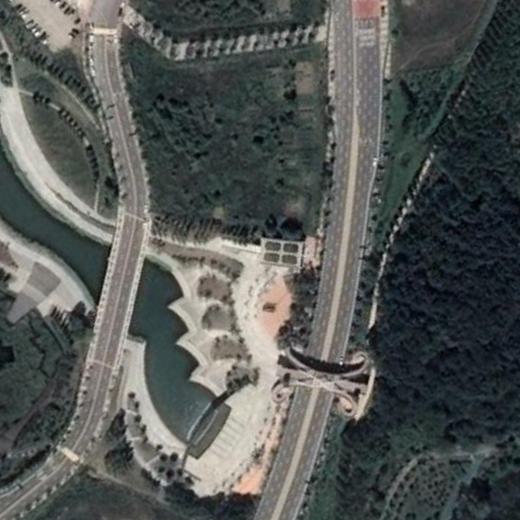}
    \end{subfigure}
    \hspace{3mm} 
    \begin{subfigure}{0.155\textwidth}
        \includegraphics[width=\linewidth]{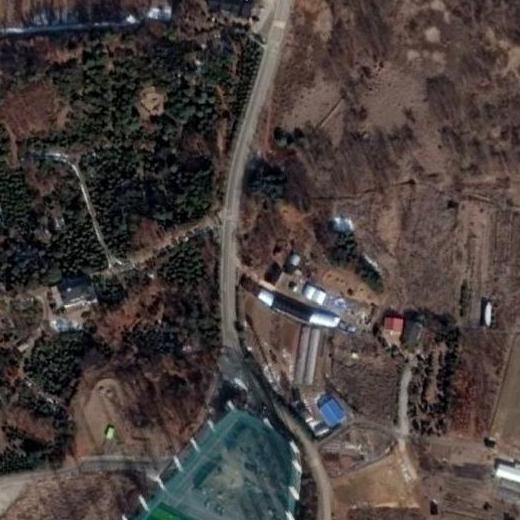}
    \end{subfigure}
    \begin{subfigure}{0.155\textwidth}
        \includegraphics[width=\linewidth]{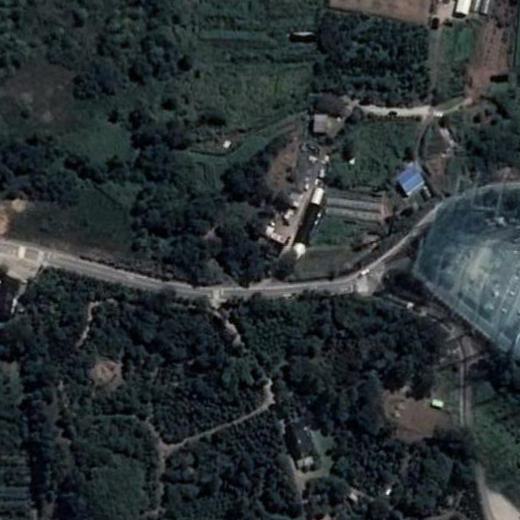}
    \end{subfigure}
    \hspace{3mm} 
    \begin{subfigure}{0.155\textwidth}
        \includegraphics[width=\linewidth]{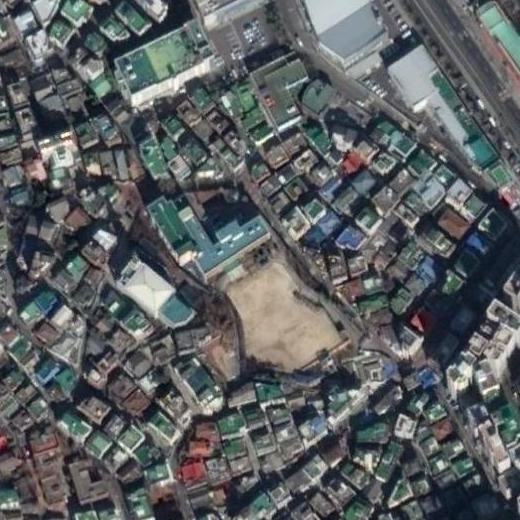}
    \end{subfigure}
    \begin{subfigure}{0.155\textwidth}
        \includegraphics[width=\linewidth]{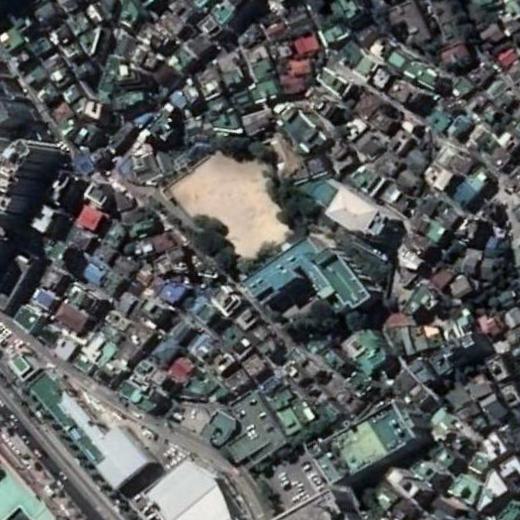}
    \end{subfigure}
    
    \caption{\textbf{Target domain examples.} We share some example image pairs from each of the target image datasets. From top row to bottom row, the domains are: Google Scanned Objects (Hard), NAVI Wild Set, NAVI Multiview, ScanNet-1500, and DeepAerial.}
    \label{fig:data_examples_appendix}
    
\end{figure*}

To illustrate the target image domains we consider in this work, Figure~\ref{fig:data_examples_appendix} presents example images pairs from each domain, namely: Google Scanned Objects~\cite{Downs2022GoogleSO}, NAVI~\cite{jampani2023navi}, ScanNet-1500~\cite{dai2017scannet}, and DeepAerial~\cite{park2020two}.
This shows that our target datasets cover a wide range of object and scene types, constituting a challenging task for generalizable image matching.

\section{Area Under Curve (AUC) Pose Results}
We also report pose AUC performance, as shown in Table~\ref{tabel: main_auc}. Because the limited performance on out-of-domain data, we report pose accuracy in the main paper.

\begin{table*}[t]
\centering
\tablestyle{5pt}{1.1}
\fontsize{7pt}{8pt}\selectfont
\vspace{-0.05in}
\setlength\tabcolsep{6pt}
\renewcommand{\arraystretch}{1.25}
\begin{tabular}{c|c|c|c|c|cc}\hline
    & \multicolumn{5}{c}{\textbf{Out-of-domain}} \\
     
    & \multicolumn{2}{c|}{\textit{Google Scanned Object}~\cite{Downs2022GoogleSO}} & \multicolumn{2}{c|}{\textit{NAVI}~\cite{jampani2023navi}} & \multicolumn{1}{c}{\textit{ScanNet}~\cite{dai2017scannet}} \\
    & \multicolumn{1}{c|}{Hard (60-90 degree)} & \multicolumn{1}{c|}{Easy (15-45 degree)} & \multicolumn{1}{c|}{Multiview} & \multicolumn{1}{c|}{Wild} \\ 
    Method & \multicolumn{1}{c|}{AUC@5\degree / 10\degree / 20\degree} & \multicolumn{1}{c|}{AUC@5\degree / 10\degree / 20\degree} & \multicolumn{1}{c|}{AUC@5\degree / 10\degree / 20\degree} & \multicolumn{1}{c|}{AUC@5\degree / 10\degree / 20\degree} & \multicolumn{1}{c}{AUC@5\degree / 10\degree / 20\degree} \\ \shline
    PDCNet~\cite{Truong2021pdcnet} & 2.6 / 4.8 / 8.4 & 13.5 / 22.4 / 33.0 & 1.7 / 3.7 / 6.6 & 2.9 / 6.1 / 10.4 & 16.4 / 33.7 / 51.2 \\
    LoFTR~\cite{Sun2021LoFTR} & 3.6 / 7.3 / 13.0 & 20.7 / 33.9 / 47.9 & 5.7 / 11.8 / 20.4 & 4.5 / 9.4 / 17.0 &  16.9 / 33.6 / 50.6\\
    \hline
    SIFT~\cite{Lowe2004sift}+MNN & 3.4 / 6.5 / 11.5 & 16.7 / 30.1 / 40.8 & 3.3 / 6.9 / 12.8 & 2.8 / 5.9 / 11.7 & 1.7 / 4.8 / 10.3\\
    SuperPoint~\cite{detone2018superpoint}+MNN & 2.5 / 5.3 / 10.0 & 15.2 / 26.1 / 38.8 & 4.5 / 9.7 / 17.8 & 3.7 / 8.0 / 15.1 & 7.7 / 17.8 / 30.6\\
    \hline
    DINOv2~\cite{Oquab2023DINOv2}+SG~\cite{sarlin2020superglue} & 1.8 / 3.6 / 7.4 & 5.5 / 11.6 / 21.3 & 3.3 / 9.7 /.155.6  & 3.8 / 8.4 / 16.3 & 3.3 / 10.0 / 22.0\\
    SuperGlue~\cite{sarlin2020superglue} & 3.4 / 6.9 / 12.2 & 17.5 / 30.1 / 42.6 & 5.1 / 11.2 / 19.9 & 4.8 / 10.2 / 18.3 & 10.4 / 22.9 / 37.2\\
    LightGlue~\cite{Lindenberger2023LightGlue} & 3.5 / 7.1 / 12.6 & 18.9 / 32.3 / 46.7 & 5.7 / 12.4 / 21.2 & 4.3 / 9.2 /15.7 &  15.1 / 32.6 / 50.3\\ 
    \multirow{1}{*}{\textbf{OmniGlue (ours)}} & 4.1 / 8.2 / 14.3 & 20.7 / 34.1 / 48.4 & 5.8 / 12.6 / 22.2 & 5.6 / 11.8 / 20.7 & 14.0 / 28.9 / 44.3\\
    \hline
\end{tabular}
\vspace{-0.1in}
\caption{\small{Relative camera pose estimation performance (AUC) and zero-shot generalization capability of models trained on MegaDepth dataset.}}
\label{tabel: main_auc}
\vspace{-0.05in}
\end{table*}

\section{Latency analysis.}
We note that novel OmniGlue modules do not hurt latency as compared the baseline SuperGlue model. Even though DINOv2 introduces additional computation, we use its features to prune the graphs and reduce the computation accordingly.

Theoretically, the computation that DINOv2 introduces is $O(n_1 (h w)^2)$, where $n_1=9$ (number of DINOv2 attention layers), $h=\frac{H}{14}$ and $w=\frac{W}{14}$ ($H$ and $W$ are input resolution to DINOv2). The computation that pruning saves is $O(2 n_2 k k^{'})$, where $n_2=9$ (number of information propagation blocks), $k=1024$ (number of target keypoints in one image), $k^{'}=\frac{k}{2}$ (number of pruned keypoints in the other image) and the coefficient $2$ is multiplied because there are $2$ inter-graph aggregation modules in each block. It is simplified as $O(n_2 k^2)$. With the resolution $W=630$ and a typical aspect ratio of 16:9, the $h w \approx k = 1024$. Thus, the introduced and saved computation are balanced. 

We report the empirical speed results in Table~\ref{table:appendix_latency}, which shows that OmniGlue runs at a similar frame rate as the baseline SuperGlue model (no graph pruning). Inference was performed on an NVIDIA A40 GPU with FlashAttention. The result is reproduced with using Glue-Factory.

\begin{table*}[ht]
\centering
\begin{tabular}{l|cc}
& SuperGlue & OmniGlue \\ \shline
Speed (FPS) & 52 & 51 \\
\end{tabular}
\caption{Latency analysis, comparing SuperGlue and our OmniGlue. For both models, we include feature extraction (SuperPoint) and feature matching inference times. Additionally, we include DINOv2 inference time in our measurements for OmniGlue.}
\label{table:appendix_latency}
\end{table*}

\section{Additional Qualitative Results}
\begin{figure*}[t]
    \centering

    \begin{subfigure}{0.33\textwidth}
        \includegraphics[width=\linewidth]{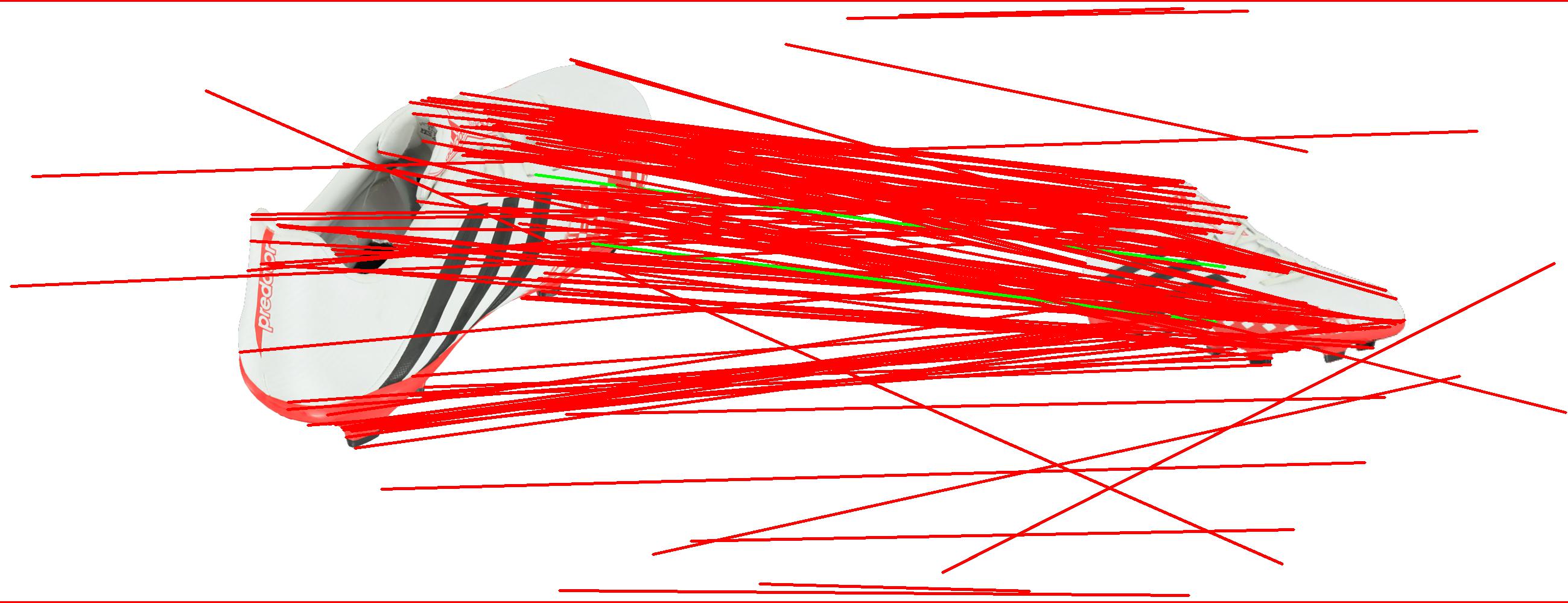}
    \end{subfigure}
    \begin{subfigure}{0.33\textwidth}
        \includegraphics[width=\linewidth]{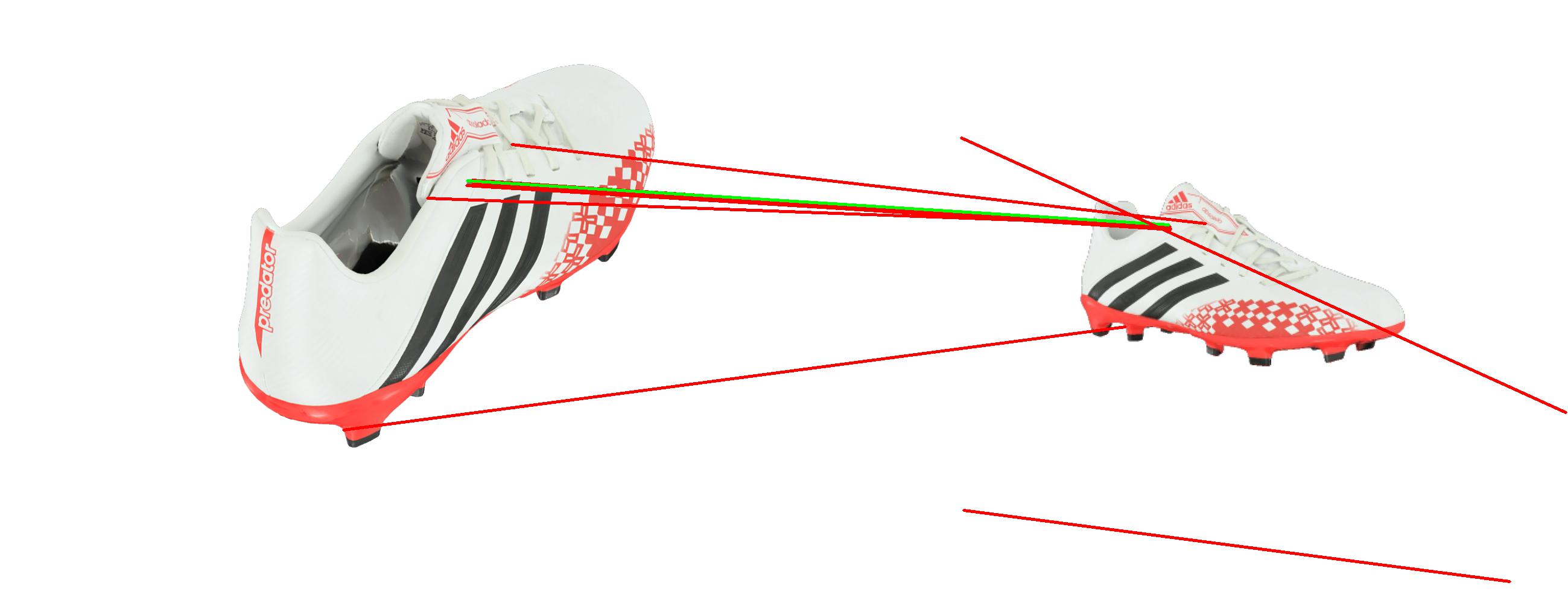}
    \end{subfigure}
    \begin{subfigure}{0.33\textwidth}
        \includegraphics[width=\linewidth]{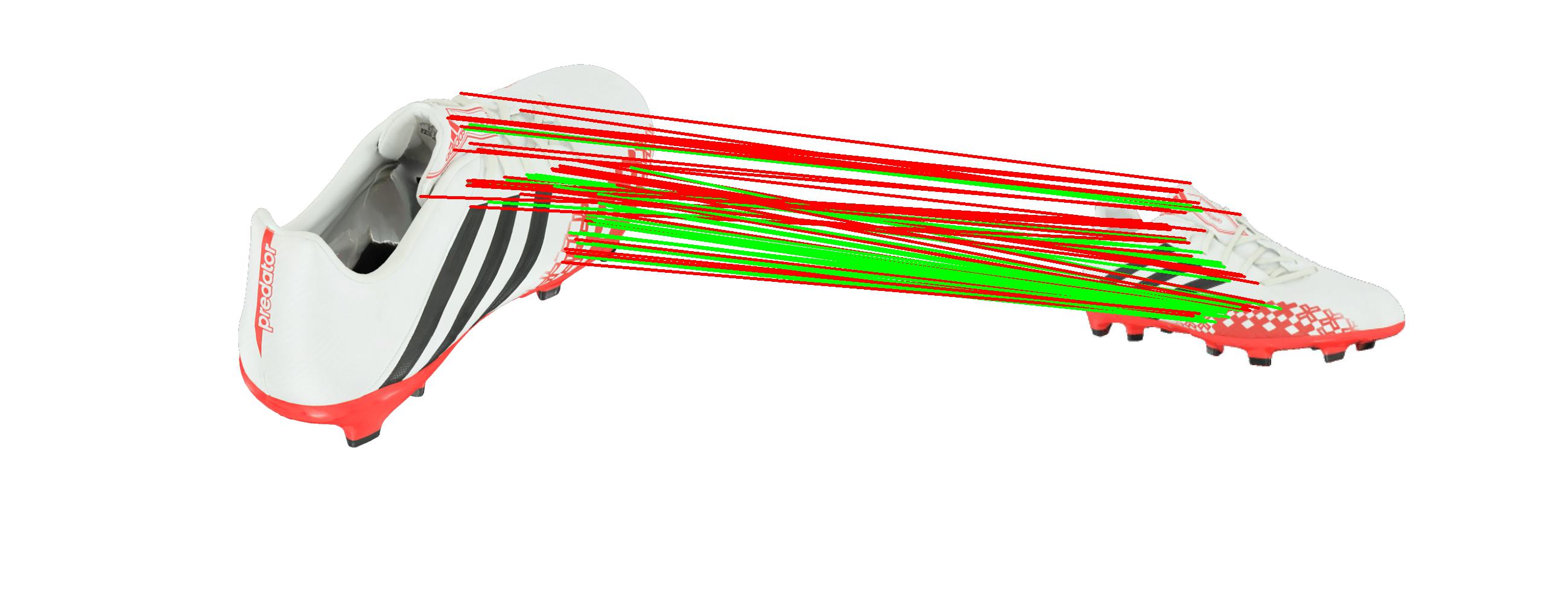}
    \end{subfigure}

    \bigskip
    
    \begin{subfigure}{0.33\textwidth}
        \includegraphics[width=\linewidth]{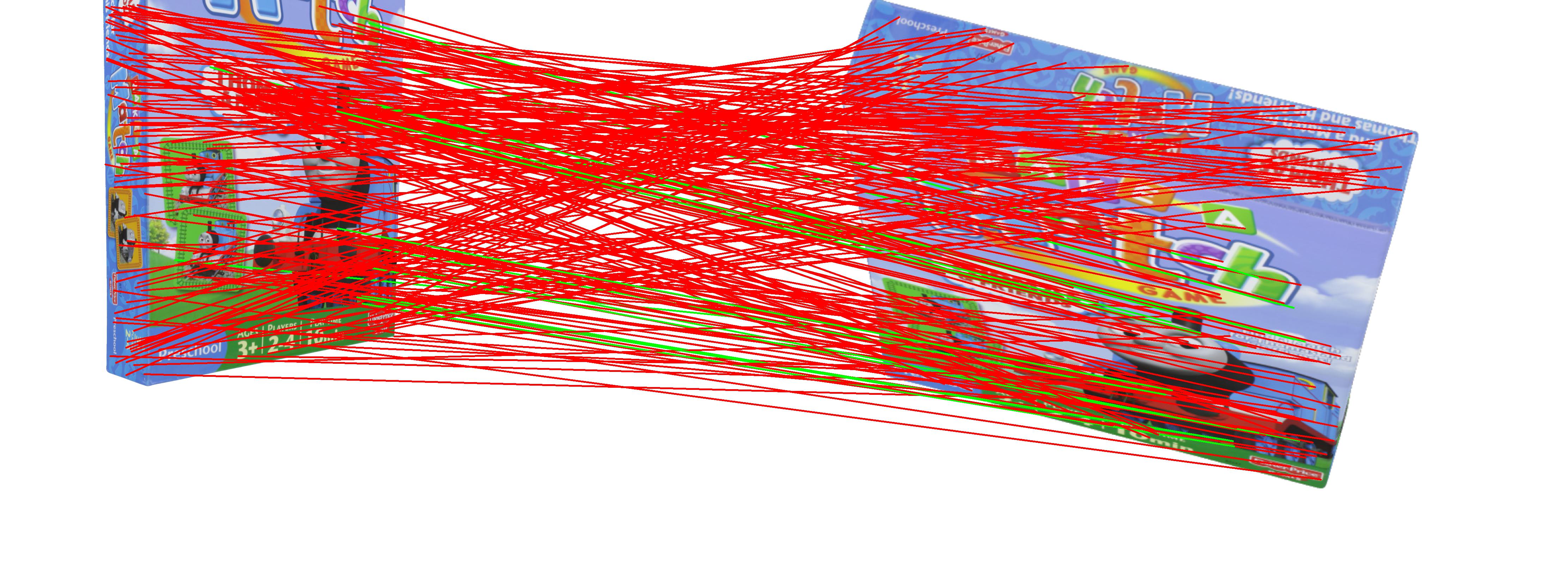}
    \end{subfigure}
    \begin{subfigure}{0.33\textwidth}
        \includegraphics[width=\linewidth]{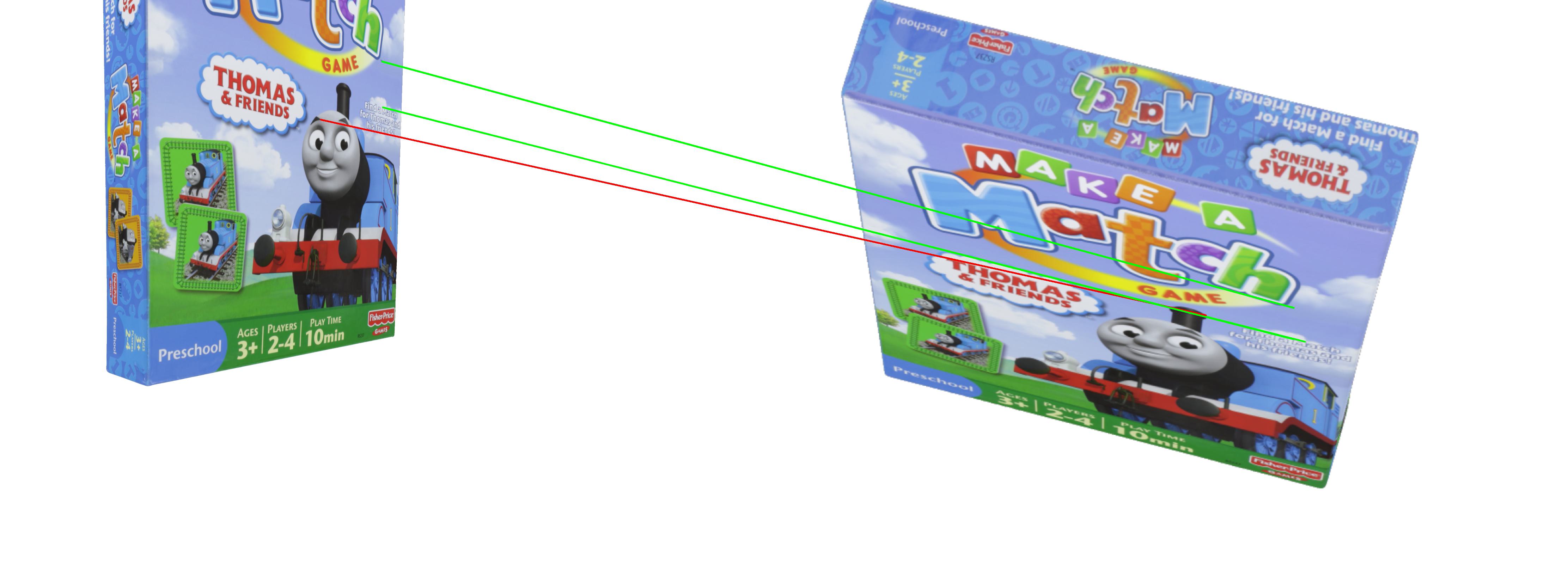}
    \end{subfigure}
    \begin{subfigure}{0.33\textwidth}
        \includegraphics[width=\linewidth]{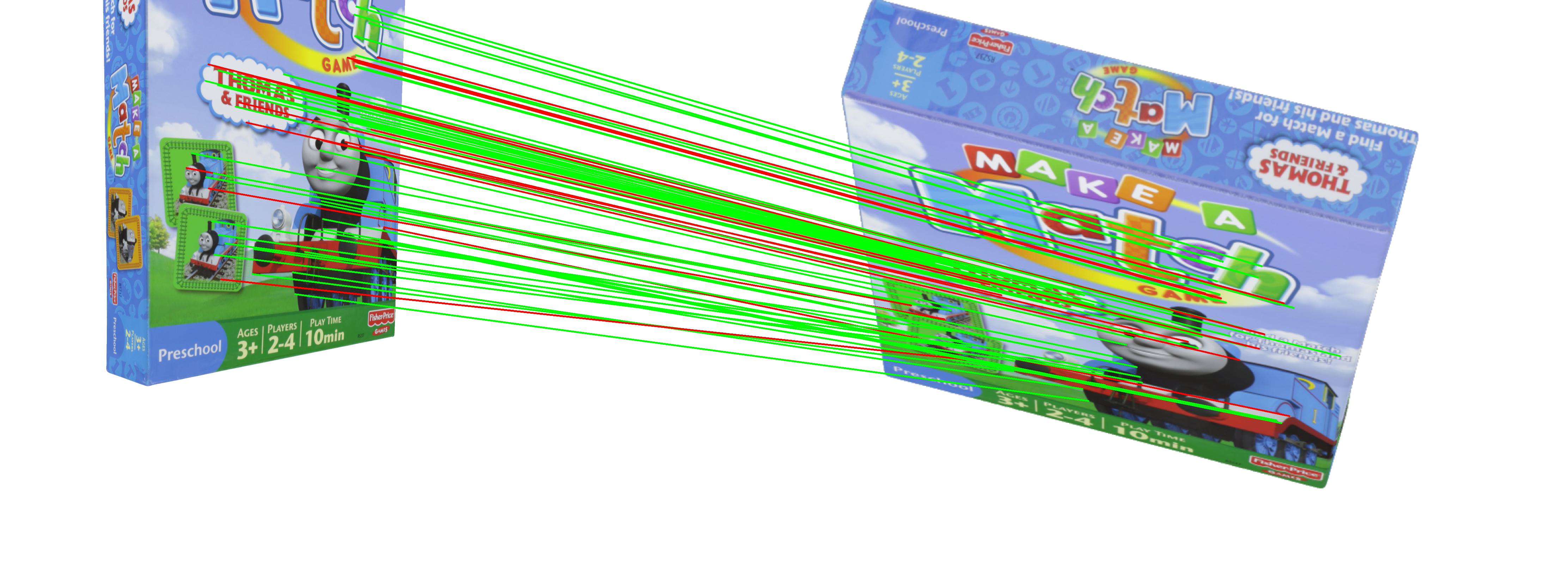}
    \end{subfigure}

    \bigskip

    \begin{subfigure}{0.33\textwidth}
        \includegraphics[width=\linewidth]{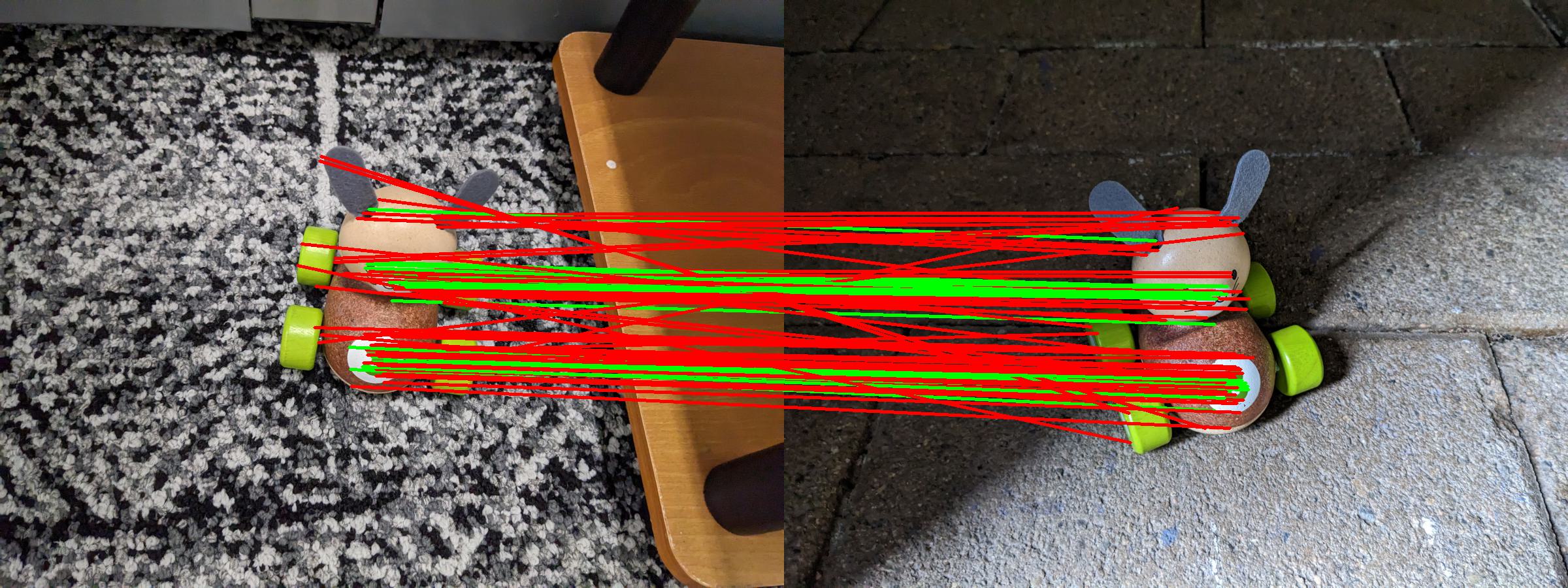}
    \end{subfigure}
    \begin{subfigure}{0.33\textwidth}
        \includegraphics[width=\linewidth]{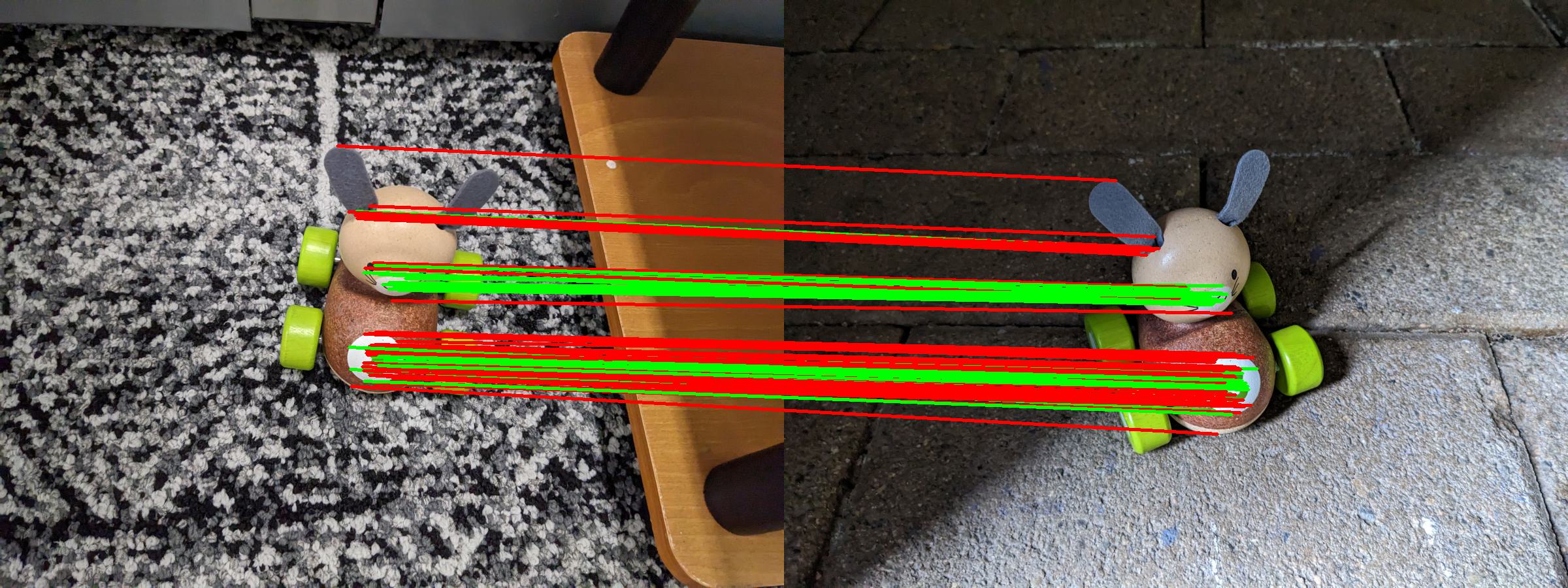}
    \end{subfigure}
    \begin{subfigure}{0.33\textwidth}
        \includegraphics[width=\linewidth]{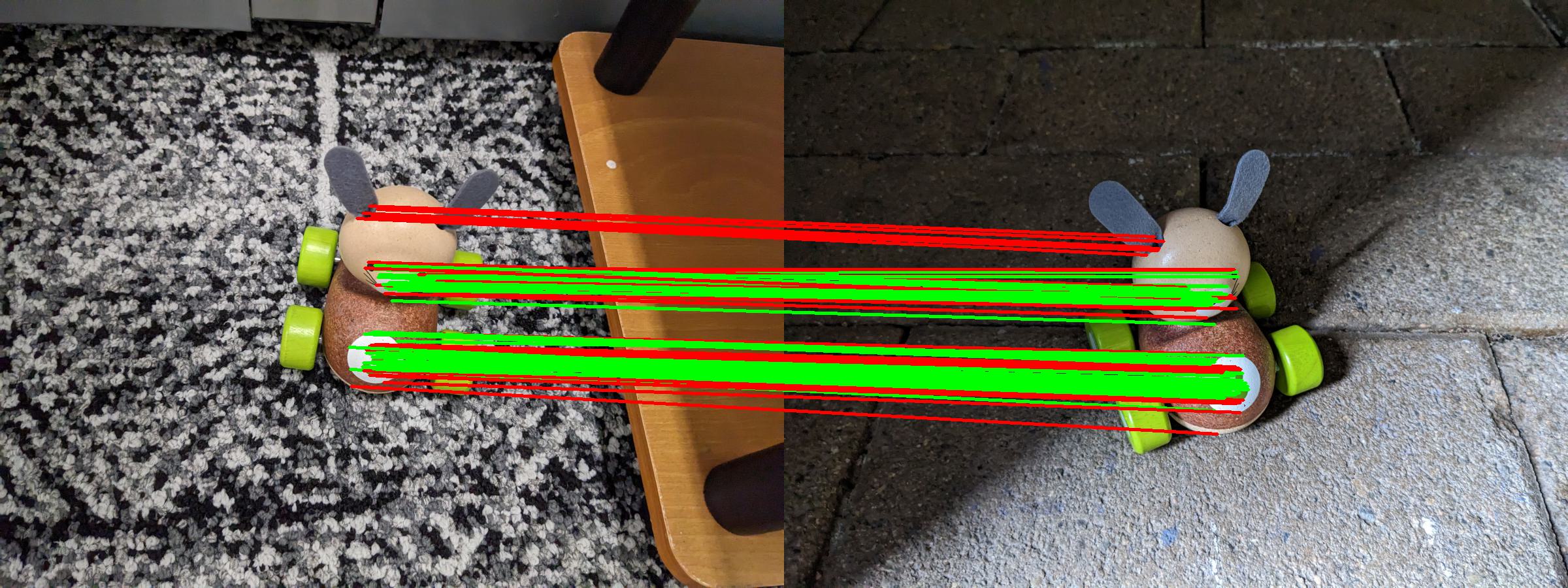}
    \end{subfigure}

    \bigskip
    
    \begin{subfigure}{0.33\textwidth}
        \includegraphics[width=\linewidth]{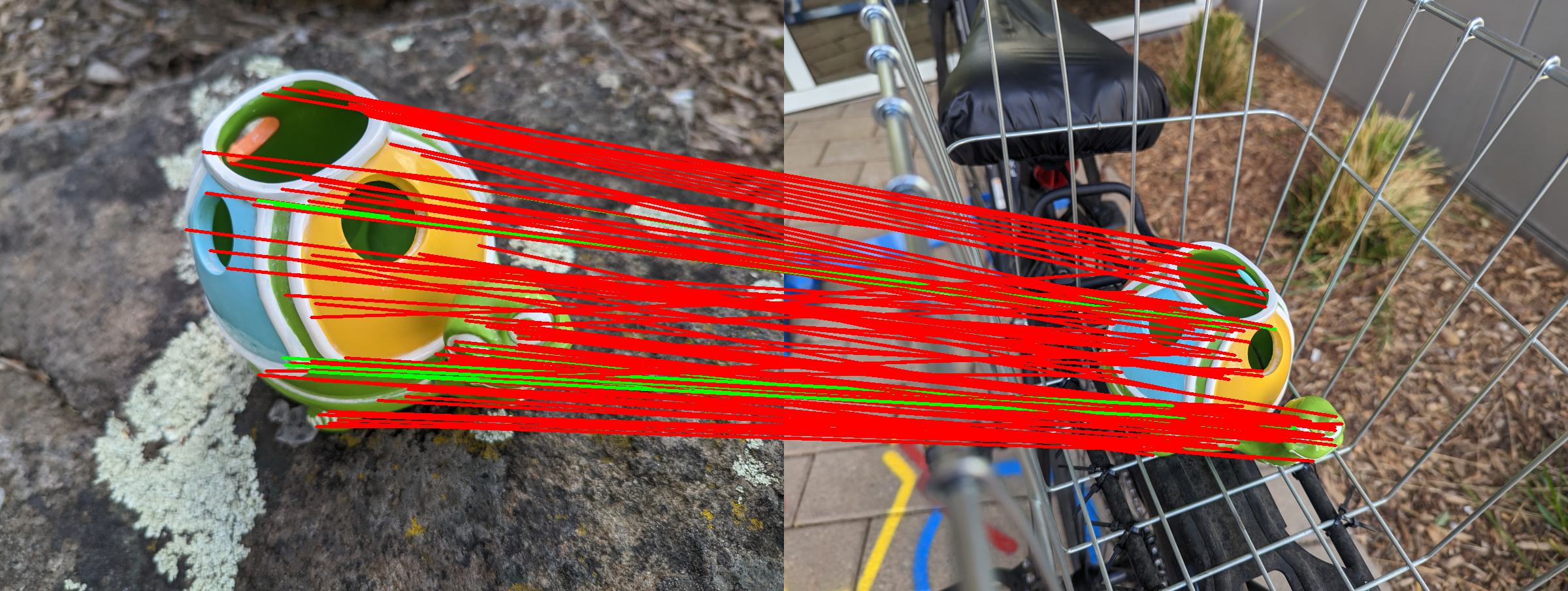}
    \end{subfigure}
    \begin{subfigure}{0.33\textwidth}
        \includegraphics[width=\linewidth]{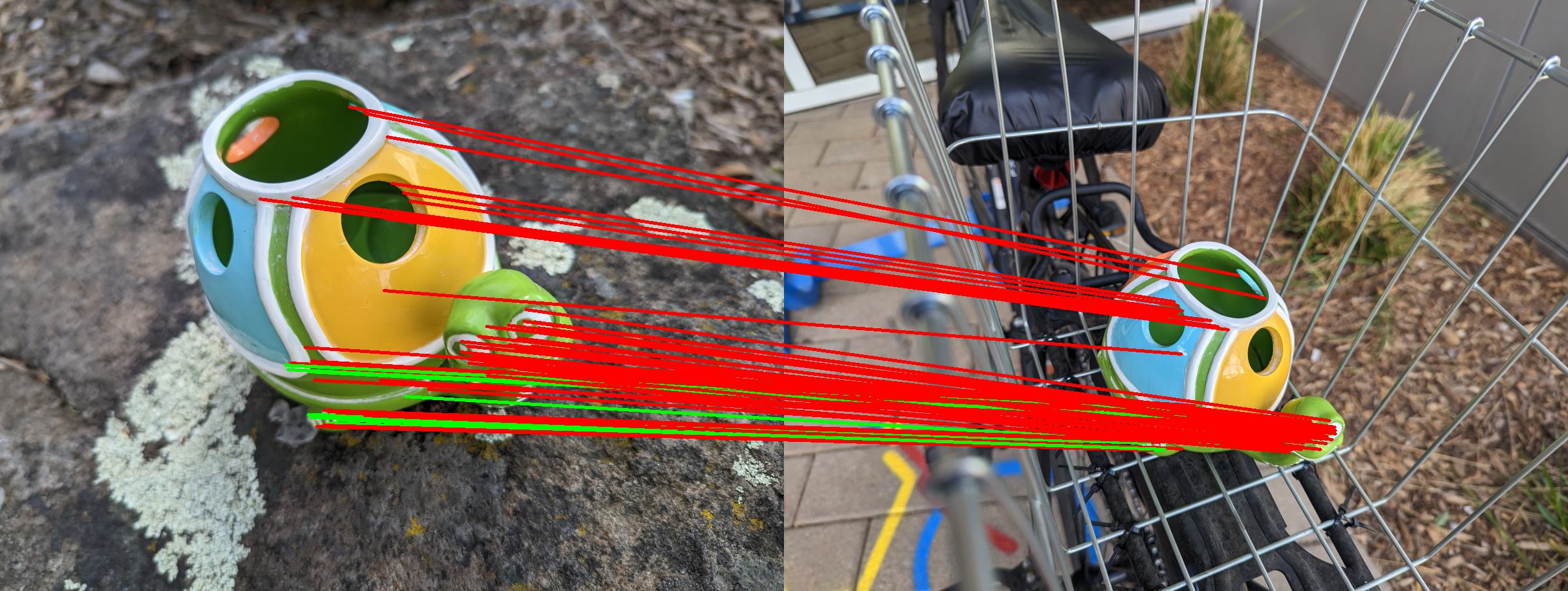}
    \end{subfigure}
    \begin{subfigure}{0.33\textwidth}
        \includegraphics[width=\linewidth]{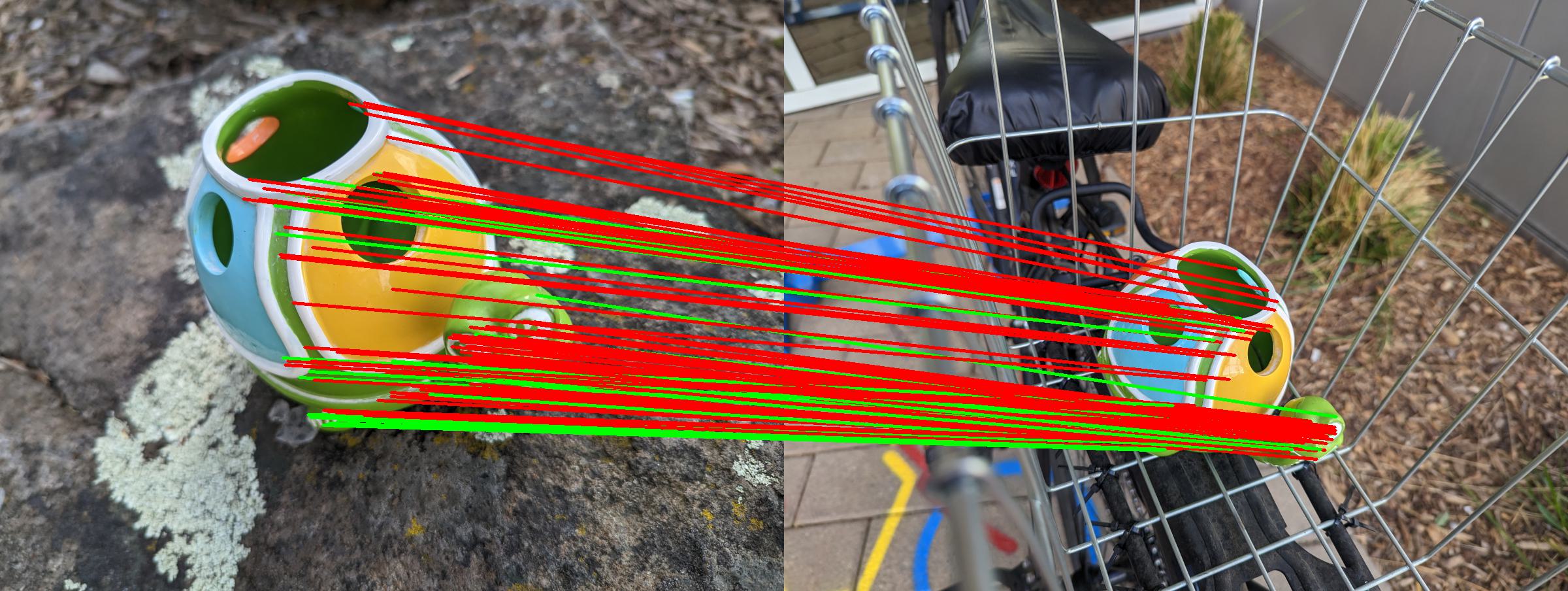}
    \end{subfigure}
    
    \bigskip
    
    \begin{subfigure}{0.33\textwidth}
        \includegraphics[width=\linewidth]{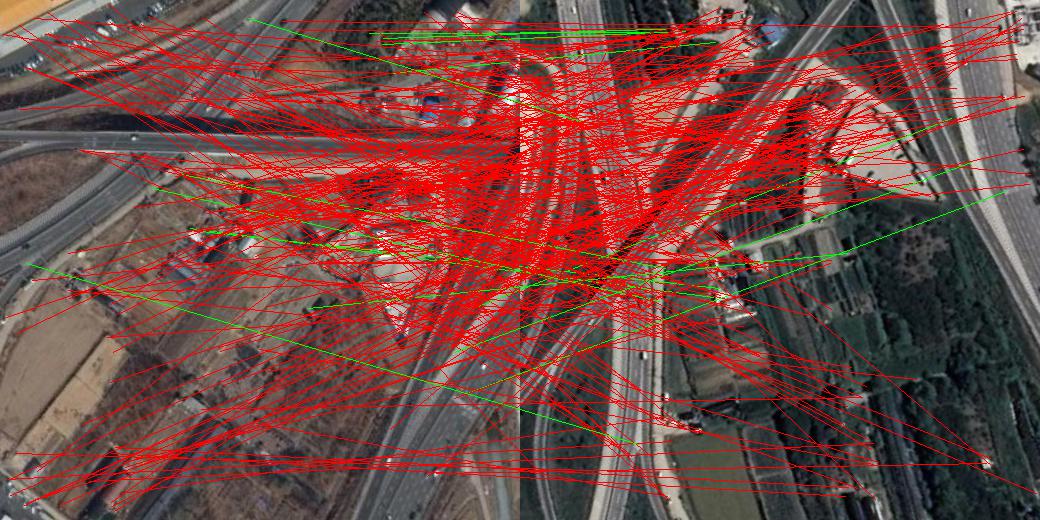}
    \end{subfigure}
    \begin{subfigure}{0.33\textwidth}
        \includegraphics[width=\linewidth]{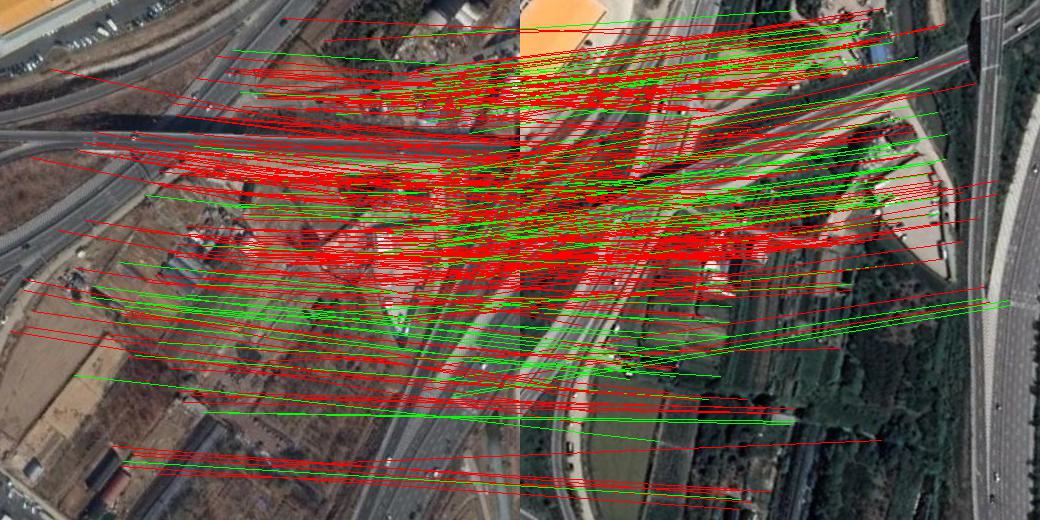}
    \end{subfigure}
    \begin{subfigure}{0.33\textwidth}
        \includegraphics[width=\linewidth]{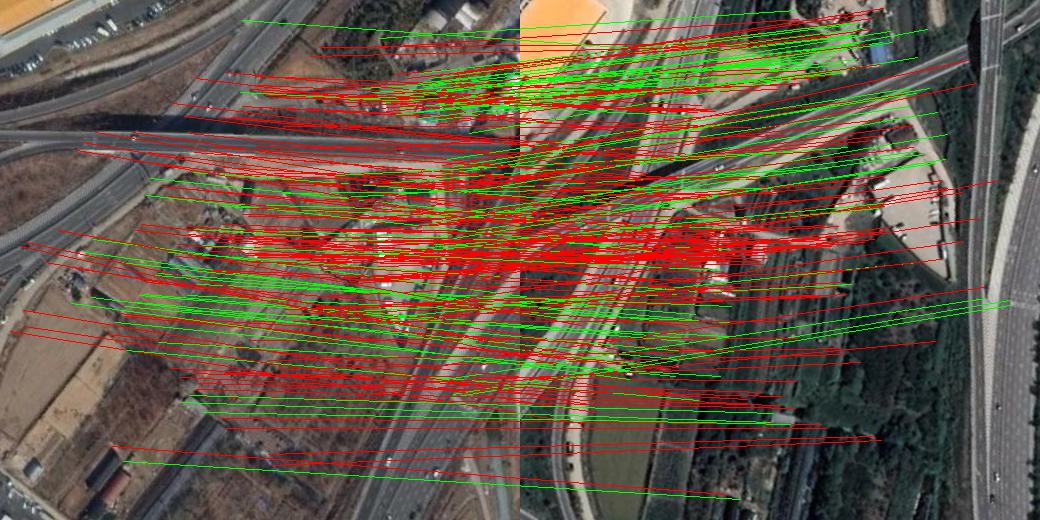}
    \end{subfigure}
    
    \bigskip
    
    \begin{subfigure}{0.33\textwidth}
        \includegraphics[width=\linewidth]{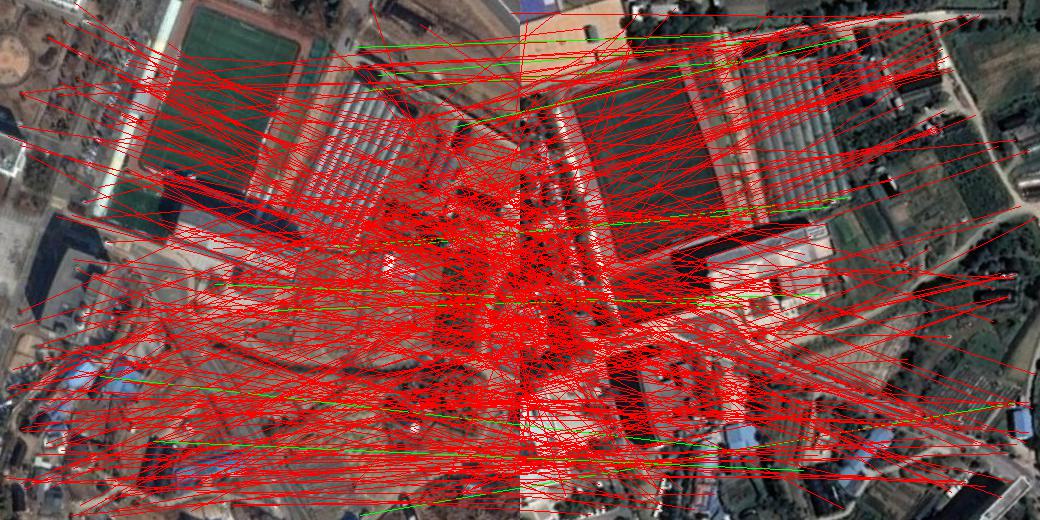}
    \end{subfigure}
    \begin{subfigure}{0.33\textwidth}
        \includegraphics[width=\linewidth]{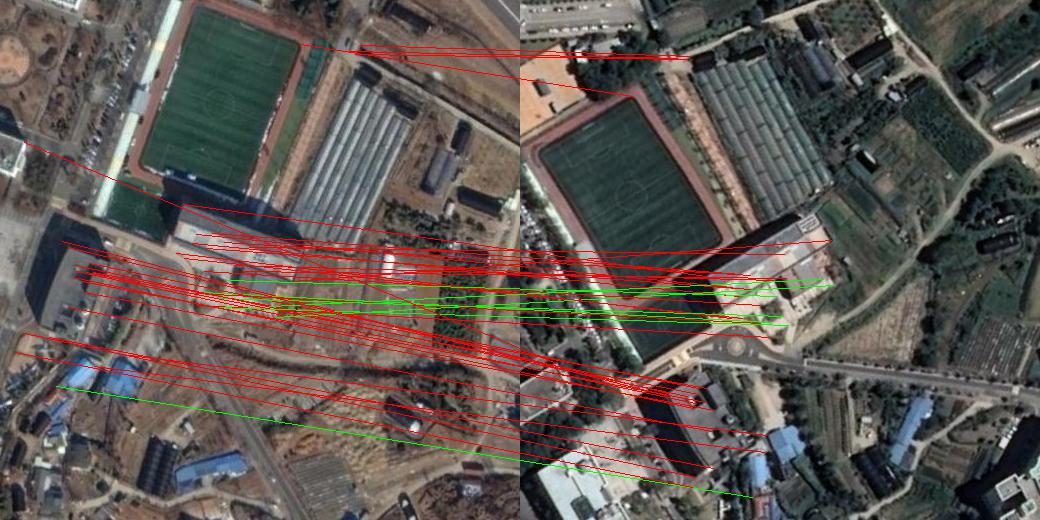}
    \end{subfigure}
    \begin{subfigure}{0.33\textwidth}
        \includegraphics[width=\linewidth]{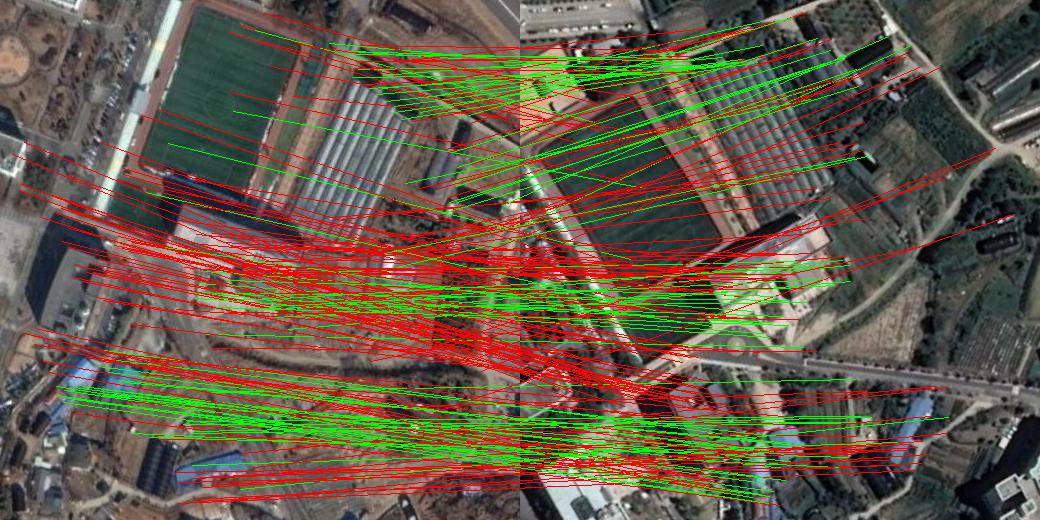}
    \end{subfigure}
    
    \caption{\textbf{Qualitative matching comparison.} We compare the following methods: mutual nearest neighbor (MNN, left), SuperGlue (center) and \modelname{} (right). Green lines denote correct correspondences, while red ones denote incorrect predictions.
    The first two rows present results on Google Scanned Objects (Hard), the following two rows on the NAVI Wild Set, and the final two rows on DeepAerial. The MNN results use SuperPoint features in the first two rows, and SIFT features in the others.}
    \label{fig:matching_viz_appendix}
    
\end{figure*}

We additionally present qualitative results of \modelname{} in Figure~\ref{fig:matching_viz_appendix}.
We compare our method (last column) with two reference matching methods: mutual nearest neighbors (MNN, first column) and SuperGlue~\cite{sarlin2020superglue} (second column).
We show MNN with SIFT~\cite{Lowe2004sift} features for two domains, and with SuperPoint~\cite{detone2018superpoint} features for one.
We observe that \modelname{} produces improved matches for image pairs with significant changes in viewing conditions, across a range of domains.

{\small
\bibliographystyle{ieee_fullname}
\bibliography{egbib}
}

\end{document}